\documentclass{elsarticle}

\usepackage{graphicx} 
\usepackage{amsmath}
\usepackage{longtable}
\usepackage{subcaption}
\usepackage{xspace} 
\usepackage{algorithm}
\usepackage{algorithmic}
\usepackage{caption} 
\usepackage{soul}

\date{August 2024}

\begin{document}
\sloppy 
\begin{frontmatter}



\title{Deepfake Detection with Optimized Hybrid Model: EAR Biometric Descriptor Extraction via Improved RCNN}


\author{Ruchika Sharma, Rudresh Dwivedi}

\affiliation{organization={Department of Computer Science and Engineering, Netaji Subhas University of Technology},
            addressline={Dwarka}, 
            city={Delhi},
            postcode={110078}, 
            state={NCT of Delhi},
            country={India}}

\begin{abstract}
Deepfake is a widely used technology employed in recent years to create pernicious content such as fake news, movies, and rumors by altering and substituting facial information from various sources. Given the ongoing evolution of deepfakes investigation of continuous identification and prevention is crucial.  Due to recent technological advancements in AI (Artificial Intelligence) distinguishing deepfakes and artificially altered images has become challenging. This approach introduces the robust detection of subtle ear movements and shape changes to generate ear descriptors. Further, we also propose  a novel optimized hybrid deepfake detection model that considers the ear biometric descriptors via enhanced RCNN (Region-Based Convolutional Neural Network). Initially, the input video is converted into frames and preprocessed through resizing, normalization, grayscale conversion, and filtering processes followed by face detection using the Viola-Jones technique. Next, a hybrid model comprising DBN (Deep Belief Network) and Bi-GRU (Bidirectional Gated Recurrent Unit) is utilized for deepfake detection based on ear descriptors. The output from the detection phase is determined through improved score-level fusion. To enhance the performance, the weights of both detection models are optimally tuned using the SU-JFO (Self-Upgraded Jellyfish Optimization method). Experimentation is conducted based on four scenarios: compression, noise, rotation, pose, and illumination on three different datasets. The performance results affirm that our proposed method outperforms traditional models such as CNN (Convolution Neural Network), SqueezeNet, LeNet, LinkNet, LSTM (Long Short-Term Memory), DFP (Deepfake Predictor) \cite{raza2022novel}, and ResNext+CNN+LSTM \cite{vamsi2022deepfake} in terms of various performance metrics viz. accuracy, specificity, and precision.\\ \\
\end{abstract}

\begin{graphicalabstract}
	\includegraphics[scale=0.7]{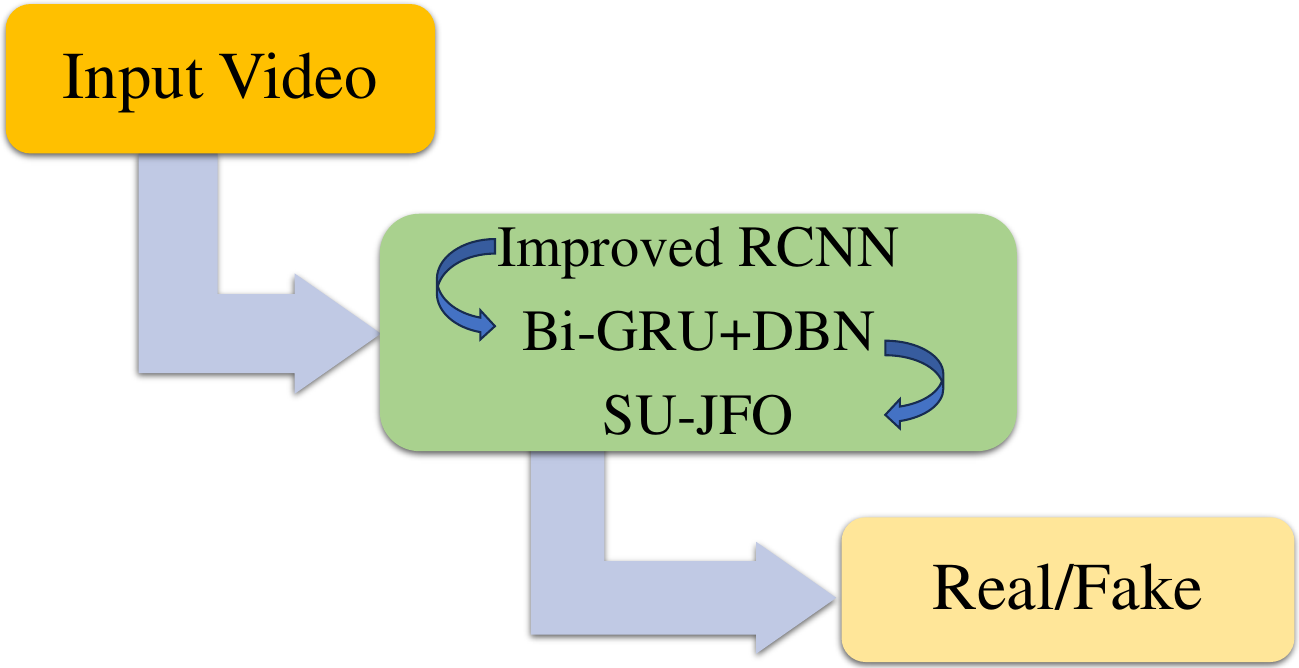}
\end{graphicalabstract}

\newpage

\begin{highlights}
\item Our proposed model has detected deepfakes using an improved RCNN. Deepfake detection through ear biometrics includes the size and shape to enhance feature extraction for more realistic deepfake detection taking compression, noise, rotation , poses and illumination cases into consideration.

\item  Our proposed model is a hybrid model combining Bi-GRU and DBN models with optimized weights to detect deepfakes.

\item  A novel fusion process integrates the outputs from both the models (Bi-GRU and DBN) improving the detection accuracy and reliability.

\item  Our proposed work introduces a novel activation function called as hyper sig activation function which enhances the efficiency and accuracy in detecting deepfakes.

\end{highlights}
\newpage
\begin{keyword}
Deepfake, Artificial Intelligence, Bidirectional Gated Recurrent Unit, Long Short-Term Memory, Deep Belief Network, Jelly Fish Optimization
\end{keyword}

\end{frontmatter}

\section{Introduction}
A relatively new area of AI technology called "deepfake" is gaining popularity on social networking sites involving layering one person's face over another. The ease of access to new technologies has led to the expansion of deepfake videos on social media platforms. Deepfake has come out as one of the most significant challenges faced on social media platforms nowadays. Such instances can be observed in films where celebrities faces are digitally imposed onto other bodies \cite{silva2022deepfake}. The widespread use of deepfakes has impacted society by propagating misinformation on digital platforms \cite{vamsi2022deepfake}. Deepfakes are concerned specifically with the manipulation of images, videos, and sounds that are generated by GAN (Generative Adversarial Networks) primarily resulting from the advancements in NNs (Neural Networks) \cite{tran2022generalization}. Deepfake technology enhances the advancements in the field of entertainment, cross-cultural interaction, and education which improves not only the quality of education but also the overall quality of life \cite{kang2022detection}. The proposed work enhances deepfake detection by focusing on ear biometric thereby improving accuracy and robustness in detecting subtle ear movements and shapes. Accurate tracking of the ear movements is a challenging task that requires human assistance. In our work, we have proposed a solution that synchronizes different feature extraction mechanism with better computational efficacy.

However, Deepfake is widely used to produce erroneous data and digital proof misleading the public as well as disrupting the rules of the society. This strategy has been developed among the most advanced ways to launch network attacks. Deepfake can create fake images and videos that are difficult for humans to discern leading to societal harm \cite{xu2023novel}. With Deepfake which largely relies on DNN (Deep Neural Network), creating a video clip of an individual expressing themselves or performing an action they didn't perform only requires a few clicks. Such videos have impacted individuals to a greater extent than anticipated potentially altering future perceptions of media accuracy \cite{xu2021deepfake}. While technological innovation is generally benign for recreational use but some individuals may exploit it for malicious or political purposes resulting in serious consequences.

Typically, ML (Machine Learning) is coupled with other detection techniques. Researchers utilize artificial features such as head postures, facial expressions, or specific patterns of facial movements and emotions during the speech to classify the semantics \cite{tolosana2022deepfakes}. However, it is uncommon for investigators to employ neural networks to discern variations between fake and authentic videos. On the contrary, the research for deepfake video detection is predominantly in its beginning.

DL (Deep Learning) has proven to be an effective and valuable strategy in numerous AI (Artificial Intelligence), NLP (Natural Language Processing), and other applications \cite{mitra2021machine}. In recent years, the development of multiple new algorithms that utilize DL for identifying deepfakes has occurred along with the efforts to enhance the understanding of their workings \cite{ismail2022integrated}. Big data analytics, human-level management, and computer vision are among the complex issues that DL has successfully addressed. Consequently, DL techniques are used to develop programs that could raise the security concerns. In this context, our paper introduces an optimized hybrid deep learning model for deepfake detection. The key contribution of the deepfake detection model are given in the following:
\begin{itemize}
   \item[\textbf{1.}] Proposed an improved RCNN-based Ear detection taking into account ear attributes such as the size and shape resulting in an enhanced accuracy and robustness to detect deepfakes against variations in ear appearance under different conditions such as compression, rotation, noise, pose and illumination.
   
   \item[\textbf{2.}] Proposed a hybrid deep learning model for deepfake detection by combining Bi-GRU (Bi-Directional Gated Recurrent Unit) and DBN (Deep Belief Networks) models. Bi-GRU captures temporal dependencies and sequence patterns are required for detecting deepfakes whereas DBN allows us to learn complex features. Also, the weights of both the models are optimally tuned using SU-JFO (Self-Upgraded Jellyfish Optimization) method ensuring better convergence and learning efficiency to accurately distinguish between real and fake content. The SU-JFO algorithm adapts advanced position updating strategies which enhances convergence speed in comparison to standard Jelly Fish algorithm due to the incorporation of more refined parameter such as distribution coefficient and additional swarm dynamics leading from more effective convergence to optimal solution.
   
   \item[\textbf{3.}] Proposed an improved score-level fusion process that integrates the outcomes of trained models namely Bi-GRU and DBN. The integration of multiple models reduces the likelihood of false positives and hence achieves more accurate and reliable deepfake detection, making it more robust against adversarial attacks.
   
   \item[\textbf{4.}] Proposed a novel activation function named as hyper sig activation function. We are the first to introduce this activation function formed by combining the properties of hyperbolic sigmoid and bipolar sigmoid function thereby enhancing the accuracy of ear detection. This refines the bounding box predictions for ear localization, ensures non-linearity, and gradient flow thereby improving the model's precision, stability, and improved generalization. This novelty handles complex scenarios such as varying poses and lightening resulting in robust ear detection performance. 
\end{itemize}

The rest of our paper is organized as follows: The literature review in the context of deepfake detection is described in section \ref{section2}. The proposed model and the methodology adopted for the proposed model has been described in section \ref{section3}. The experimental evaluations and discussion of our proposed model are reported in section \ref{section4}. Section \ref{section5} concludes the paper.

\section{Literature Review}\label{section2}

Ahmed et al. \cite{ahmed2023perception} focused on the connection between modalities and people's propensity to accept and disseminate various types of deepfakes. To investigate the effects of single versus multimodality on people's assessed claims validity and communication motives, the investigation used a web-based survey carried out in the United States. In particular, three misinformation conditions such as cheap fakes, audio deepfakes, and video deepfakes were randomly allocated to participants. Additionally, the impact of cognitive ability on a desire to contribute throughout conditions as well as the impression of claim correctness were also examined. The results showed that the consumers were more likely to believe that the audio deepfakes were less accurate than video deepfakes in comparison to inexpensive fakes. However, people are less likely to share video deepfakes in comparison to cheapfakes. Additionally, the authors discovered that those with high cognitive abilities were less tending to share or understand deepfakes as reliable. After thorough experimental analysis, the study suggested that modalities associated with deepfakes should be considered.
  
Raj et al. \cite{raj2023fdt} suggested an FDT (Fake Detection Tool) which streamlines the fake detection process by combining many modification approaches facilitating users to recognize and visualize fake content. To stream tweets with images of faces based on hashtags the program was also integrated with Twitter. It offers an output data frame and uses pie charts for improved visualization when presenting information on sentiments, virality, and other topics. The proposed method uses a variety of big datasets for learning how to handle fakes in the real world. It was a free, effective, and easily navigable phony detection program.

Dong et al. \cite{dong2023contrastive} proposed an innovative deepfake detection approach with excellent generalizability. Supervised comparison learning was utilized to improve the generalizability of unfamiliar datasets and interventions. Furthermore, researchers developed a cross-modality data enhancement technique that combines RGB as well as SRM (Steganalysis rich model) features to fully retrieve identification hints. Moreover, they suggest a module for enhancing textural and semantic information using multi-scale features. Wide-ranging experiments indicate that this method enhances the generalization of models within dataset as well as under cross-dataset situations.

Alnaim et al. \cite{alnaim2023dffmd} suggested a DFFMD (Deepfake Face Mask Dataset) using a unique Inception-ResNet-v2 that includes batch normalization, feature-based approaches residual connections, and preprocessing steps. As opposed to conventional techniques, a combination based on features residual relationships, batch normalization, and preprocessing phases improves the recognition accuracy of deepfake video when facemasks are present. When the investigation's findings were compared with the current cutting-edge techniques, face-mask deepfakes were identified with an accuracy of 99.81\% as opposed to the conventional Inception-ResNet-v2 having an accuracy of 77.48\% and  VGG19 having an accuracy of 99.25\% . Subsequent research endeavors ought to assess the precision of devising an additional experimental study aimed at augmenting the identification of deepfake facial masks.

Patel et al. \cite{patel2023improved} presented a new and upgraded D-CNN (Deep Convolutional Neural Network) structure with good generalization and acceptance accuracy for identifying deepfakes. The model undergoes training using images from many sources enhancing its generalizability. After scaling the images are sent into the D-CNN model. An Adam optimizer and binary-cross entropy were used to increase the D-CNN model's training percentage. Seven distinct datasets comprising 10,000 genuine images and 5000 deepfake images were examined for the reconstruction task. The accuracy of the suggested model in AttGAN ( Attribute-Generative Adversarial Network), GDWCT (Group-Wise Deep Whitening-And-Coloring Transformation), StyleGAN (Style-Based Generative Adversarial Network), StyleGAN2, and StarGAN is 98.33\%, 99.33\%, 95.33\%, 94.67\%, and 99.17\% respectively in real and deepfake images indicating its feasibility in testing configurations.

Sadiq et al. \cite{sadiq2023deepfake} focused on the detection of text written by machines on social media sites such as Twitter. Using the open-access Tweepfake dataset, a straightforward DL algorithm in conjunction with embedded words was used in this study to classify tweets as either human-generated or bot-generated. Using FasText word embeddings, a CNN  architecture was developed to detect deepfake tweets. In order to demonstrate the enhanced efficacy of the suggested approach, this study utilized many ML models as reference methods. A variety of features were used in these baseline approaches such as FastText, FasText subword embeddings, Term Frequency, and Term FrequencyInverse Document Frequency. Additionally, the usefulness and benefits associated with the suggested method were shown by comparing its results to other DL models such as LSTM and CNN-LSTM in effectively addressing the particular work. According to the experimental findings, CNN architectural design and use of FasText embeddings are comparable for the efficient categorization of the tweet data achieving an exceptional accuracy rate of 93\%. 

Elhassan et al. \cite{elhassan2022dft} suggested methodology for fake video detection which offers a greater efficiency and accuracy. This paper builds on earlier research that presented the key ideas by applying additional multi-transfer learning techniques such as DenseNet121, EfficientNetB0, VGG16, DenseNet169, EfficientNetB7, VGG19, InceptionV3, MobileNet, Xception and ResNet50 to improve the method's capacity to identify and categorize Deepfake videos using natural signal characteristics extracted from the teeth and mouth frames. 

Kolagati et al. \cite{kolagati2022exposing} utilized a deep hybrid NN model for deepfake video detection. The authors retrieve data related to several facial features from the videos through facial landmark recognition. The multilayer perceptron was trained using this data to identify differences between actual and deepfake films. They extract features and train a CNN on the videos concurrently. Both designs were used to create a multi-input deepfake detection. The model was trained using a portion of the Dessa Dataset and the Deepfake Detection Competition Dataset. The model yields good classification results scoring 0.87 and having an accuracy of 84\%.

Raza et al. \cite{raza2022novel} proposed to use an effective framework to detect deepfake media. This work suggested a novel DFP (Deepfake Predictor) technique that depends on a hybrid VGG16 and CNN architecture. When developing NN algorithms, the deepfake dataset consisting of both real and fake faces were used. Comparatively, the TL (Transfer Learning) methods used were Xception, NASNet, VGG16, and MobileNet. In terms of deepfake detection, the suggested DFP method produced results with 94\% accuracy and 95\% precision. Their DFP method outperforms other cutting-edge research and TL strategies. By precisely identifying the deepfake content and protecting the deepfake victims, the study approach assists cyber security professionals in combating cyber crimes related to deepfakes.

	Vamsi et al. \cite{vamsi2022deepfake} stated that media manipulation has been widespread in recent years due to technological advancements and the simplicity with which false information may be produced. Deepfake media also known as AI-altered films are becoming more prevalent on social media platforms posing a serious threat to media integrity. It was anticipated that identifying this type of content would be extremely difficult. An approach to deepfake detection using CNN and LSTM (Long and short term memory) named as ResNext were employed as a technique to recognize Deepfake videos. Based on the Celeb-DF dataset the constructed DL model achieved an accuracy of 91\% accuracy rate. 

	Li et al. \cite{li2024adani} proposed a novel approach AdaNI (Adaptive Noise Injection) which detects deepfakes against adversarial attacks. AdaNI adjust the noise based on feature relevance. The proposed approach is adaptable as it can be easily integrated into existing DNN (Deep Neural Network) making it robust.

	Wang et al. \cite{wang2024multi}  proposed to method to detect deepfakes that have been compressed. The proposed approach utilizes frequency adaptive notch filter at an initial stage followed by spatial denoising technique and then attention based fusion method. The study reveals as most of the data on social media is compressed, therefore the methodology makes the model robust against compression scenarios. 

	Guo et al. \cite{guo2021fake}  suggested a new approach that detects deepfakes based on the manipulation traces left at the time of deepfake generation process. The aaproach proposed a novel method which involves amlgation of CNN (convolutional neural network) and attention mechanism together named as AMTEN (Adaptive Manipulation Traces Extraction Network). The proposed approach reached an accuracy of 98\%.
 
\begin{center}
\begin{longtable}[!htbp]      {|p{1.85cm}|p{2.8cm}|p{3cm}|p{3cm}|}
    \hline
    \begin{tabular}[c]{@{}c@{}}\textbf{Author(s)}\\ \textbf{Citation}\end{tabular} & \textbf{Methodologies} & \textbf{Features} & \textbf{Challenges}\\
         \hline
        Ahmed et al. \cite{ahmed2023perception}  & Deepfake across different modalities & When investigating deepfakes, the proposed model should avoid employing a singular perspective as citizen engagement may vary across different manifestations  & Required more representative studies for generalization.\\ \hline
        Raj et al. \cite{raj2023fdt}  &FDT & The proposed tool is well-organized, freely accessible software and also user-friendly & Needs additional functionality for understanding deepfakes to the software.\\ \hline
        Dong et al. \cite{dong2023contrastive} & Multi-scale feature enhancement module & Both intra-dataset and cross-dataset were performed. The proposed model enhances the model generalization & Explore the robustness of the model compared to troubles from attacks to enhance the proposed model.\\ \hline
         Alnaim et al. \cite{alnaim2023dffmd} & Deepfake detection method(DFFMD) & 	It was observed 99.81\% accuracy for detecting face-mask Deepfakes as compared to Inception-ResNet-v2 of 77.48\% and VGG19 of 99.25\% & Need to develop the experimental work for detecting deepfake facemasks using DL.\\ \hline
        Patel et al. \cite{patel2023improved} & D-CNN & The real and fake images are identified based on the D-CNN model obtained with higher accuracy of different datasets. & Need to explore different effective image upscaling systems and evaluate their impact on performance to gain enhanced insights.  \\  \hline
         Sadiq et al. \cite{sadiq2023deepfake} & CNN-LSTM & To identify deepfake text the proposed model attains 93\% accuracy for detection & 	Has to apply a more efficient detecting system in the social media platform.\\ \hline
        Elhassan et al. \cite{elhassan2022dft} & DL, Multi-learning approaches & Based on the teeth and mouth frames, identified deepfake videos and attained better accuracy in the suggested model.  & Need to analyze different algorithms for attaining higher accuracy in the identification of deepfake facial images.\\ \hline
       Kolagati et al. \cite{kolagati2022exposing} & Deep Hybrid NN model. & The proposed model attained 84\% accuracy and 0.87 AUC score for identifying the deepfake videos. & Need to investigate enhanced model with additional balanced datasets.\\ \hline
        Raza et at.\cite{raza2022novel} & DFP, VGG16, CNN & The suggested model attains 94\% accuracy and 95\% precision for detection of deepfake as compared to the traditional methods. & Have to include a blockchain-based method as well as cloud web system by safety online system for deepfake detection.\\ \hline
       	Vamsi et al.\cite{vamsi2022deepfake} &  CNN, and LSTM & In digital media, forensics detected deepfake with DL model with an accuracy of 91\% & Have to enhance the proposed model with a mobile application for preventing the distribution of false data through digital media.\\ \hline  
       	
     	Li et al. \cite{li2024adani} & AdaNI (Adaptive noise injection) & With CIFAR-10 dataset achieving an accuracy of 91.28\% and with CIFAR-100 the model reached an accuracy of 81.15\%. & Have to select right magnitude of noise which needs extensive tuning.\\ \hline
     
   	    Wang et al. \cite{wang2024multi} & Multi-task decision approach & Proposed approach to  detect compressed deepfakes & Need to explore on unseen deepfake dataset .\\ \hline 
     
       Guo et al. \cite{guo2021fake} & AMNET (Adaptive Manipulation Traces Extraction Network) & Proposed approach achieved an accuracy of 98\%  & Have to enhance the model to detect manipulation traces during compression , noise and other scenarios.\\ \hline
       
        \caption{Features and Challenges of Existing Deepfake Detection Techniques.} \label{table1}
    \end{longtable}
      \end{center}
 \vspace{-2mm}
Detecting deepfakes using DL algorithms presents several significant research challenges including issues related to model robustness, data quality, and computational requirements. Incorporating and synchronizing multiple aspects for more accurate deepfake detection is complex. Hence, developing deep learning approaches that effectively combine different aspects to enhance the detection accuracy is essential. Table \ref{table1} shows the features and challenges of the deepfake detection model compared to existing models. The accuracy of deepfakes continues to improve steadily leading to an increased demand for novel detection strategies. Deep and shallow learners are the two primary classifier classes utilized in deepfake detection \cite{patel2023improved,kolagati2022exposing}. Shallow classifiers can differentiate between real and fake images and videos based on irregularities in their attributes such as the absence of reflections in the eyes or discrepancies in jawline structure and ear features. To address these challenges our proposed hybrid model introduces a novel deepfake detection method centered on ear biometrics. The model follows a three-stage process—Preprocessing, Feature Extraction, and Detection combining DBN and Bi-GRU models with optimized weights using SU-JFO algorithm. This approach improves the detection accuracy by enhancing the feature extraction and recognition through ear-based biometric detection effectively overcoming existing limitations.

\section{Proposed Deepfake Detection with Optimized Hybrid Model} \label{section3}
In the proposed work, a new hybrid optimized model is introduced for deepfake detection. The model comprises three stages: preprocessing, feature extraction, and detection based on ear biometrics. The detection process in the proposed model is outlined as follows:
\begin{itemize}
    \item [\textbf{1.}] Initially, the preprocessing phase involves converting the input video into frames. For each frame, preprocessing steps include resizing and normalizing, converting to gray scale, filtering, and face detection using the Viola-Jones Algorithm.
    \item[\textbf{2.}] Subsequently, an improved RCNN-based ear detection is conducted along with extracting ear attributes and AAM-based features from the preprocessed phase.
    \item[\textbf{3.}] The last phase involves the detection of deepfakes through a hybrid recognition method which combines the models  DBN and Bi-GRU. The weights of these models are optimized using the SU-JFO algorithm. Subsequently, an enhanced score-level fusion procedure is used to determine the detection results. 
    Figure \ref{figure1} illustrates the overall design of the proposed model.
    \begin{figure}[!ht]
    \centering
    \includegraphics[width=\linewidth]{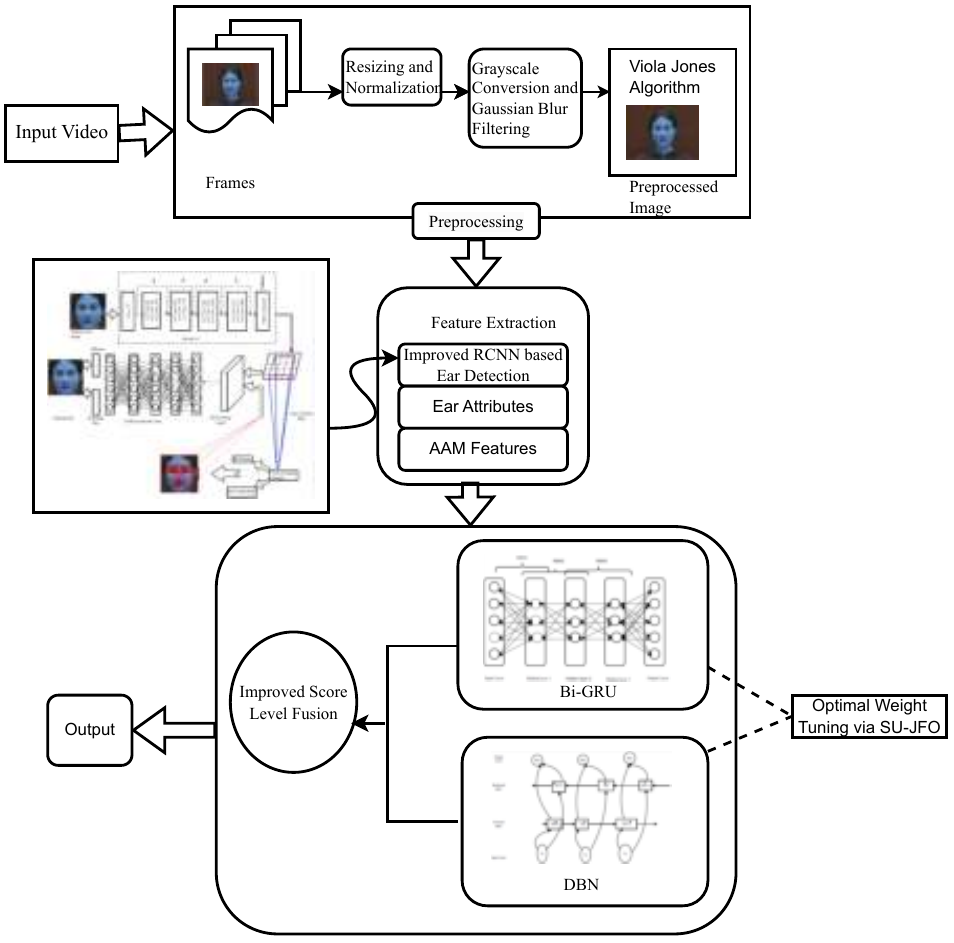}
    \caption{Structural Framework of Proposed Method.}
    \label{figure1}
\end{figure}
\end{itemize}

The purpose of using a hybrid deepfake detection model is to capture the strengths of multiple techniques for achieving greater accuracy and robustness in detecting deepfakes. The incorporation of ear biometric descriptors via an improved RCNN adds an additional layer of feature extraction focusing on segments of facial data that are less likely to be manipulated in deepfakes. This multi-faceted approach aims to enhance detection performance across various challenging scenarios including variations in noise, rotation, pose, illumination, and compression. By combining RCNN for extracting ear biometric descriptors with DBN and Bi-GRU, the model addresses various challenges posed by deepfakes including noise, rotation, pose variations, illumination changes, and compression. RCNN is used to detect and extract distinct ear biometric features which are less commonly altered in deepfakes while DBN captures high-level abstract features essential for classification. Bi-GRU further enhances detection by learning temporal patterns in video frames enabling the model to identify inconsistencies over time. Additionally, the model incorporates the SU-JFO technique to fine-tune the weights of each component improving its adaptability and overall performance under different conditions. Although the individual components such as RCNN, DBN, Bi-GRU, and optimization techniques have been applied in related fields but the specific combination of these methods for deepfake detection is novel. This approach builds on the premise that the strengths of each technique complement one another resulting in a more comprehensive and effective solution. The inclusion of ear biometric descriptors as a feature further differentiates this model leveraging the distinctiveness and stability of ear attributes in scenarios where other facial features may be manipulated. Extensive experimentation across various scenarios supports the effectiveness of this hybrid model demonstrating its ability to outperform traditional approaches in terms of accuracy, specificity, and precision.

\subsection{Preprocessing Phase}
Preprocessing serves as the initial stage for eliminating noise from the given input data. Consider the input video as $I_{v}$, the image is then converted into various frames represented as $I^{F}$. Each frame representing an image undergoes preprocessing in the following steps:
\begin{enumerate}
		\item[\textbf{1.}] \textbf{Resizing and Normalization Process:} The input frames $I^{F}$ are resized to dimensions of $224 \times 224$ pixels. Subsequently, pixel values are normalized using min-max normalization methods. This method preprocesses the frames to scale the values within a predetermined range ensuring that the scale values of various images of famous people are equivalent. The mathematical expression utilized in the min-max normalization \cite{horng2009improved} procedure is presented in Eq. \ref{eq1}, where $I^{F}_{Norm}$ denotes the normalized value of the input frames, $I^{F}_{max}$ denotes the highest value of features in the image, and  $I^{F}_{min}$ shows the minimum value of features in the frame image. Consequently, the value obtained using normalized approaches falls between 0 and 1. 
	\begin{equation}\label{eq1}
		I^{F}_{Norm}=\frac{I^{F}-I^{F}_{min}}{I^{F}_{max}-I^{F}_{min}}
	\end{equation}
	
	\item[\textbf{2.}] \textbf{Gray scale Conversion and Filtering Process:} Following the normalization process, the normalized frame (color image) is converted into a gray scale image to reduce the impact of color changes. 
	Additionally, Gaussian blur \cite{ibrahim2021gaussian} is employed to filter the gray scale image reducing the noise and smoothing the image input. Gaussian blur is a popular method for noise reduction and detail smoothing in images. The resulting filtered image is denoted as $I_{f}$.\\
	
		\item[\textbf{3.}] \textbf{Viola Jones for Face Detection:} Subsequently, the filtered images provided are subjected to face detection using a Viola-Jones algorithm \cite{bavkar2022multimodal}. This method is capable of real-time face detection requiring clear visibility of frontal upright faces. The approach initially applies basic analysis to the source image displayed through a window opening to identify human facial features. Once sufficient features are detected then an image window is recognized as a face. The process continues by increasing the window's size to capture faces of different sizes with each window scale undergoing a separate evaluation process. Every window is evaluated at several stages to minimize the number of features. While initial stages involve fewer features, subsequent stages become progressively more complex. The features of a sub-window are evaluated at each stage of the cascade classifier. If the  score of the evaluated features doesn't exceed the specified threshold, the sub-window is immediately rejected. It aid to reduce the computational costs to exclude non facial regions. The Viola-Jones face detection approach comprises three primary components: attentional cascade architecture, classifier training with Adaboost, and integral image enabling effective real-time face recognition. The result of the preprocessing image is denoted as $I^{P}_{f}$ and subsequently processed through the proposed model's subsequent steps. 
\end{enumerate}

\subsection{Feature extraction phase:}
Feature extraction is an essential step for transforming raw pixel data into a suitable feature set making it easier for the algorithm to analyze images. The proposed work extracts improved RCNN-based Ear detection, Ear attributes, and AAM-based features from the preprocessed image.

\subsubsection{Improved RCNN based Ear detection ($F_{IRCNN}$)}
\begin{figure}[!t]
    \centering
    \includegraphics[width=\linewidth]{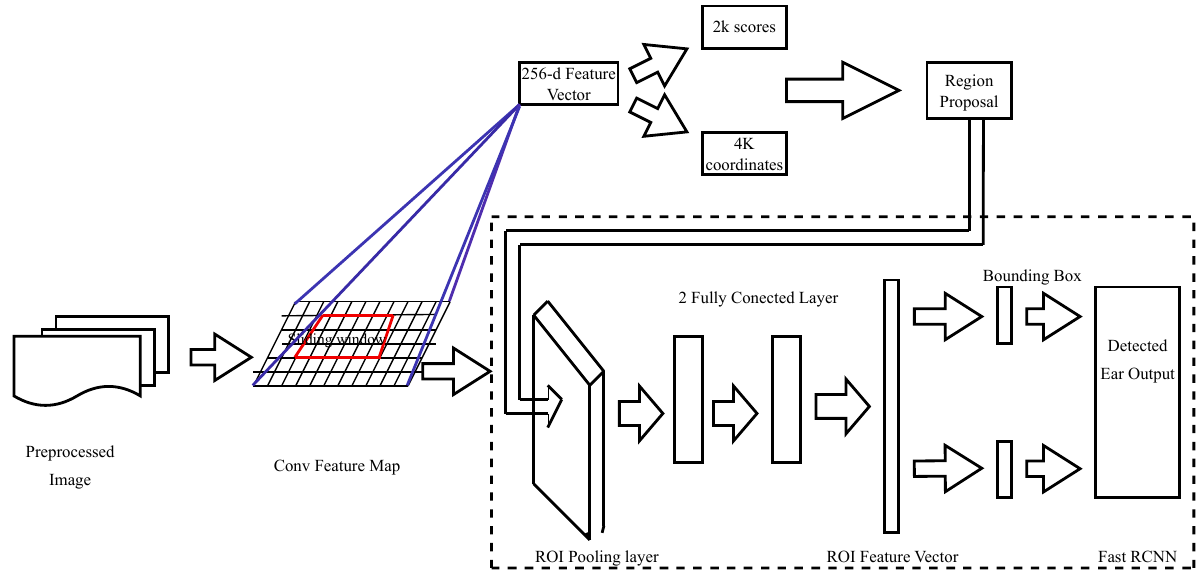}
    \caption{Conventional RCNN architecture.}
    \label{figure2}
\end{figure}

For detecting ears from the preprocessed image $I^{P}_{f}$, an RCNN algorithm \cite{el2018ear} is proposed. The procedural steps of the algorithm are described as follows :\\
\vspace{-2mm}
\begin{enumerate}
	\item[\textbf{1.}] After processing the input facial image with CNN, a convolutional feature map of that image is generated. 
	\vspace{1mm}
	\item[\textbf{2.}] An RPN is employed to process the feature map. Lower dimensions (256-d) are attained by employing a sliding window that traverses spatially across the feature map. In the proposed approach, a small window is moved through a convolutional feature map transforming it into a 512-d vector. Each anchor is created by this sliding window and the RPN's convolutional layer processes each anchor for further processing. If the object-based score of an anchor box exceeds a predefined threshold its coordinates are forwarded as a region proposal.
	\vspace{1mm}
	\item[\textbf{3.}] To obtain the most precise dimensions that accommodate the ear object, region suggestions undergo a series of steps including ROI pooling, a fully connected layer, a bounding box regressor, and a SoftMax classification layer. The regressor's output predicts the bounding box's dimensions (x,y, width, and height) while the classifier's final output which is represented as probability $p$ indicates the presence of the object of interest in the expected box. The traditional architecture of an RCNN is depicted in Fig. \ref{figure2}.
\end{enumerate}

While the traditional RCNN model relies on VGG as its structural foundation the proposed model employs the ResNet50 model as its backbone. In the RCNN structure, the ROI pooling layer is traditionally followed by two fully connected layers. However, in the proposed RCNN model five fully connected layers are utilized. Additionally, the bounding box regression employs a hybrid activation function known as the hyper sig activation function which is a combination of the hyperbolic sigmoid and bipolar sigmoid activation functions. Both functions belong to the class of hyperbolic tangent functions and return values within the interval [-1, 1]. The bipolar sigmoid $BS\left[ f(x) \right]$ and hyperbolic activation $HA\left[ f(x) \right]$ \cite{abubakar2014modified} functions are defined by Eq. \ref{eq2} and Eq. \ref{eq3}, respectively. 

\begin{equation}\label{eq2}
    BS\left[ f(x) \right]= -1+\frac{2}{1+e^{-x}}
\end{equation}
\begin{equation} \label{eq3}
    HA\left[ f(x) \right]= \frac{e^{x}-e^{-x}}{e^{x}+e^{-x}}
\end{equation}
These equations are then combined to form the hyper sig activation function as shown in Eq. \ref{eq4}. Eq. \ref{eq5} through Eq. \ref{eq9} outlines the derivation for the proposed activation function while Eq. \ref{eq10} defines the hyper sig activation function that arises as a result.


\begin{equation}\label{eq4}
    HS\left[ f(x) \right]= \left[ -1+\frac{2}{\left( 1+e^{-x} \right)} \right]+\left[ \frac{e^{x}-e^{-x}}{e^{x}+e^{-x}} \right]
\end{equation}

\begin{equation} \label{eq5}
    HS\left[ f(x) \right]= \left[ \frac{-1\left( 1+e^{-x}\right)+2}{1+e^{-x}} \right]+\left[ \frac{e^{x}-e^{-x}}{e^{x}+e^{-x}} \right]
\end{equation}
 \begin{equation} \label{eq6}
   = \left[ \frac{ -1-e^{-x}+2}{1+e^{-x}} \right]+\left[ \frac{e^{x}-e^{-x}}{e^{x}+e^{-x}} \right]
\end{equation}
\begin{equation}\label{eq7}
=\frac{\left( 1-e^{-x} \right)\left( e^{x}+e^{-x} \right)+\left( 1+e^{-x} \right)\left( e^{x}-e^{-x} \right)}{\left( 1+e^{-x} \right)\left( e^{x} +e^{-x}\right)}
\end{equation}
\begin{equation}\label{eq8}
  =  \frac{e^{x}+e^{-x}-e^{-x+x}-e^{-x-x}+e^{x}-e^{-x}+e^{-x+x}-e^{-x-x}}{e^{x}+e^{-x}+e^{-x+x}+e^{-x-x}}
\end{equation}

\begin{equation}\label{eq9}
  =  \frac{e^{x}+e^{-x}-e^{0}-e^{-2x}+e^{x}-e^{-x}+e^{0}-e^{-2x}}{e^{x}+e^{-x}+e^{0}+e^{-2x}}
\end{equation}
\begin{equation}\label{eq10}
=\frac{2e^{x}-2e^{-2x}}{e^{x}+e^{-x}+e^{-2x}+1}
\end{equation}

\begin{figure}[!ht]
    \centering
    \includegraphics[width=\linewidth]{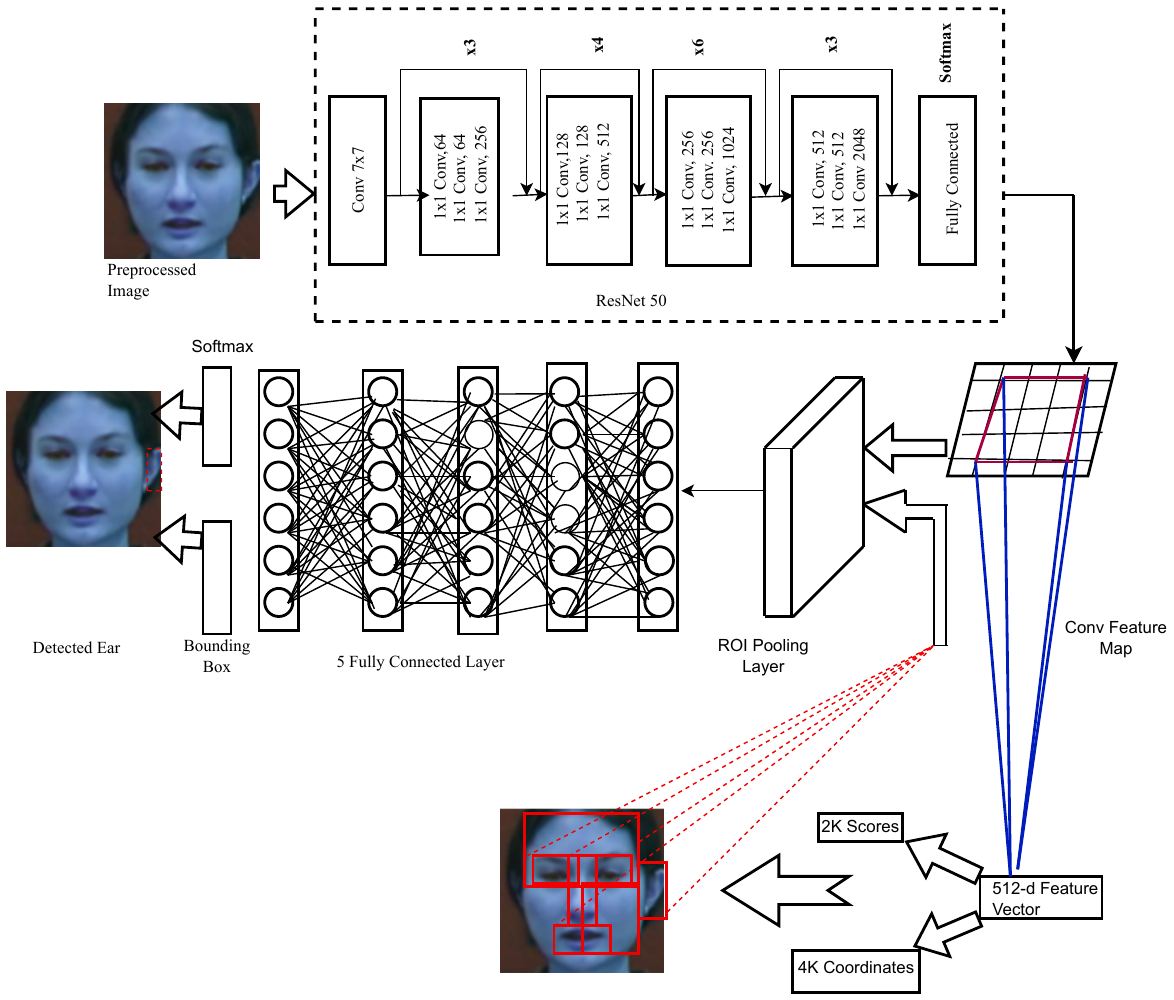}
    \caption{Improved RCNN architecture.}
    \label{figure3}
\end{figure}

Fig. \ref{figure3} illustrates the overall architecture of the improved RCNN method. The ear features detected from the preprocessed image through the improved RCNN are denoted as $F_{IRCNN}$.

\subsubsection{Ear Attributes ($F_{EA}$)}
Our work considers sizes and shapes as ear features to form the descriptor. It identifies specific ear locations that can serve as landmarks for size and shape feature extraction. The extracted feature of ear attributes is denoted as $F_{EA}$.

\begin{itemize}
    \item[\textbf{1.}] \textbf{Size:} Aspect ratio represented as a ratio of height to width and the ear's horizontal and vertical dimensions are the two parameters for describing the overall shape of the ear. An ellipse or circle is fitted to the ear region to obtain major and minor axes, center coordinates, and orientation parameters. This provides a concise representation of the ear contour. The distance between designated landmark sites is measured to determine the precise dimensions and ratio of various ear components.

    \item[\textbf{2.}] \textbf{Shape:} Analyzing the curvature of specific points or curves on the ear provides insights into its overall shape. Analyzing the geometric properties of the ear is typically necessary to extract shape details from ear attributes.  
\end{itemize}
\subsubsection{AAM features ($F_{AAM}$)}
Ear detection utilizes the statistical shape and appearance model known as AAM \cite{iqtait2018feature} to facilitate ear object tracking and detection. AAM records the variability in ear appearances and shapes within the specific dataset in connection with ear detection. The appearance model represents the variance in pixel value or texture within an object's shape. These variances are modeled using Principal Component Analysis (PCA) or other statistical techniques. Therefore, the extracted AAM-based features from the preprocessed image are denoted as $F_{AAM}$.

Thus, the total extracted features from the preprocessed image are termed as $F_{f}=\left[ F_{IRCNN}, F_{EA}, F_{AAM} \right]$ which are subjected to the detection process.

\subsection{Detection of DeepFake with Optimized Hybrid Model}
Following the process of feature extraction, the feature set ($F_{f}$) is given as input to the hybrid model for deepfake detection. The hybrid model is capable for reducing the computation time while enhancing the detection accuracy. Subsequently, an improved score-level fusion approach is utilized for deepfake detection tasks. Additionally, the Self-Upgraded Jellyfish Optimization (SU-JFO) algorithm optimizes the weight of both the models i.e DBN and Bi-GRU to improve deepfake detection performance.

\subsubsection {Bidirectional Gated Recurrent Unit (Bi-GRU)} 
GRU unit consists of a reset gate $R_{g}\left( t \right)$ and an update gate $Z_{g}\left( t \right)$ computed using Eq. \ref{eq11} and Eq. \ref{eq12}, respectively. These two gates govern the present input $F_{f}$ and the past state $H\left( t-1 \right)$ which determines the output $H\left( t \right)$. The outputs of the gate are computed using Eq.\ref{eq13} and Eq. \ref{eq14}, where $W_{R}^{'}$, $W_{Z}^{'}$, $W_{H}^{'}$, $V_{R}^{'}$, $V_{Z}^{'}$, $V_{H}^{'}$ represent the weight matrices of the reset gate, update gate, and output respectively. $G_{R}$, $G_{Z}$, $G_{H}$ represents the synthesis of bias vectors for input $F_{f}\left( t \right)$ and the previous state $H\left( t-1 \right)$, and the logistic sigmoid function and the hyperbolic tangent activation function are denoted as $\sigma$ and $tanh$, respectively.

\begin{equation}\label{eq11}
  R_{g}\left( t \right)=\sigma(W_{R}^{'}F_{f} (t)+ V_{R}^{'}H\left( t-1 \right)+ G_{R})
\end{equation}

\begin{equation}\label{eq12}
    Z_{g}\left( t \right)=\sigma(W_{Z}^{'}F_{f} (t)+ V_{Z}^{'}H\left( t-1 \right)+ G_{Z})
\end{equation}
\begin{equation}\label{eq13}
  H^{'}\left(t \right)=tanh\left( W_{H}^{'}F_{f}\left( t \right)+V_{H}^{'}\left( R_{g}\left( t \right)\odot H\left( t-1 \right) \right)+ G_{H} \right)
\end{equation}
\begin{equation}\label{eq14}
    H\left( t \right)=\left( 1-Z_{g}(t) \right)\odot H\left( t-1 \right)+Z_{g}\left( t \right)\odot H^{'}\left( t \right)
\end{equation}

\begin{figure}[!t]
    \centering
    \includegraphics[width=\linewidth]{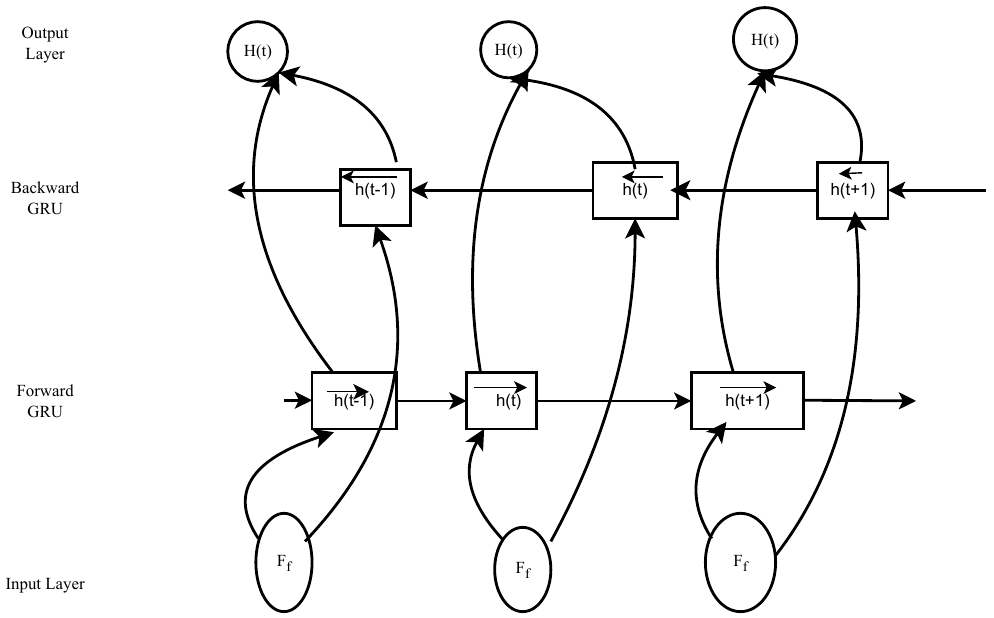}
    \caption{Architecture of Bi-GRU model.}
    \label{figure4}
\end{figure}

Models featuring a bidirectional structure can leverage the past and future data while processing the current data. Fig. \ref{figure4} displays the structure of the Bi-GRU model \cite{liu2021bi}. The Bi-GRU model is constructed by analyzing the states of two unidirectional GRUs pointing in opposing directions. One GRU starts at the beginning of the data series and moves forward while another GRU starts at the end and moves backward. This allows knowledge from both the past and the future to influence the current state. Eq. \ref{eq15} and Eq. \ref{eq16} compute the forward and backward GRU outputs, and Eq. \ref{eq17} indicates the Bi-GRU model's final output denoted as $H\left( t \right).    \overrightarrow{H\left( t\right)}$ denotes the state of forward GRU, $\overleftarrow{H\left( t\right)}$ represents the state of backward GRU, and $\oplus$ represents the operation of concatenating two vectors. According to this work, the SU-JFO algorithm optimally tunes the weight of the Bi-GRU model represented as $W'_{1}$.
\begin{equation}\label{eq15}
    \overrightarrow{H\left( t \right)}=Fwd.GRU\left( F_{f}\left( t\right),\overrightarrow{H\left( t-1 \right)} \right)
\end{equation}
\begin{equation}\label{eq16}
    \overleftarrow{H\left( t \right)}=Bwd.GRU\left( F_{f}\left( t\right),\overleftarrow{H\left( t-1 \right)} \right)
\end{equation}
\begin{equation}\label{eq17}
    H\left( t\right)=\overrightarrow{H\left( t \right)}\oplus\overleftarrow{H\left( t\right)}
\end{equation}

\subsubsection{Deep Belief Network (DBN)}
\begin{figure}[!t]
    \centering
    \includegraphics[width=\linewidth]{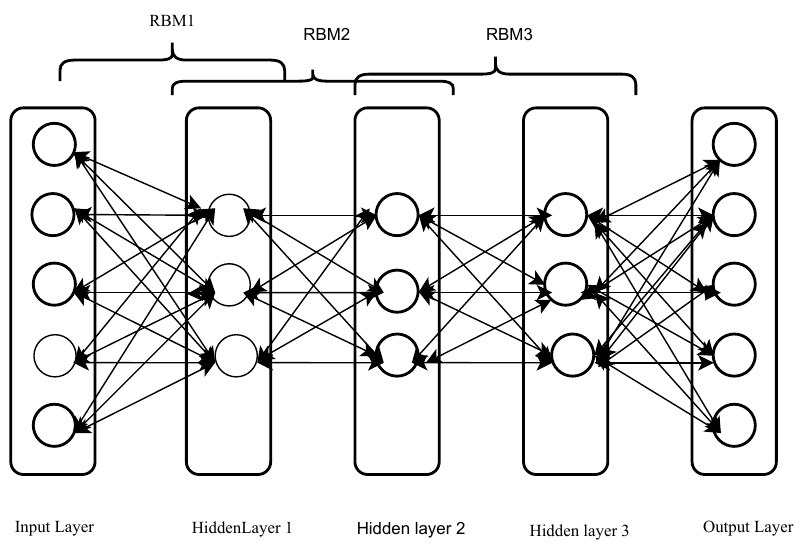}
    \caption{Architecture of DBN model.}
    \label{figure5}
\end{figure}

A generative graphical model consisting of stacked RBMs (Restricted Boltzmann Machine) is DBN \cite{koo2020improved}. Due to its deeper structure, DBN can extract a hierarchical arrangement of the input. DBN utilizes a training method that trains each layer greedily one by one. Eq. \ref{eq18} computes the joint distribution given the visible input unit and hidden layers. $P\left( h^{k-1},h^{k} \right)$ represents the conditional distribution of the top RBMs visible and hidden layers. Fig. \ref{figure5} illustrates the architecture of the DBN model. 
\begin{equation}\label{eq18}
    P\left( x,h^{1},...,h^{k} \right)=\left( \prod_{l=0}^{k-2}P\left( h^{l}|h^{l+1} \right) \right)P\left(h^{k-1},h^{k}  \right)P\left( x|h^{l} \right)
\end{equation}

In the DBN each layer is constructed similar to an RBM making the training process similar to training an RBM layer. To initiate network training, classification is employed in the DBN. Pre-training involves layer-wise unsupervised learning of stacked RBMs and fine-tuning employing supervised learning using a classification algorithm which can be sequentially used to achieve two-phase training. At each stage, an optimization problem must be addressed. During the pre-training phase of each layer l, the optimization problem is tackled using a given feature set: 
$F_{f}=\left\{ \left( x^{(1)},y^{(1)},...,\left( x^{(\left| F_{f} \right|)},y^{\left| \left(  F_{f}\right) \right|} \right) \right) \right\}$ with inputs $x$ and labels $y$. The weight of the DBN model is optimally tuned using the SU-JFO algorithm represented as $W'_{2}$. The output of the DBN model is termed as $DBN_{out}$.
\subsection{Optimal weight tuned by proposed SU-JFO algorithm}

\subsubsection{Solution Encoding}
The weight of the hybrid detection model is considered as $W^{'}_{i}=\left[ W^{'}_{1}, W^{'}_{2}\right]$. In this context, the weights of the DBN model are denoted by $W^{'}_{2}$ while those of the Bi-GRU model are represented by $W^{'}_{1}$. In the proposed model, the population size is fixed at 10, and the values of the upper bound and lower bound of the solution are fixed at 1 and 0 respectively.
\subsubsection{Objective Function}
The proposed model's objective is defined as $O_{F}$. It represents the minimization of the error between the actual and predicted values as per Eq. \ref{eq19}.
\begin{equation}\label{eq19}
    O_{F}=Min(Error)
\end{equation}
\subsubsection{ Model of SU-JFO algorithm}
In the proposed work, both the weights of the detection model are optimally tuned using the SU-JFO algorithm. The Jellyfish Search(JS) optimizer \cite{chou2021novel} is a unique meta-heuristic algorithm inspired by the movements of a jellyfish in the water. Three fundamental principles form the basis of our proposed optimization method given as follows:
\begin{itemize}
\item[\textbf{1.}] \textbf{Time Control Mechanism:} Controls the switching between two types of jellyfish movements following the ocean current or moving within the swarm.
\item[\textbf{2.}] \textbf{Food Attraction:} Jellyfish migrate in search of food in water gravitating towards the locations with higher nourishment availability.
\item[\textbf{3.}] \textbf{Location and Objective Function:} The amount of food discovered by Jellyfish is defined by their location and related objective function.  
\end{itemize}
\textbf{Step 1. Update the position of Jellyfish in Ocean Current:}
The ocean current's high nutrition content attracts jellyfish. The direction of the ocean current $(\overrightarrow{trend})$ is determined by averaging all vectors from every jellyfish to the jellyfish currently in the optimal position. Ocean currents are simulated by Eq. \ref{eq20}. As a result, $\overrightarrow{trend}$ is established using Eq. \ref{eq22}.
\begin{equation} \label{eq20}
	\overrightarrow{trend}=\frac{1}{N}\sum_{}^{}\overrightarrow{trend_{i}}=\frac{1}{N}\sum_{}^{}(W^{''}-a_{c}W'_{i})=W^{''}-a_{c}\sum_{}^{}\frac{W'_{i}}{N}=W^{''}-a_{c}\mu
\end{equation}

\begin{equation}\label{eq21}
\text{Set} \quad df = a_{c}\mu
\end{equation}

\begin{equation}\label{eq22}
	\overrightarrow{trend}=W^{''}-df
\end{equation}
where N represents the number of jellyfish, $W^{''}$ represents the jellyfish currently at the optimal position, $a_{c}$ represents the factor governing attraction, $\mu$ indicates the mean location of all jellyfish, $df$ represents the difference between the current best location of a jellyfish and the mean location of all jellyfish, and $\pm \beta\sigma$ denotes the distance around the mean location containing some probability of all jellyfish calculated using Eq. \ref{eq21} and Eq. \ref{eq23}, where sigma $\sigma$ represents the standard deviation as per Eq. \ref{eq24}.

\begin{equation} \label{eq23}
	df=\beta\times \sigma\times R^{f}\left( 0,1 \right)
\end{equation}
\begin{equation} \label{eq24}
	\text{Set} \quad \sigma =  R^{\alpha}\left( 0,1 \right)\times \mu  
\end{equation}
Hence, \begin{equation} \label{eq25}
	df = \beta\times R^{f}\left( 0,1 \right)\times  R^{\alpha}\left( 0,1 \right)\times \mu 
\end{equation}

Upon simplification, we denote the Eq. \ref{eq25} as Eq. \ref{eq26} where $\alpha_{c}= \beta\times R\left( 0,1 \right)$
\begin{equation} \label{eq26}
	df = \beta\times R\left( 0,1 \right)\times \mu
\end{equation}
Therefore, 
\begin{equation} \label{eq27}
	\overrightarrow{trend} = W^{''}-\beta\times R\left( 0,1 \right)\times \mu
\end{equation}
Eq. \ref{eq28} shows the updated random position in Eq. \ref{eq26} by using Eq. \ref{eq27}.
\begin{equation} \label{eq28}
	R=R(t)\times i_{min}+\frac{\left( i_{max}-i_{min}\right)*t}{T_{max}}
\end{equation}
where, $t$ signifies the current iteration, ${T_{max}}$ signifies the maximum iteration, ${i_{max}}$ denotes the maximum inertia value equal to 0.9, ${i_{min}}$ denotes the minimum inertia value equal to 0.9 and the value of $R(0)$ is 0.7. Eq. \ref{eq29} gives the new location of each jellyfish as follows:
\begin{equation} \label{eq29}
	W^{'}_{i}(t+1)= W^{'}_{i}(t)+R(0,1)\times \overrightarrow{trend}
\end{equation}

Substituting all values in Eq. \ref{eq29} and it may be written as per Eq. \ref{eq30}, where the distribution coefficient is based on the condition $\beta>0$ which is associated with the length of $\overrightarrow{trend}$.  $\beta=3$ is obtained based on the sensitivity analysis outcome in numerical experiments.
\begin{equation} \label{eq30}
	W^{'}_{i}(t+1) = W^{'}_{i}(t)+R(0,1)\times (W^{''}-\beta\times R(0,1)\times \mu)
\end{equation}
\textbf{Step 2. Updated the position of jellyfish in swarm:} During this phase, jellyfish within the swarm exhibit two distinct movement patterns: passive and active. Initially most jellyfish engage in Type A motion gradually transitioning to Type B motion. In Type A motion jellyfish move within their respective positions. The updated location of each jellyfish is determined in Eq. \ref{eq31}, where UB and LB represent the upper and lower bounds respectively. $\gamma$ denotes the motion coefficient associated with the dimension of motion around the jellyfish's position with a sensitivity analysis yielding a value of 0.1. 
\begin{equation} \label{eq31}
	W^{'}_{i}(t+1)=W^{'}_{i}(t)+\gamma\times R(0,1)\times \left( UB-LB \right)
\end{equation}
Similarly, the type B movement of the jellyfish is considered as the exploitation phase of the local search area. In the process described, a random jellyfish (denoted as $j$) distinct from the central jellyfish of interest $(i)$ is chosen. A vector is then formed from the jellyfish of interest $(i)$ to the selected jellyfish $(j)$ determining the direction of motion as per Eq. \ref{eq32}. Each jellyfish adjusts its movement direction to optimize its search for food within the swarm as shown in Eq. \ref{eq33} and Eq. \ref{eq34}, where $O_{F}$ denotes the objective function of location $W$. 

\begin{equation} \label{eq32}
	\overrightarrow{Step}=W^{'}_{i}(t+1)-W^{'}_{i}(t)
\end{equation}
where, $\overrightarrow{Step}=R\left( 0,1 \right)\times \overrightarrow{Direction}$
\begin{equation} \label{eq33}
	{\overrightarrow{Direction}} = \begin{cases} 
		W_j'(t) - W_i'(t) & \text{if } O_F(W_i') \geq O_F(W_j') \\ 
		W_i'(t) - W_j'(t) & \text{if } O_F(W_i') < O_F(W_j')
	\end{cases}
\end{equation} 

\begin{equation} \label{eq34}
	W^{'}_{i}(t+1)=W^{'}_{i}(t)+\overrightarrow{Step}
\end{equation}

For updating the exploitation phase, Eq. \ref{eq34} is utilized to obtain the new position in the swarm as per Eq. \ref{eq35}, where $\overrightarrow{R_{n}}$ and $\overrightarrow{R_{n1}}$ are random numbers in the range of 0 and 1, and the distribution coefficient is in the range of 0.5 to 10. In this proposed work, the value of the distribution coefficient denoted as $d$ is taken as 3.

\begin{equation}  \label{eq35}
	W^{'}_{i}(t+1)=W^{'}_{i}(t)+\overrightarrow{Step}+\overrightarrow{R_{n}}*\left(\bar{W}'' - \beta*R_{n1}*\mu\right)
\end{equation}

Next, a time control mechanism is employed to calculate the motion type over time. This mechanism controls all jellyfish movement around the ocean region and within the jelly swarm. The following steps provide a detailed explanation of the time control mechanism.\\ 

\textbf{Step 3. Time control mechanism:} In this time control mechanism, jellyfish movement in the current region of ocean containing a large quantity of food is controlled. When climatic conditions change jellyfish move to another ocean current area and form a swarm. Initially, Type A jellyfish are favored and as the time progresses Type B jellyfish are preferred. The movement of jellyfish within the jelly swarm and their interaction with the ocean current are regulated using a time control approach. The time control function denoted by the symbol $c(t)$ is the main component and its value ranges from 0 to 1. Eq. \ref{eq36} is employed to calculate the time control function. Additionally, algorithm \ref{alg:example} displays the pseudocode of the time control mechanism.

\begin{equation} \label{eq36}
	c(t) = \left| \left( 1 - \frac{t}{Max_{iter}} \right) \times \left( 2 \times R(0,1) - 1 \right) \right|
\end{equation}

\begin{figure}[!htbp]
	\centering
	\includegraphics[width=\linewidth]{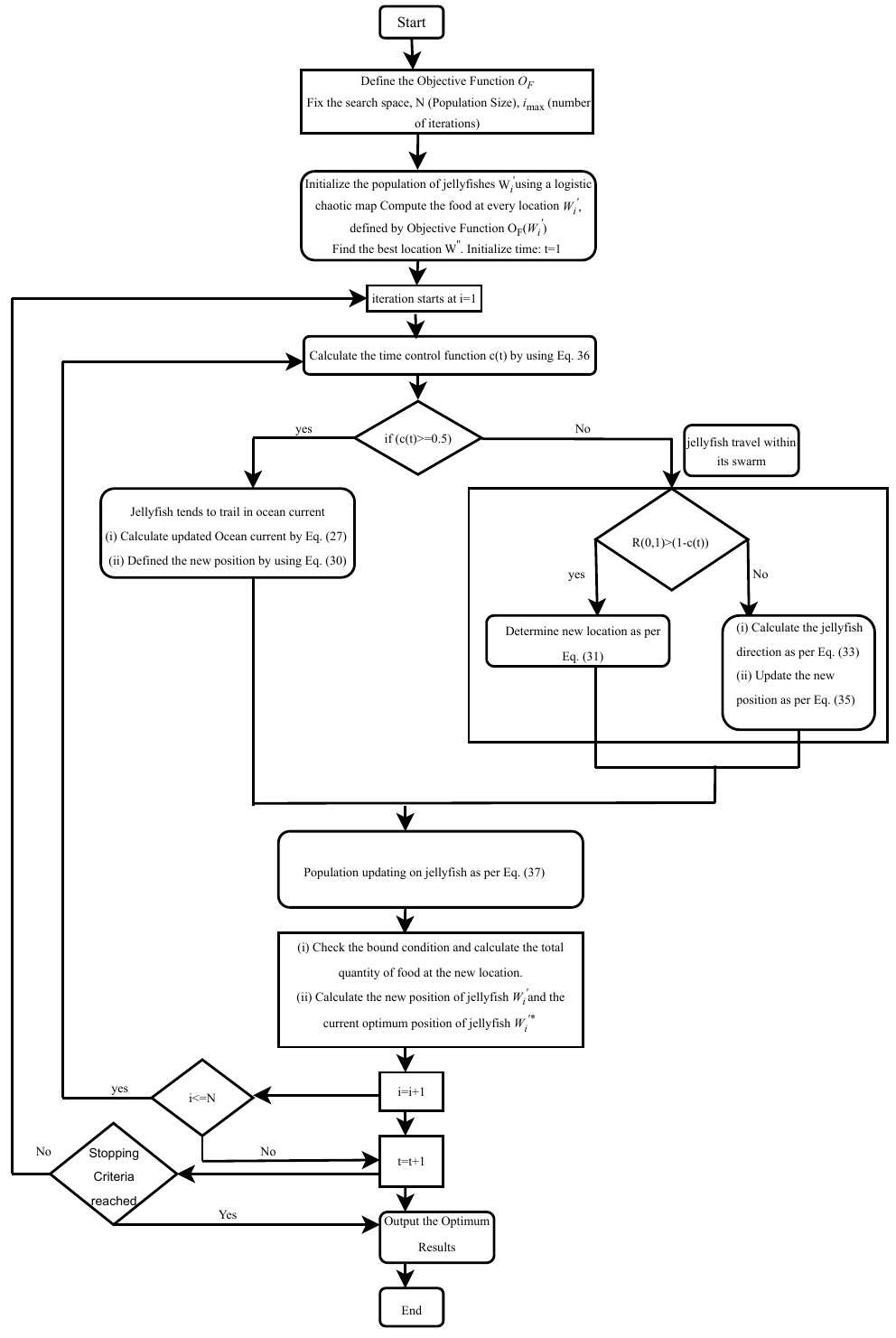}
	\caption{Flowchart of the SU-JFO algorithm.}
	\label{figure7}
\end{figure}

\begin{algorithm}[H] 
	\caption{Time control mechanism} \label{alg:example}
	
	\begin{algorithmic}
		\STATE  Start
		\STATE \hspace{3mm} For each i=1:N (population) do \\
		\STATE \hspace{10mm}Evaluate the time control c(t)  as per Eq. \ref{eq35}.\\
		\STATE \hspace{10mm}If c(t)>=0.5 \\
		\hspace{15mm} Jellyfish tends to trails ocean current.\\
		\hspace{10mm}Else: Within the swarm, jellyfish travel.\\
		\STATE \hspace{15mm} If R(0,1)>(1-c(t)):\\
		\hspace{20mm} Jellyfish move passively in a Type A manner.\\
		\hspace{15mm}Else: \\
		\hspace{20mm} Jellyfish move actively in a Type B manner.\\ 
		\STATE    \hspace{15mm} End if\\
		\STATE \hspace{10mm} End if \\
		\STATE  \hspace{3mm} End for \\
		\STATE End
	\end{algorithmic}
\end{algorithm}
\textbf{Step 4. Population updating on jellyfish:} In the stage of population update, the new updated positions of jellyfish are determined based on their population. Eq. \ref{eq37} provides the updated jellyfish population \cite{abdel2021improved} where $R1$, $R2$, and $R3$ are the indices of three solutions randomly selected from the population and denote the control parameter which is a random number between 0 and 1 used to control movement of the current solution. Figure \ref{figure7} depicts the flowchart of the SU-JFO algorithm. 
\begin{equation} \label{eq37}
	W^{'}_{i}(t+1)=W^{'}_{i}(t)+R*\left( \overrightarrow{W^{'}_{R1}}(t)-\overrightarrow{W^{'}_{R2}}(t)\right)+(1-R)*\left( W^{''}-\overrightarrow{W^{'}_{R3}}(t)  \right)
\end{equation}
Finally, the jellyfish travels around the boundary search area and it returns to the opposite boundary area. The re-entering procedure is calculated as per Eq. \ref{eq38}, where $W^{'}_{i,d}$ signifies the position of $i^{th}$ jellyfish in dimension $d$. The updated position after validating the constraints of boundary is denoted as $W^{'*}_{i,d}$, $UB_{d}$ and $LB_{d}$ signifies the upper boundary and lower boundary in the $d^{th}$ dimension respectively. 
\begin{equation} \label{eq38}
	W_{i,d}^{\prime *} = 
	\begin{cases} 
		(W_{i,d}' - UB_d) + LB(d) & \text{if } W_{i,d}' > UB_d \\ 
		(W_{i,d}' - LB_d) + UB(d) & \text{if } W_{i,d}' < LB_d 
	\end{cases}
\end{equation}

   \subsubsection{Improved Score level Fusion}
   Finally, the detection outcome is determined by the improved score-level fusion techniques \cite{hamd2020score} based on the output of the Bi-GRU and DBN models. The outcome is in the form of either 0 or 1  indicating whether the given data is real or fake, respectively. An improved normalization process is conducted for both the Bi-GRU and DBN models to bring both model outputs into a common range of the hybrid model as the output scores of Bi-GRU and DBN might have different scales. Eq. \ref{eq39} shows the improved normalization process of the Bi-GRU model denoted as $SC_{Bi-GRU_{out}}$, where H(t) represents the predicted output of Bi-GRU, $DBN_{out}$ denotes the predicted output of DBN, and $LF_{Bi-GRU}$ denotes the local factor of Bi-GRU calculated as per Eq. \ref{eq40}. Similarly, the improved normalization of the DBN model $SC_{DBN_{out}}$ is calculated as per Eq. \ref{eq41}, and the local factor of DBN ($LF_{DBN}$) is expressed in Eq. \ref{eq42}. The local factor in Eq. \ref{eq40} and Eq. \ref{eq42} depicts how much the current output deviates from the central tendency of the Bi-GRU's and DBN predictions respectively. It ensures that the predictions are scaled appropriately based on the distributions of outputs which adds robustness thereby allowing the model to focus on meaningful predictions. Next, the normalization scores are fused and the improved score level $FS$ is computed using Eq. \ref{eq43} where $C_{v}$ indicates the coefficient of the variable calculated in Eq. \ref{eq44}. Additionally, the terms $E^{-}_{i}$ and $E^{+}_{i}$ can be expressed as per Eq. \ref{eq45} and Eq. \ref{eq46} respectively. The term $E^{+}_{i}$ and $E^{-}_{i}$ accounts for false positive and false negative respectively. The target label is denoted as $T_{ij}$ which represents the actual class label for a given data point $i$ indicating whether the data is real or fake. $min\left( H(t) \right)$ indicates the minimum value of the Bi-GRU output and $max\left( H(t) \right)$ indicates the maximum value of the Bi-GRU output, and $median\left( H(t) \right)$ designates the median value of the Bi-GRU output. $min\left( DBN_{out} \right)$ and $max\left( DBN_{out} \right)$ denote the minimum and maximum values of the DBN output, respectively while $median\left( DBN_{out} \right)$  denotes the median value of the DBN output. After normalization and score-level fusion computation, a combined score is generated. The final output is determined by a threshold  of (0.5). The fused score is further compared with the threshold value to determine whether the output is fake or real.
     \begin{equation} \label{eq39}
         SC_{Bi-GRU_{out}}=\frac{H(t)-min\left( H(t) \right)}{max\left( H(t) \right)-min\left( H(t) \right)*LF_{Bi-GRU}}
     \end{equation}
     \begin{equation} \label{eq40}
        LF_{Bi-GRU}=\left[ \frac{H(t)-median\left( H(t) \right)}{mean( H(t)-median\left( H(t) )\right)} \right]
     \end{equation}
     \begin{equation} \label{eq41}
          SC_{DBN_{out}}=\frac{DBN_{out}-min\left(DBN_{out} \right)}{max\left( DBN_{out} \right)-min\left( DBN_{out} \right)*LF_{DBN}}
     \end{equation}
     \begin{equation} \label{eq42}
          LF_{DBN}=\left[ \frac{DBN_{out}-median\left( DBN_{out} \right)}{mean(DBN_{out} -median\left( DBN_{out} \right))} \right]
     \end{equation}
     \begin{equation} \label{eq43}
FS=\left[ C_{v}*SC_{Bi-GRU_{out}} \right]+\left[ (1-C_{v})*SC_{DBN_{out}} \right]
     \end{equation}
     \begin{equation} \label{eq44}
         C_{v}=\frac{E^{-}_{i}}{E^{+}_{i}+E^{-}_{i}}
     \end{equation}
     \begin{equation} \label{eq45}
         E_i^- = \sqrt{\sum \left[\max T_{ij} - T_{ij} \right]^2}
     \end{equation}
     \begin{equation} \label{eq46}
         E_i^+ = \sqrt{\sum \left[\min T_{ij} - T_{ij} \right]^2}
     \end{equation}
     
     \section{\textbf{Experimental Results and Discussion}} \label{section4}
     This section describes the experimentation followed by performance evaluations carried out on three datasets i.e. WLDR Dataset (Dataset1) \cite{agarwal2019protecting}, DeepfakeTIMIT Dataset (Dataset2) \cite{deepfaketimit}, and Celeb-DF Dataset (Dataset3) \cite{yuezun2019celeb}. The description of datasets is described in Subsection \ref{subsection4.1}. 
     \subsection{Datasets}\label{subsection4.1}
      The deepfake detection model has been evaluated on three datasets: The World Leader Dataset has been referred to as dataset1 \cite{agarwal2019protecting}, DeepfakeTIMIT as dataset2 \cite{deepfaketimit} and Celeb-DF Dataset as dataset3 \cite{yuezun2019celeb} throughout the paper. The description of these datasets are provided in subsubsections \ref{subsection4.1.1}, \ref{subsection4.1.2}, \ref{subsection4.1.3} respectively.

     \subsubsection{World Leader Dataset (Dataset1)}\label{subsection4.1.1}
     The World Leader Dataset (WLDR) (dataset1) \cite{agarwal2019protecting} comprises an extensive footage of genuine videos showcasing five U.S. political figures, their political impersonators, and deepfake videos created using face-swapping technology. The deepfake videos portray each political figure alongside their respective impersonator.

     \subsubsection{VidTIMIT Dataset (Dataset2)}\label{subsection4.1.2}
     In the deepfakeTIMIT dataset (dataset2) \cite{deepfaketimit}, the selection process includes identifying 16 pairs of individuals who resemble each other from the VidTIMIT database available publicly. For each of the 32 subjects, two different models were trained a lower quality model and a higher quality model. A lower quality model is labeled as ``$LQ$" with a ``$64\times64$ input/output size", and a higher quality model is labeled as ``$HQ$" with a ``$128\times128$ input/output size . Within this dataset, the selection process involved manually identifying 16 pairs of individuals with a remarkable resemblance to each other from the publicly accessible VidTIMIT database. In the VidTIMIT database with 10 videos available per person, 320 videos were generated for each model version resulting in 620 videos showcasing swapped faces. Concerning audio, the original track from each video has not been subjected to any modification and hence remained unaltered. 
     \subsubsection{Celeb-DF Dataset (Dataset3)} \label{subsection4.1.3}
      Celeb-DF dataset (dataset3) \cite{yuezun2019celeb} is a deepfake forensics dataset collected from youtube. The dataset comprises 5639 deepfake videos  collected from celebrities, people of different ages, and genders. 
      
     \subsection{Performance Analysis}\label{subsection4.2}
     We conducted a thorough analysis to analyze our SU-JFO with traditional strategies. This assessment encompassed a comprehensive investigation of multiple performance indicators such as Accuracy, F-measure, Precision, MCC, and ROC. Furthermore, the SU-JFO technique was compared with state-of-the-art such as DFP \cite{raza2022novel} and ResNext+CNN+LSTM \cite{vamsi2022deepfake} as well as established techniques such as CNN, SqueezeNet, LeNet, LSTM, and LinkNet. Within this framework, an assessment was performed on both the SU-JFO method and conventional approaches utilizing  dataset1 \cite{agarwal2019protecting}, dataset2 \cite{deepfaketimit} and  dataset3 \cite{yuezun2019celeb}. 
    
\begin{figure}
	\vspace{-15mm}
	\centering
	
	
	
	\textbf{Enhanced RCNN-Ear Detection Based Images} 
	
	\subfloat[][]
	 {\includegraphics[width=.3\textwidth,height=3cm]{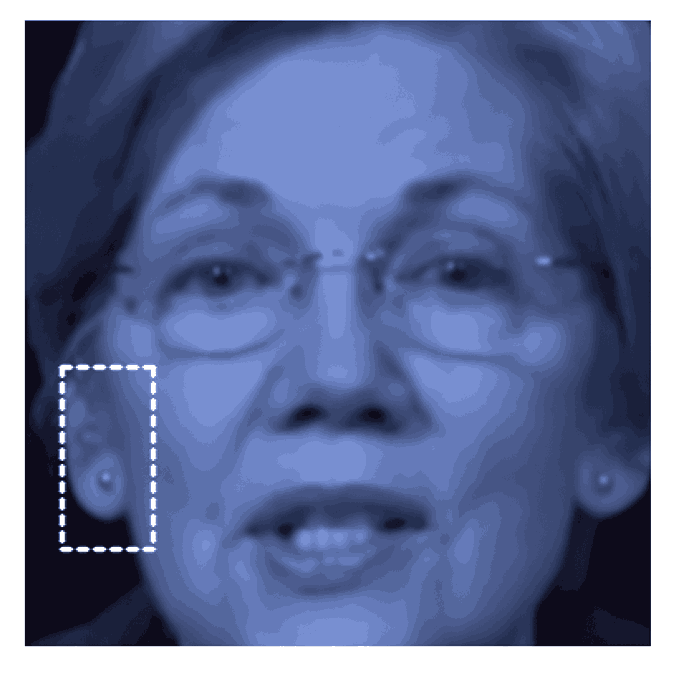}} \quad
	\subfloat[][]
	{\includegraphics[width=.3\textwidth,height=3cm]{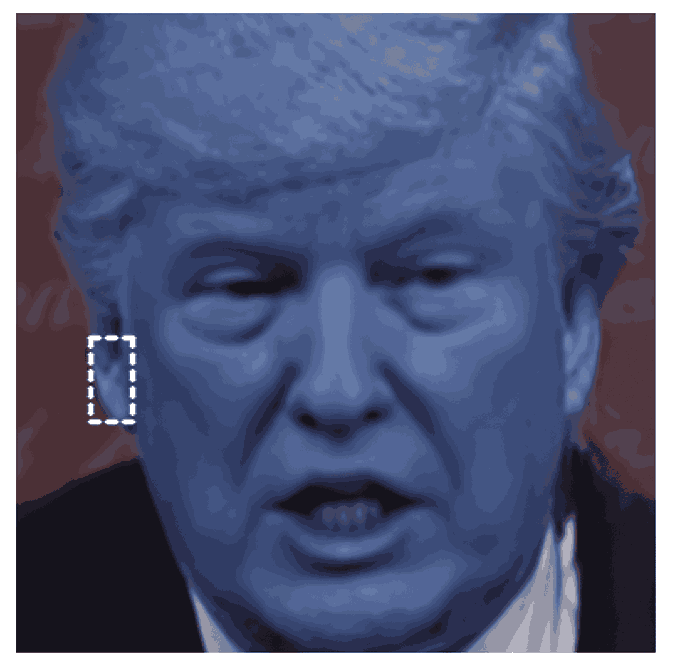}} \quad
	\subfloat[][]
	{\includegraphics[width=.3\textwidth,height=3cm]{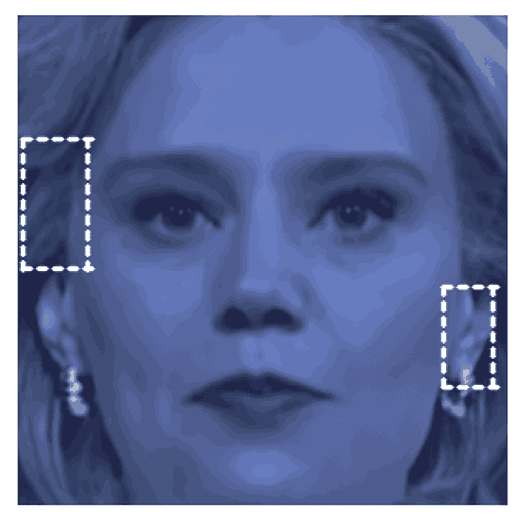}} \\
	\noindent 
\textbf{Compression Case Images} 
	\par 
	
		\subfloat[][]
	{\includegraphics[width=.3\textwidth,height=3cm]{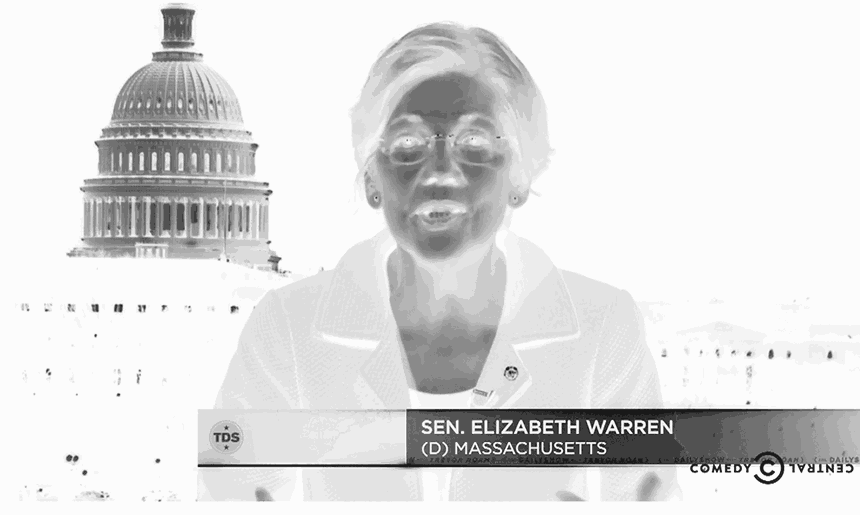}} \quad
	\subfloat[][]
	{\includegraphics[width=.3\textwidth,height=3cm]{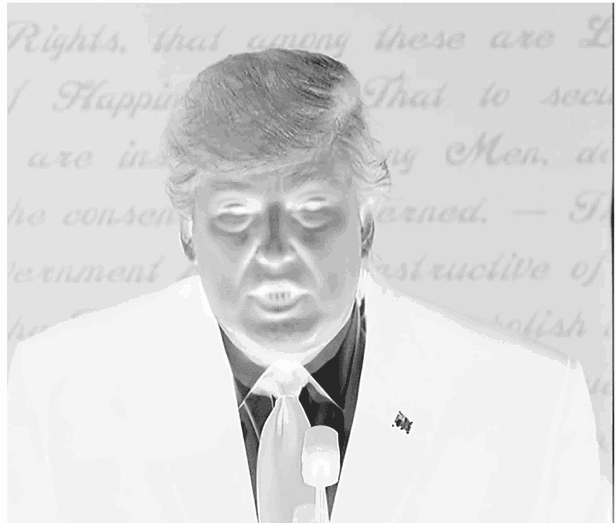}} \quad
	\subfloat[][]
	{\includegraphics[width=.3\textwidth,height=3cm]{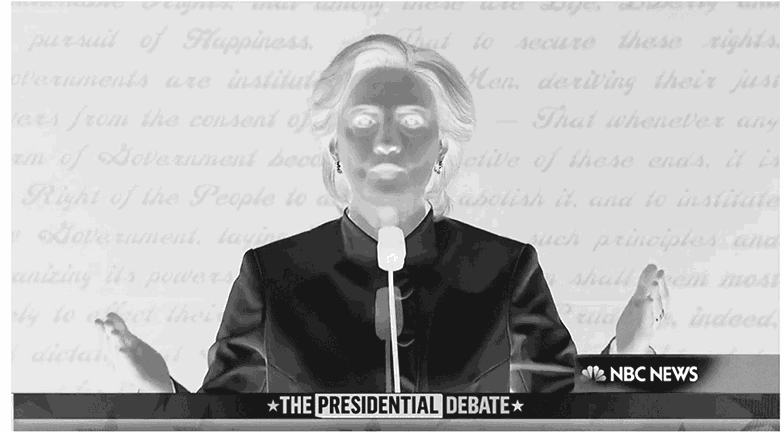}} \\
		\noindent 
\textbf{Noise Case Images} 
	\par 
		\subfloat[][]
	{\includegraphics[width=.3\textwidth,height=3cm]{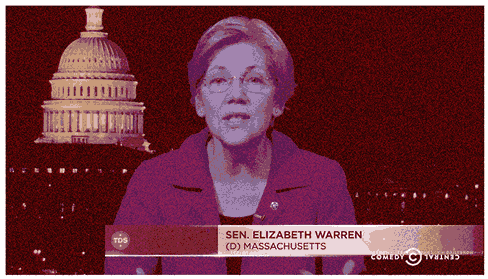}} \quad
	\subfloat[][]
	{\includegraphics[width=.3\textwidth,height=3cm]{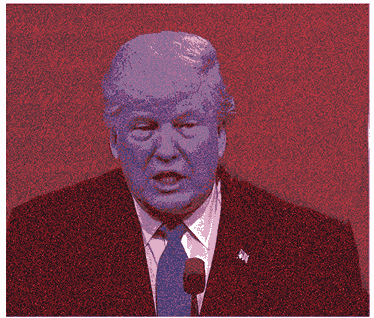}} \quad
	\subfloat[][]
	{\includegraphics[width=.3\textwidth,height=3cm]{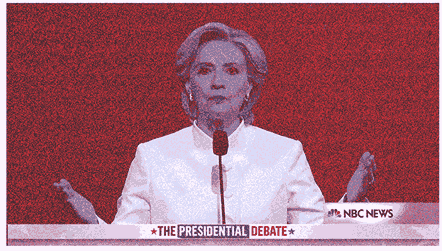}} \\
	\noindent 
	\textbf{Pose Illumination Case Images} 
	\par 
		\subfloat[][]
	{\includegraphics[width=.3\textwidth,height=3cm]{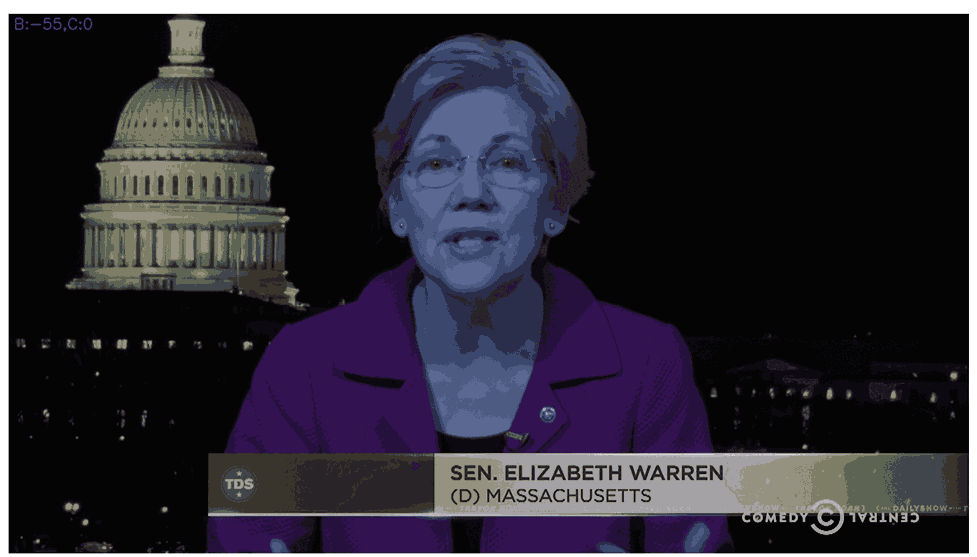}} \quad
	\subfloat[][]
	{\includegraphics[width=.3\textwidth,height=3cm]{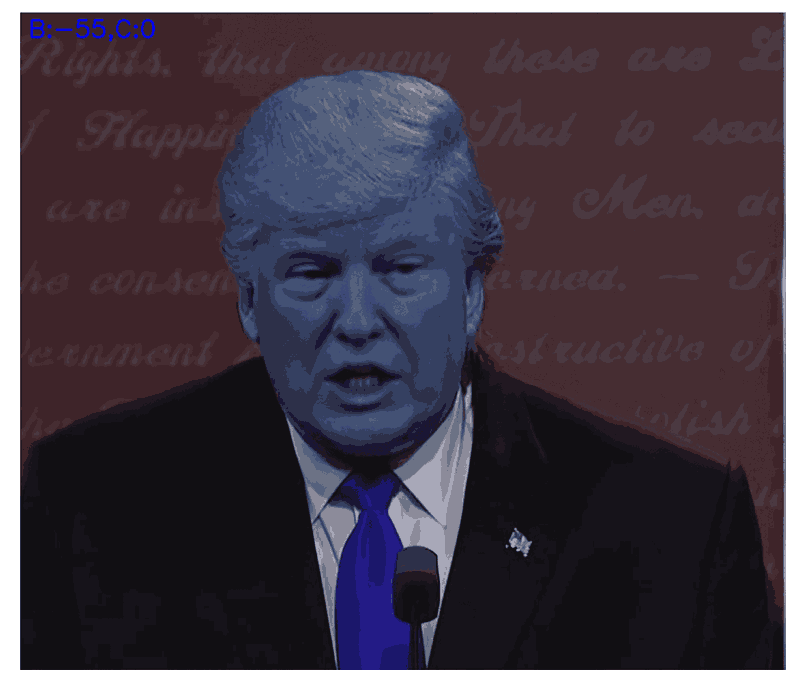}} \quad
	\subfloat[][]
	{\includegraphics[width=.3\textwidth,height=3cm]{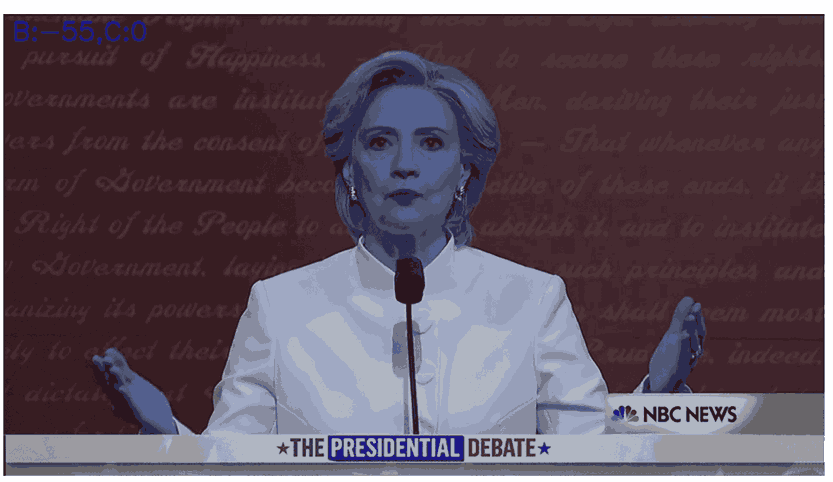}} \\
	\noindent 
	\textbf{Rotation Case Images} 
	\par 
		\subfloat[][]
	{\includegraphics[width=.3\textwidth,height=3cm]{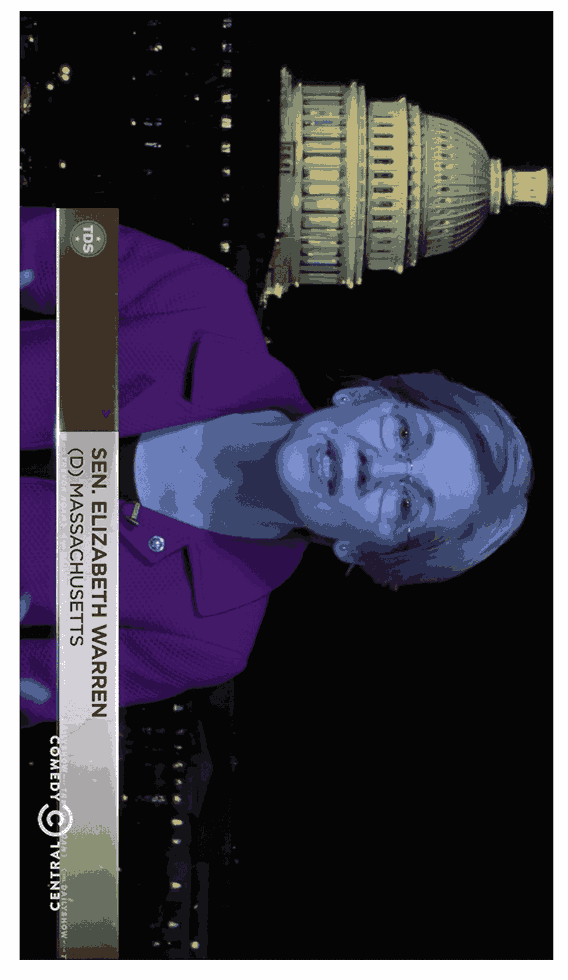}} \quad
	\subfloat[][]
	{\includegraphics[width=.3\textwidth,height=3cm]{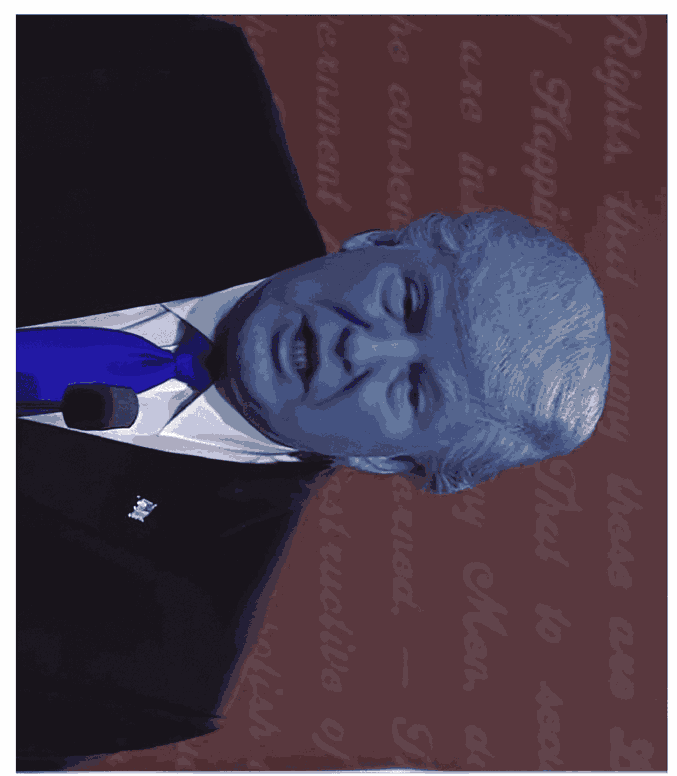}} \quad
	\subfloat[][]
	{\includegraphics[width=.3\textwidth,height=3cm]{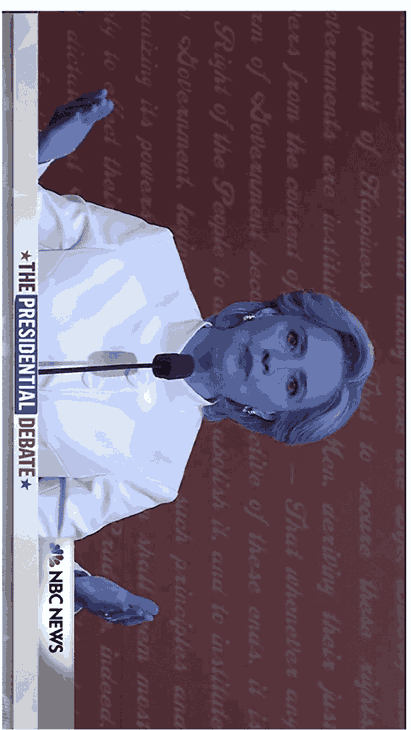}} \\
	\caption{ Images for Deepfake detection using dataset1.}
	\label{figure8}
\end{figure}

\begin{figure}
	\vspace{-18mm}
	\centering
		\noindent 
	\textbf{Enhanced RCNN-Ear Detection Based Images} 
	\par 
	\subfloat[][]
	{\includegraphics[width=.3\textwidth,height=3cm]{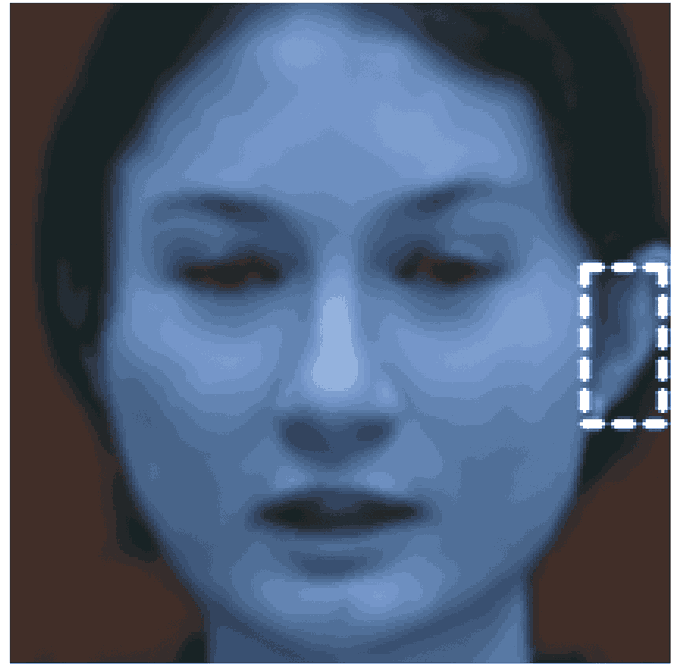}} 
	\subfloat[][]
	{\includegraphics[width=.3\textwidth,height=3cm]{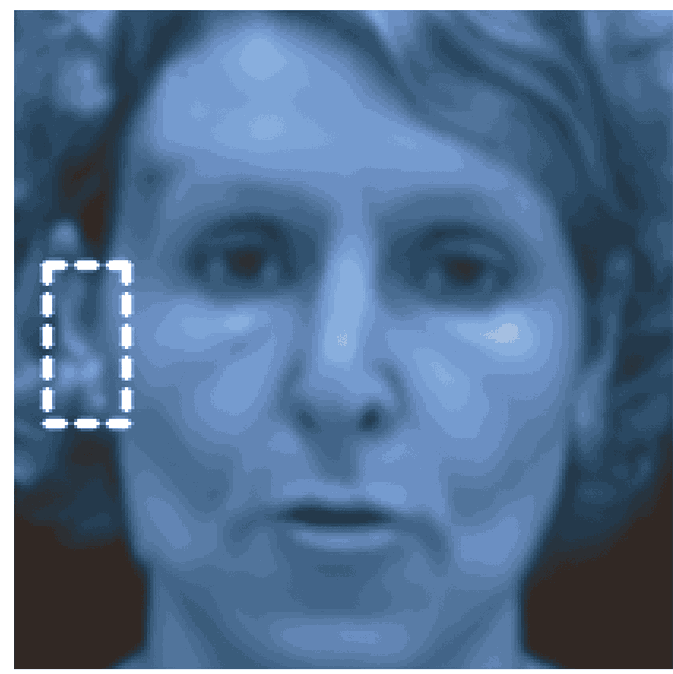}} 
    \subfloat[][]
    {\includegraphics[width=.3\textwidth,height=3cm]{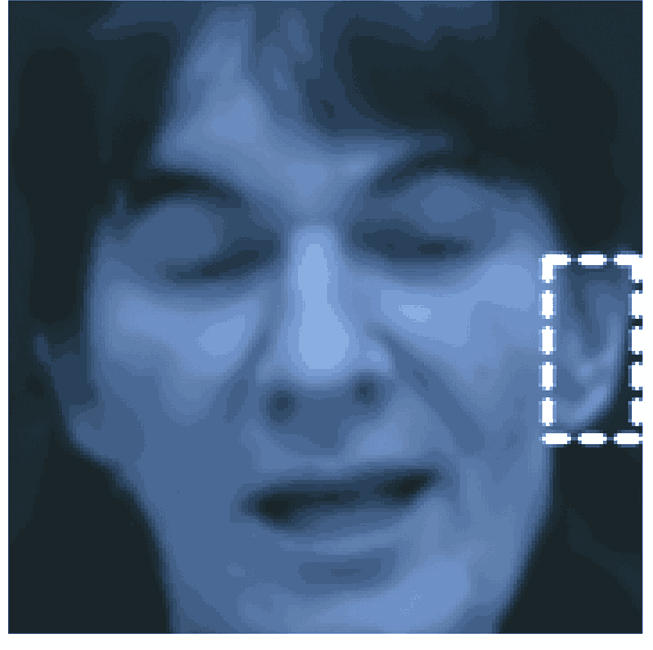}} \\
     \noindent 
   \textbf{Compression Case Images} 
  \par 
	\subfloat[][]
	{\includegraphics[width=.3\textwidth,height=3cm]{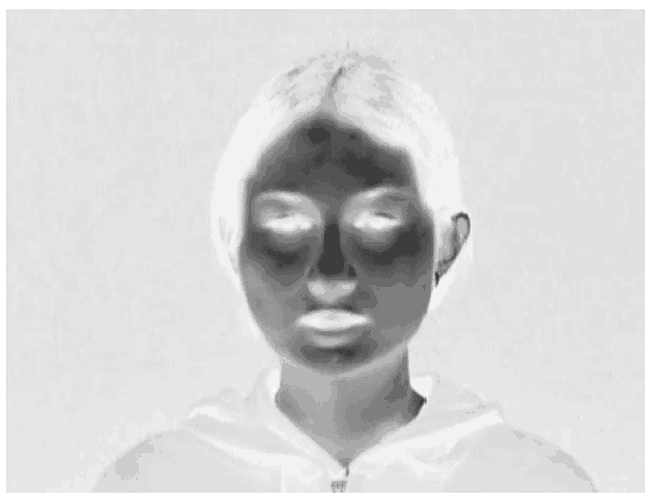}} \quad
	\subfloat[][]
	{\includegraphics[width=.3\textwidth,height=3cm]{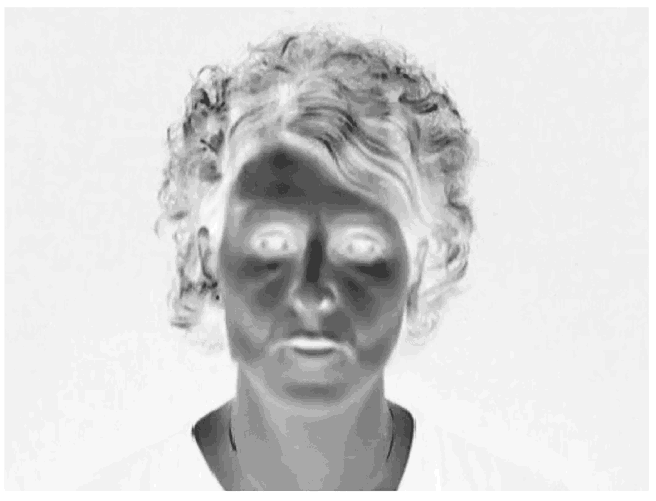}} \quad
	\subfloat[][]
	{\includegraphics[width=.3\textwidth,height=3cm]{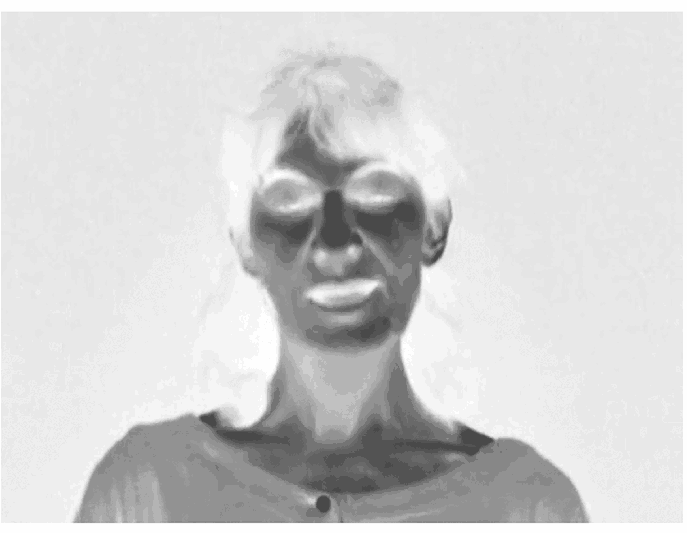}} \\
		\noindent 
	\textbf{Noise Case Images} 
	\par 
	\subfloat[][]
	{\includegraphics[width=.3\textwidth,height=3cm]{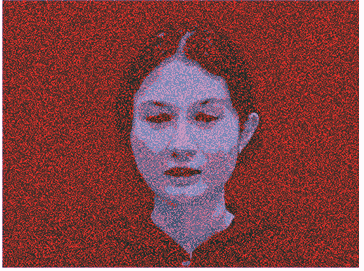}} \quad
	\subfloat[][]
	{\includegraphics[width=.3\textwidth,height=3cm]{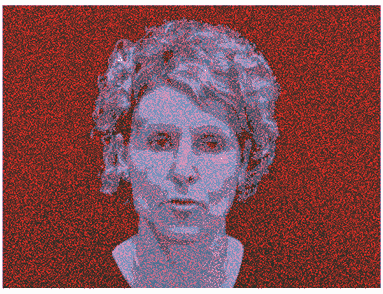}} \quad
	\subfloat[][]
	{\includegraphics[width=.3\textwidth,height=3cm]{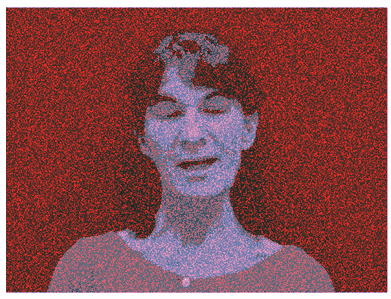}} \\
		\noindent 
	\textbf{Pose Illumination Case Images} 
	\par 
	
	\subfloat[][]
	{\includegraphics[width=.3\textwidth,height=3cm]{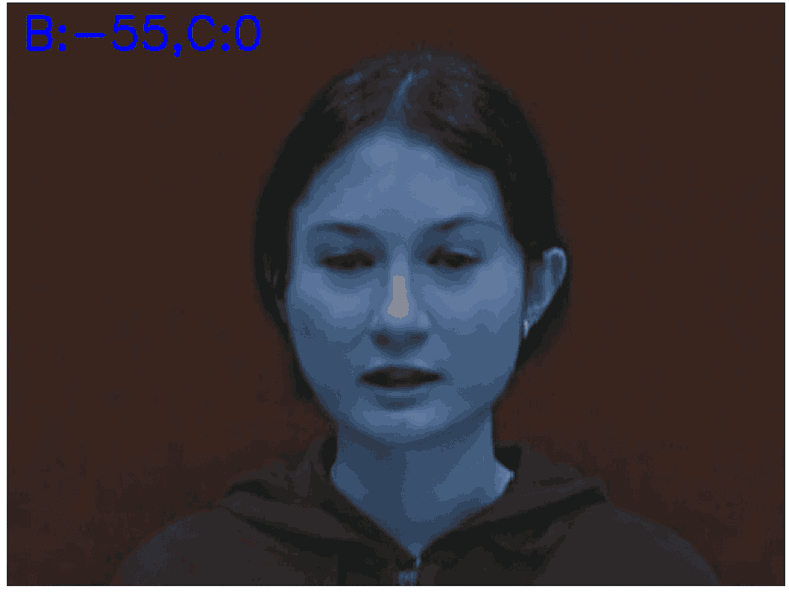}} \quad
	\subfloat[][]
	{\includegraphics[width=.3\textwidth,height=3cm]{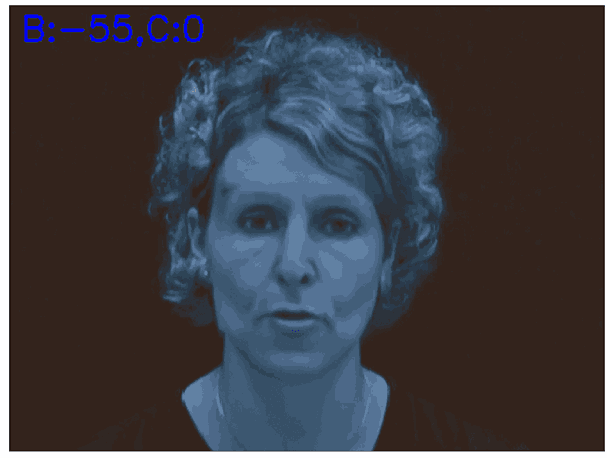}} \quad
	\subfloat[][]
	{\includegraphics[width=.3\textwidth,height=3cm]{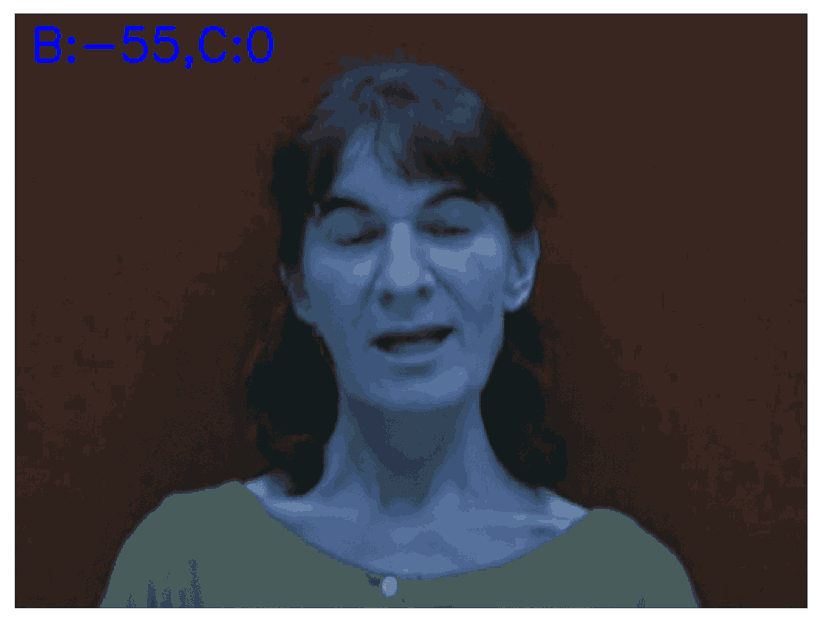}} \\
	\noindent 
\textbf{Rotation Case Images} 
\par 
	\subfloat[][]
	{\includegraphics[width=.3\textwidth,height=3cm]{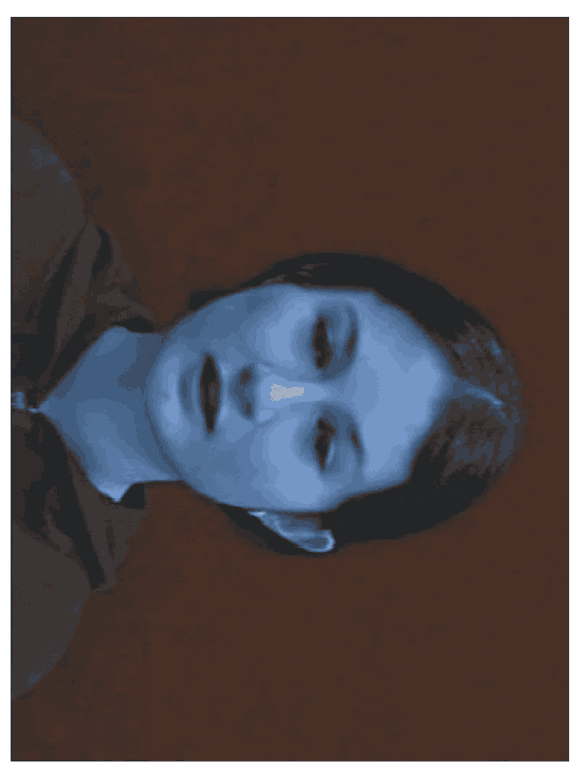}} \quad
	\subfloat[][]
	{\includegraphics[width=.3\textwidth,height=3cm]{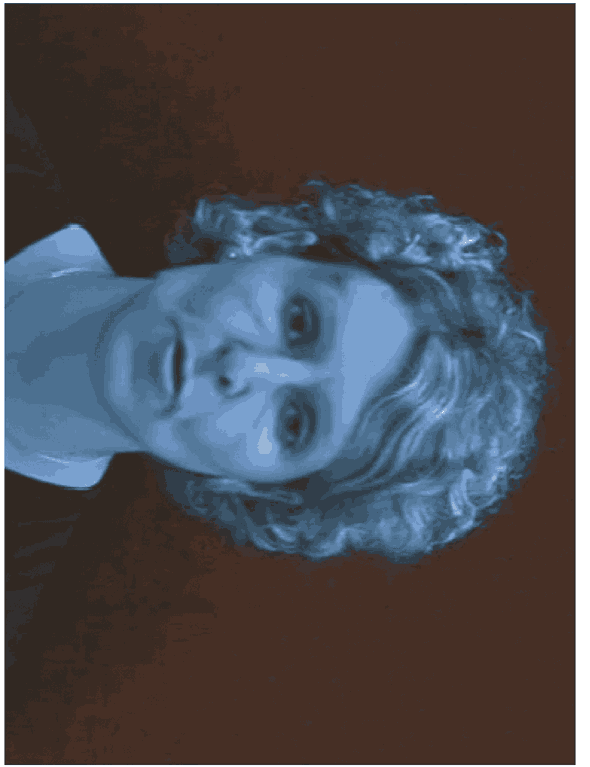}} \quad
	\subfloat[][]
	{\includegraphics[width=.3\textwidth,height=3cm]{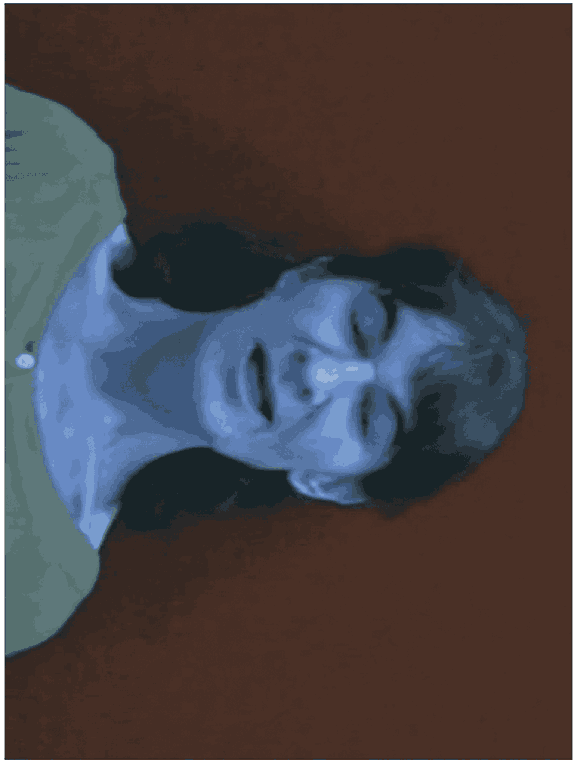}} \\
	\caption{ Images for Deepfake detection using dataset2.}
	\label{figure9}
\end{figure}

\begin{figure}
	\vspace{-15mm}
	\centering
	
	
	\textbf{Enhanced RCNN-Ear detection based images} 
	\par 
	
	\subfloat[][]
	
	{\includegraphics[width=.3\textwidth]{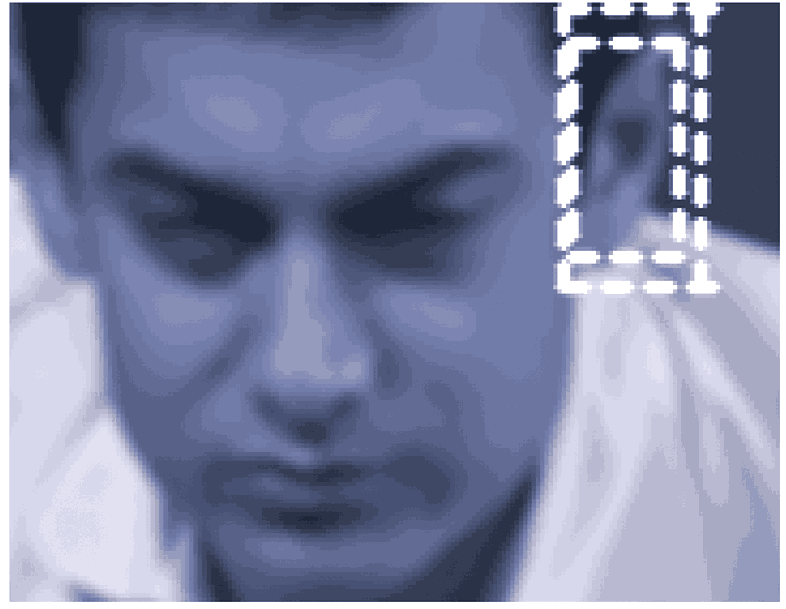}} \quad
	\subfloat[][]
	{\includegraphics[width=.3\textwidth]{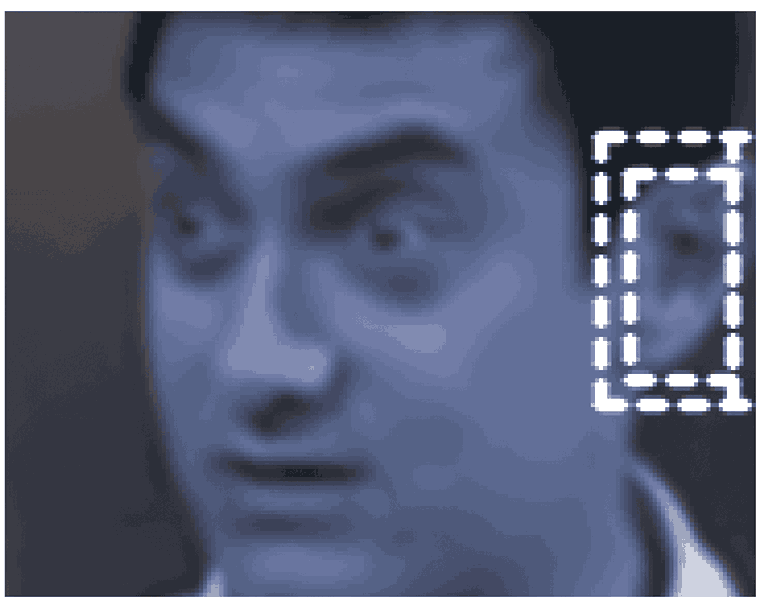}} \quad
	\subfloat[][]
	{\includegraphics[width=.3\textwidth]{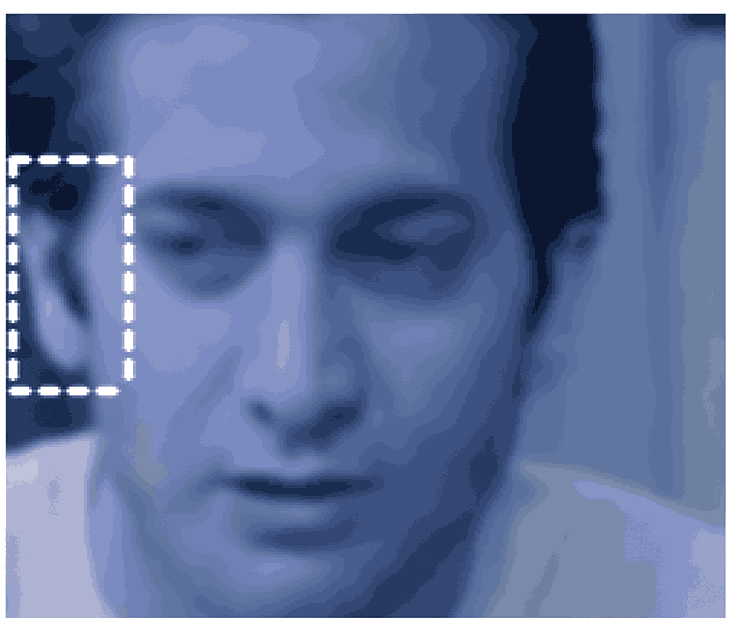}} \\
	\noindent 
	\textbf{Compression Case Images} 
	\par 
	
	\subfloat[][]
	{\includegraphics[width=.3\textwidth,height=3cm]{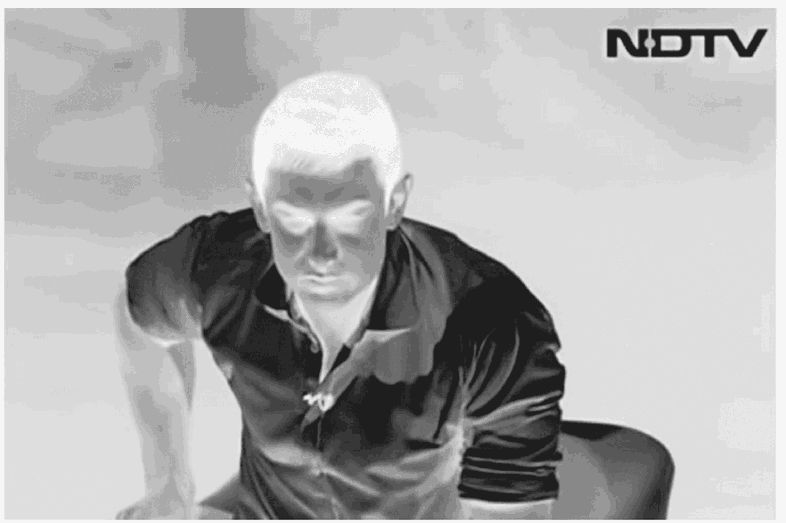}} \quad
	\subfloat[][]
	{\includegraphics[width=.3\textwidth,height=3cm]{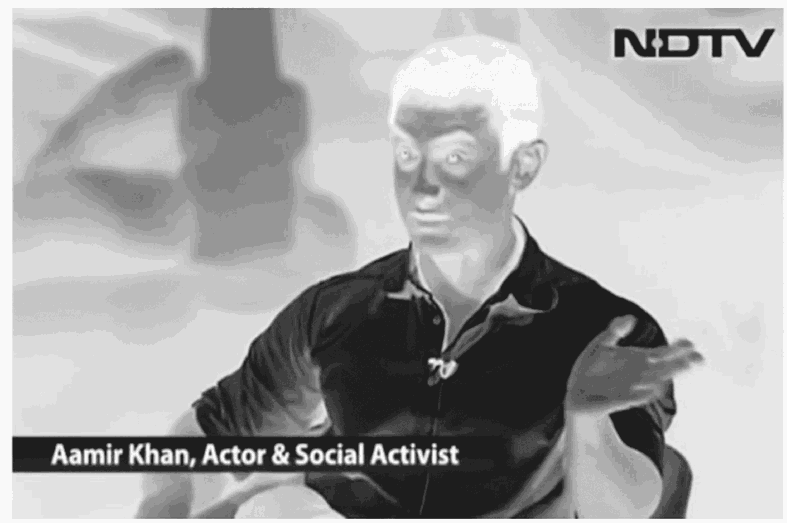}} \quad
	\subfloat[][]
	{\includegraphics[width=.3\textwidth,height=3cm]{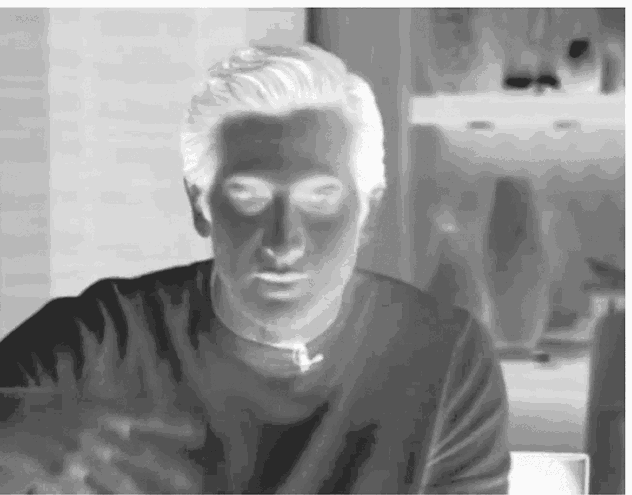}} \\
	\noindent 
	\textbf{Noise Case Images} 
	\par 
	\subfloat[][]
	{\includegraphics[width=.3\textwidth,height=3cm]{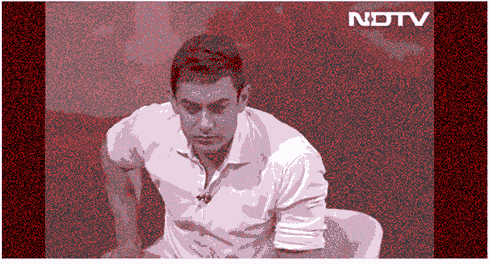}} \quad
	\subfloat[][]
	{\includegraphics[width=.3\textwidth,height=3cm]{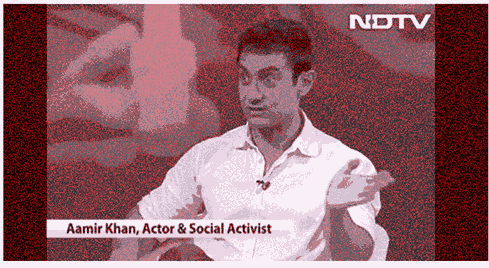}} \quad
	\subfloat[][]
	{\includegraphics[width=.3\textwidth,height=3cm]{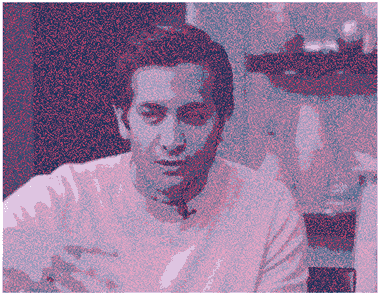}} \\
	\noindent 
	\textbf{Pose Illumination Case Images} 
	\par 
	\subfloat[][]
	{\includegraphics[width=.3\textwidth,height=3cm]{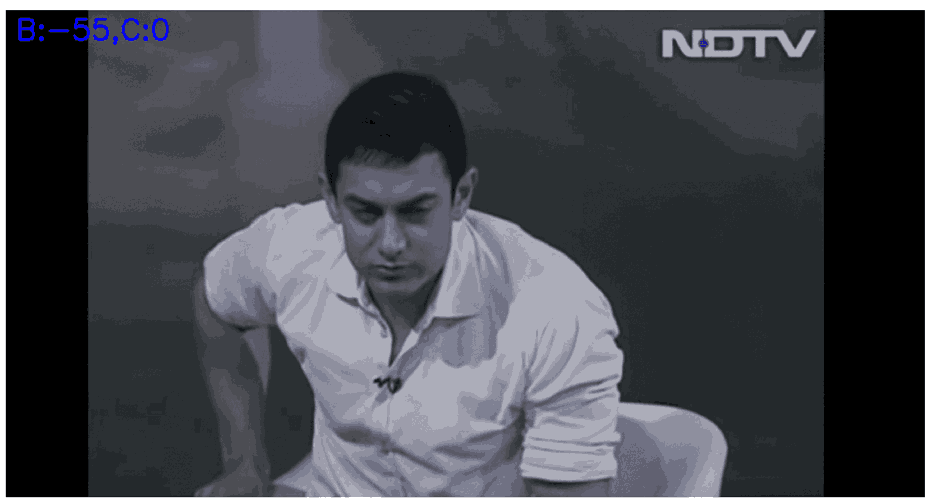}} \quad
	\subfloat[][]
	{\includegraphics[width=.3\textwidth,height=3cm]{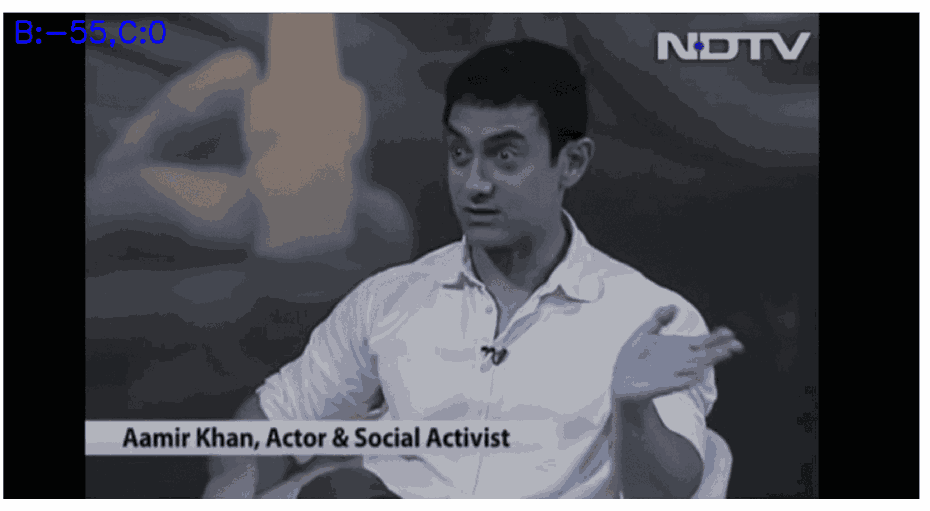}} \quad
	\subfloat[][]
	{\includegraphics[width=.3\textwidth,height=3cm]{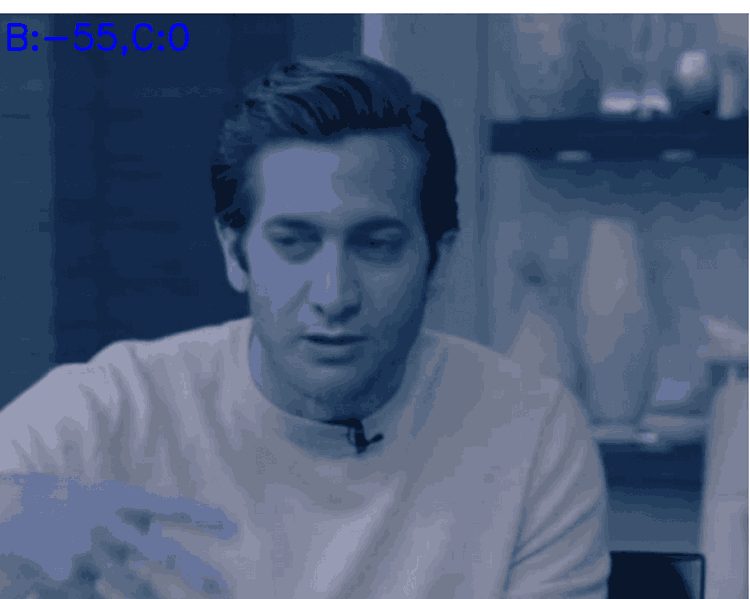}} \\
	\noindent 
	\textbf{Rotation Images} 
	\par 
	\subfloat[][]
	{\includegraphics[width=.3\textwidth,height=3cm]{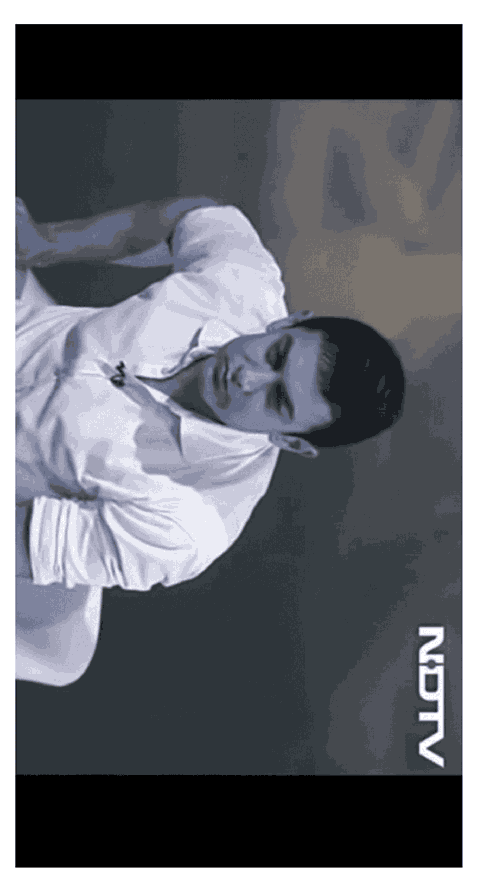}} \quad
	\subfloat[][]
	{\includegraphics[width=.3\textwidth,height=3cm]{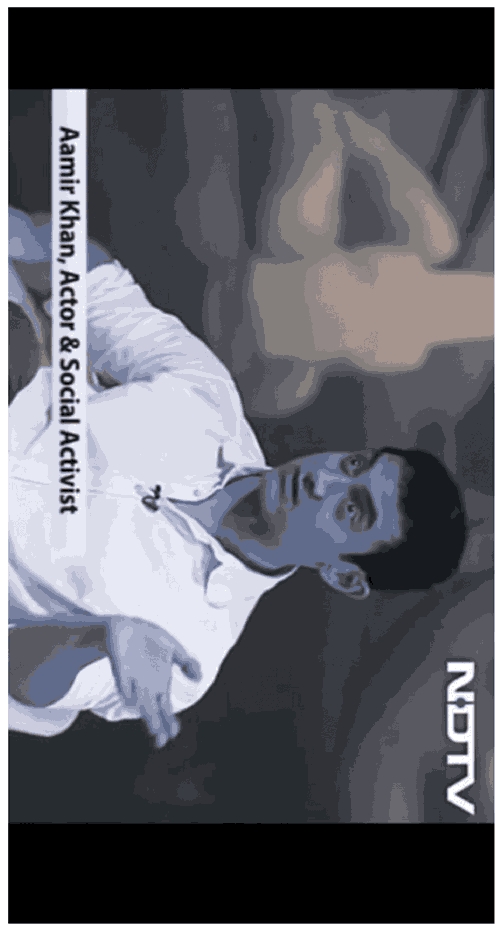}} \quad
	\subfloat[][]
	{\includegraphics[width=.3\textwidth,height=3cm]{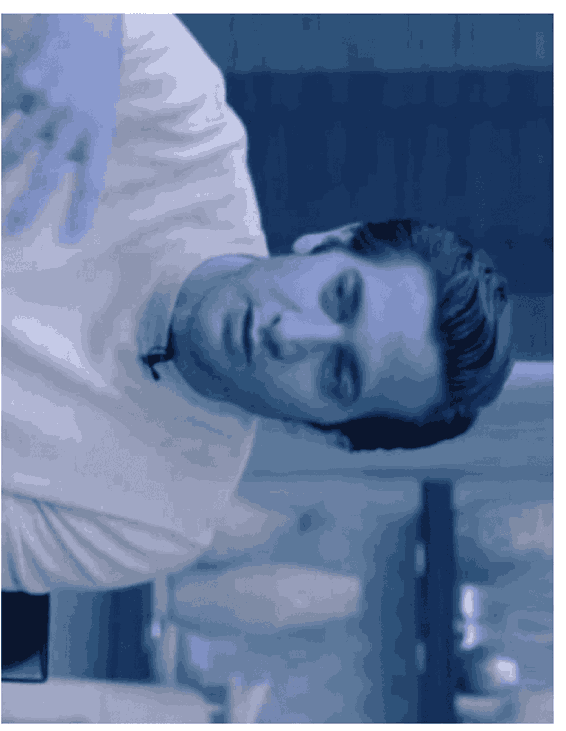}} \\
	\caption{ Images for Deepfake detection using dataset3.}
	\label{figure10}
\end{figure}

      The performance analysis encompassed four diverse types of test cases: (a) Test Case 1: Compression, (b) Test Case 2: Noise, (c) Test Case 3: Pose Illumination, and (d) Test Case 4: Rotation for each of the three datasets. Subsections \ref{section4.6}, \ref{section4.7}, and \ref{section4.8} demonstrate the test cases taken into consideration such as compression, noise, pose Illumination, and rotation for performance evaluation for dataset1, dataset2, and dataset3 respectively. Our proposed model outperforms other approaches in terms of different performance metrics.\\
      Additionally, convergence analysis is performed to compare the SU-JFO method with TSO (Tuna Swarm Optimization) \cite{xie2021tuna}, AO (Aquila Optimization) \cite{abualigah2021aquila}, HGS (Hunger Games Search) \cite{yang2021hunger}, DMO (Dwarf Mangoose Optimization) \cite{agushaka2022dwarf}, HBA (Honey Badger Algorithm) \cite{hashim2022honey}, and JFO (Jelly Fish Optimization) \cite{chou2021novel} which has been described in subsection \ref{subsection4.10}. Besides this, an ablation study was carried out on the SU-JFO approach to provide a clear understanding of including specific components within the SU-JFO described in subsection \ref{subsection4.9}. Figure \ref{figure8}, Figure \ref{figure9}, and Figure \ref{figure10} displays a sequence of images depicting Enhanced RCNN-Ear detection based images as well as cases involving compression, noise, pose illumination, and rotation. These images are crucial for deepfake detection using Dataset1 \cite{agarwal2019protecting}, Dataset2 \cite{deepfaketimit}, and Dataset3 \cite{yuezun2019celeb}. 

    \subsection{Comparative Analysis on Dataset1} \label{section4.6}
     \subsubsection{Test Case 1: Compression Scenario}
\begin{figure*}[!htbp]
   	\vspace{-10mm}
\includegraphics[height=5cm,width=\textwidth]{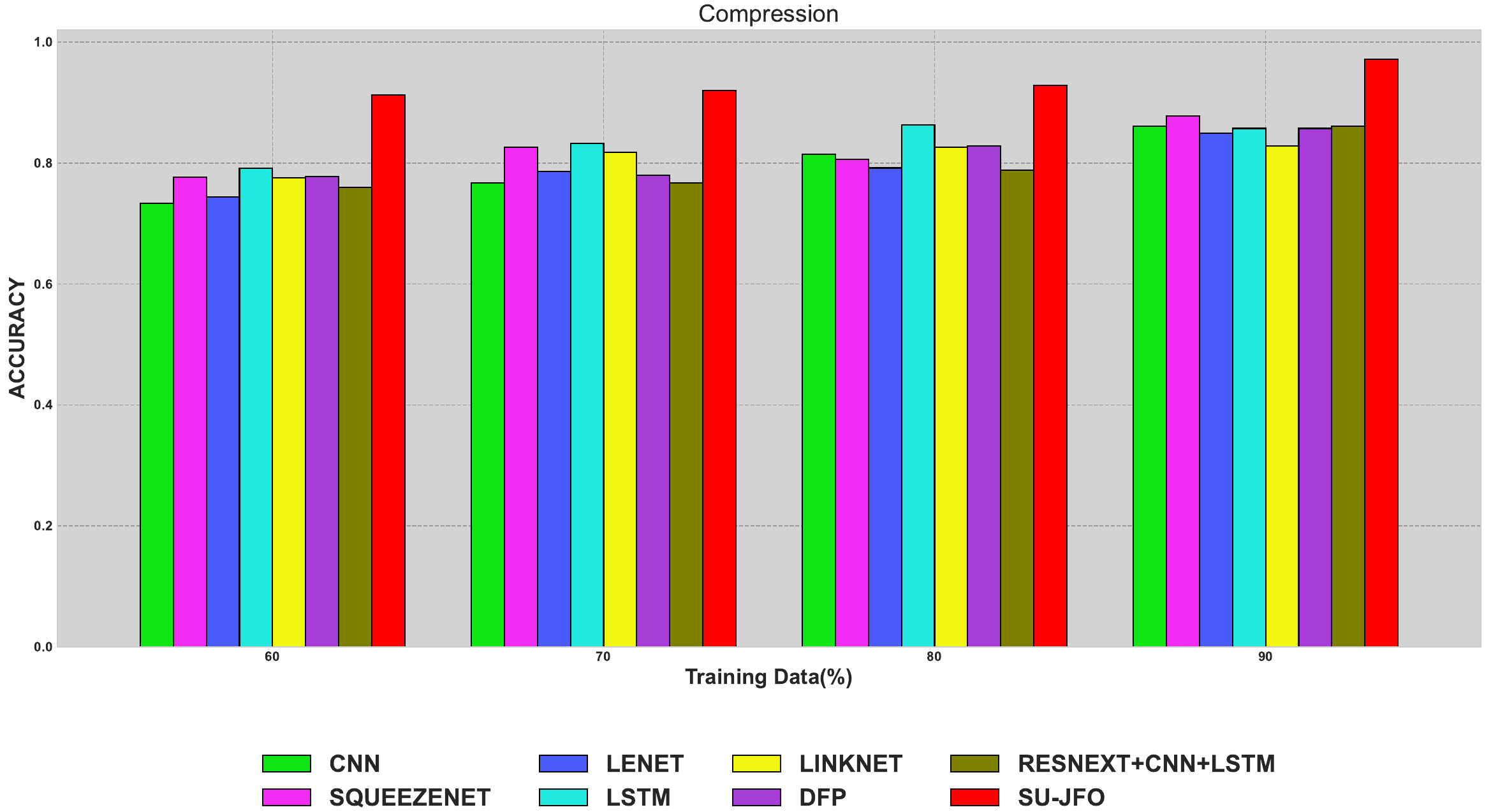}\hfill
\includegraphics[height=5cm,width=\textwidth]{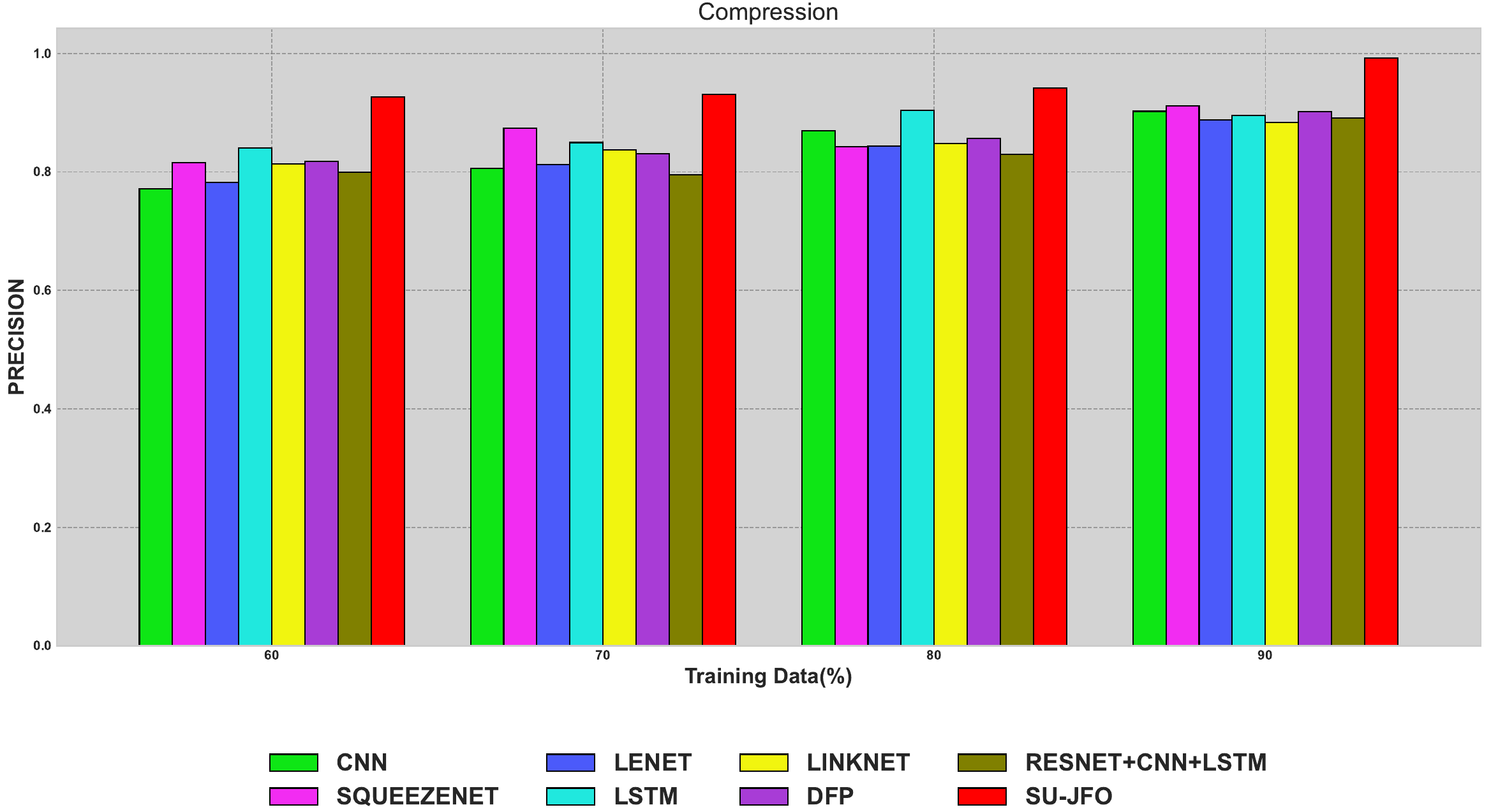}\hfill
\includegraphics[height=5cm,width=\textwidth]{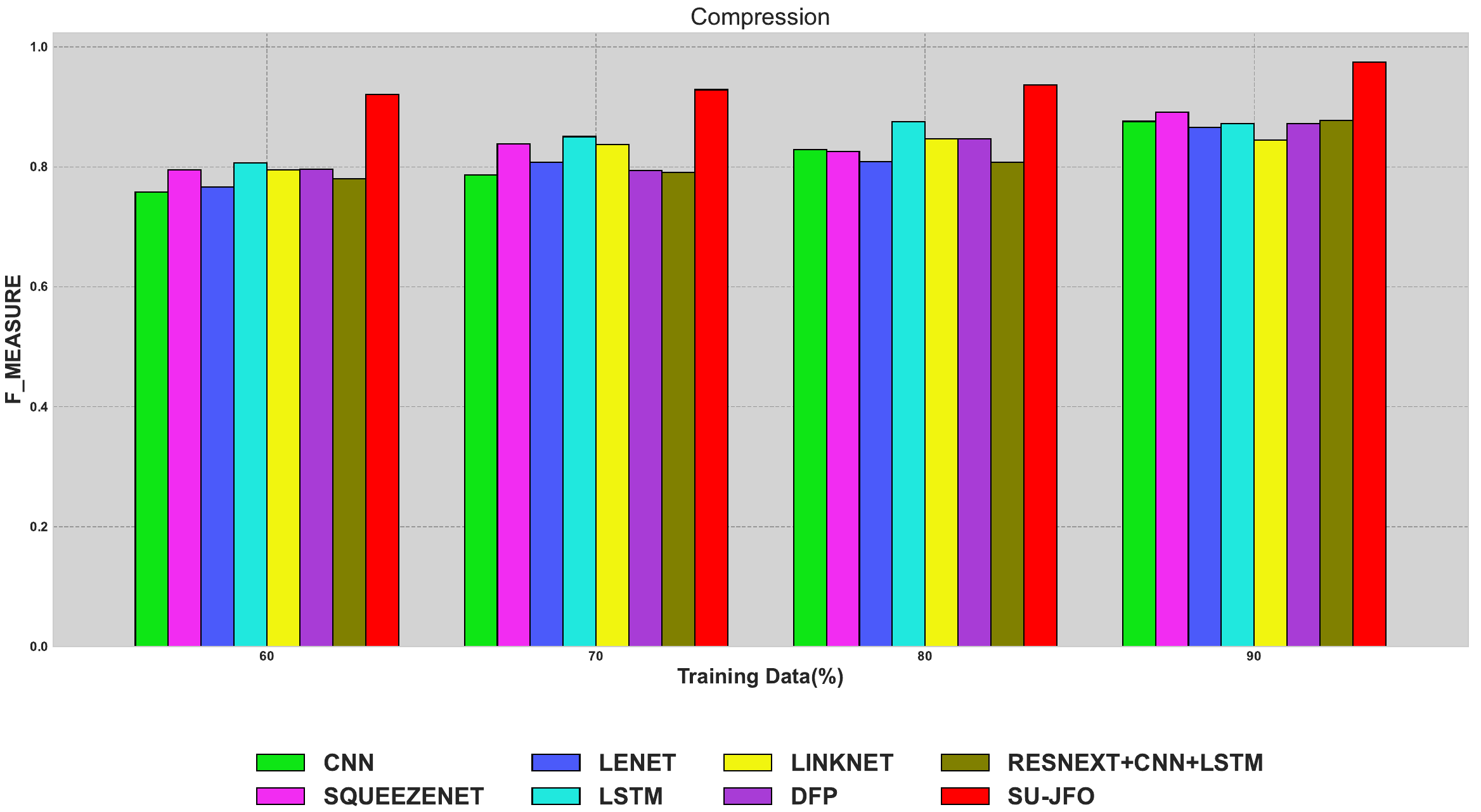}\hfill
\includegraphics[height=5cm,width=\textwidth]{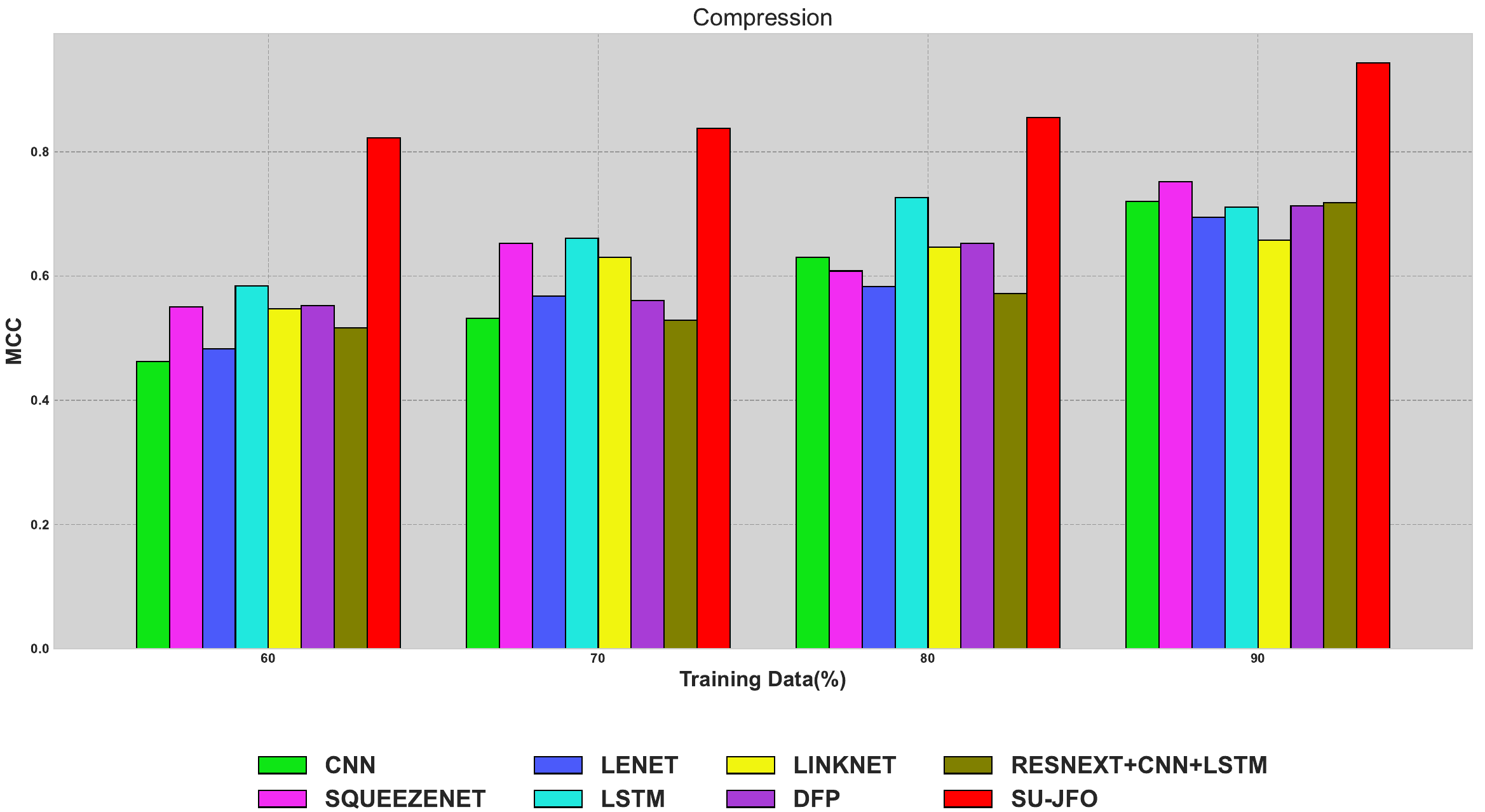}
	\caption{Assessment of SU-JFO and traditional schemes for Compression case on Dataset1 a) Accuracy b) Precision c) F-measure and d) MCC.}\label{Fig11}
\end{figure*}

      In this case, both real and fake data undergo compression, and an evaluation is carried out with SU-JFO method and conventional approaches such as CNN, SqueezeNet, LeNet, LSTM, LinkNet, DFP \cite{raza2022novel}, and ResNext+CNN+LSTM \cite{vamsi2022deepfake} considering Precision, MCC (Matthews's correlation coefficient), Accuracy, and F-measure. Specifically, the SU-JFO method is compared with CNN, SqueezeNet, LeNet, LSTM, LinkNet, DFP \cite{raza2022novel}, and ResNext+CNN+LSTM \cite{vamsi2022deepfake}. The obtained results are illustrated in Figure \ref{Fig11}. The SU-JFO method exhibited superior performance compared to conventional approaches. Specifically, the SU-JFO method attained the highest accuracy of 0.971 at 90\% training data. In contrast, traditional methods recorded lower accuracy rates with CNN at 0.861 , SqueezeNet at 0.877, LeNet at 0.848, LSTM at 0.857, LinkNet at 0.828, DFP \cite{raza2022novel} at 0.857, and ResNext+CNN+LSTM \cite{vamsi2022deepfake} at 0.861 respectively. Furthermore, at the 70\% training data, the SU-JFO method achieved a significantly higher F-measure value of 0.928 surpassing the values obtained by CNN, SqueezeNet, LeNet, LSTM, LinkNet, DFP \cite{raza2022novel}, and ResNext+CNN+LSTM \cite{vamsi2022deepfake}. The noteworthy findings from this evaluation shows the effectiveness of the SU-JFO method in the detection of deepfakes. The novelty of detecting deepfakes based on ear biometrics can be attributed to the hybrid detection approach and the improvements made in the features associated with detecting the ear through an enhanced score-level fusion strategy.\\  

\begin{figure*}[!t]
	\includegraphics[height=9cm,width=\textwidth]{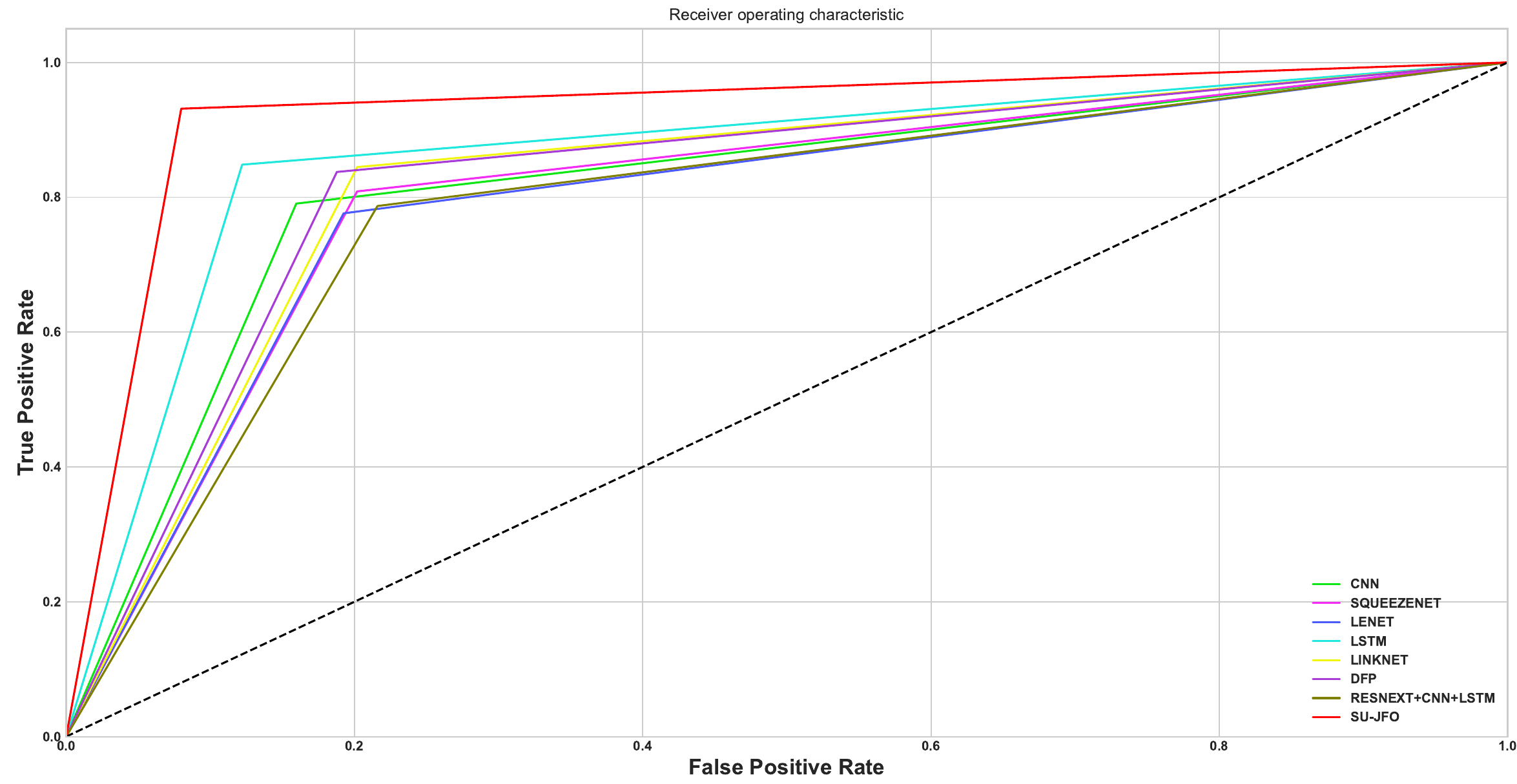}\hfill
	\caption{Evaluation on ROC for Compression Case on Dataset1.}\label{Fig12}
\end{figure*}

\noindent \textbf{{ROC Analysis for Compression Case:}}
Figure \ref{Fig12} shows the ROC evaluation comparing the performance of the SU-JFO method with traditional approaches in deepfake detection. Additionally, the SU-JFO method is compared against the models including CNN, SqueezeNet, LeNet, LSTM, LinkNet, DFP \cite{raza2022novel}, and ResNext+CNN+LSTM \cite{vamsi2022deepfake}. The figure showcases ROC curves visually demonstrating how a model's sensitivity (True Positive Rates(TPR)) and specificity (1-FPR ) adjust with different decision thresholds. Each point on the ROC curve corresponds to a specific threshold used to classify data into positive and negative categories. While a higher AUC (Area under curve) value signifies a discrimination between positive and negative categories, it does not directly measure the overall accuracy of a model's predictions. Furthermore, an experiment was conducted on the training data comprising 70\%. In particular, the SU-JFO method demonstrated a significantly increased TPR of 0.991 at a False Positive Rate (FPR) set to 1.0 outperforming other methods such as CNN, SqueezeNet, LeNet, LSTM, LinkNet, DFP \cite{raza2022novel}, and ResNext+CNN+LSTM \cite{vamsi2022deepfake} which exhibited lower TPR. This compelling evidence highlights its exceptional capability in effectively detecting deepfakes. Hence, it is evident that our approach effectively outperform in compression scenario. \\
   
\noindent \textbf{Statistical Analysis on Accuracy for Compression Case:} 
For statistical measure, each model has been evaluated against ``minimum, maximum, mean, standard deviation, and median". Table \ref{table2} describes the statistical evaluation of SU-JFO and existing methods for deepfake detection. The model was trained and executed 25 times to calculate statistical measures of accuracy including the mean, maximum, standard deviation, median, and minimum to obtain optimal results. For exact detection of deepfake the model should generate maximal accuracy values. In our work, the SU-JFO achieved higher accuracy values than the traditional methods. Considering the median statistical metric, the SU-JFO yields the greater accuracy rate of 0.924 while the CNN, SqueezeNet, LeNet, LSTM, LinkNet, DFP \cite{raza2022novel}, and ResNext+CNN+LSTM \cite{vamsi2022deepfake}  accomplished the higher accuracy values of 0.791, 0.816, 0.789, 0.845, 0.822, 0.804 and 0.778 respectively. 
\begin{center}
	\begin{table}[!htbp]
		\resizebox{\textwidth}{!}{%
			\begin{tabular}{|l|l|l|l|l|l|l|l|l|}
				\hline
				\begin{tabular}[c]{@{}c@{}}\textbf{Statistical}\\ \textbf{Metrics}\end{tabular} & \textbf{CNN} & \textbf{SqueezeNet}  & \textbf{LeNet}  & \textbf{LSTM}  & \textbf{LinkNet}  & \textbf{DFP}  & \begin{tabular}[c]{@{}c@{}}\textbf{ResNext+CNN+}\\ \textbf{LSTM}\end{tabular}
   & \begin{tabular}[c]{@{}c@{}}\textbf{SU-JFO}\end{tabular}      \\ 
 \hline  
Mean & 0.794 & 0.821 & 0.793 & 0.836 & 0.812 & 0.811 & 0.794 & 0.933\\ \hline
Maximum & 0.861 & 0.878 & 0.849 & 0.863 & 0.829 & 0.857 & 0.861 & 0.971\\ \hline 
Standard Deviation  & 0.048 & 0.037 & 0.037 & 0.028 & 0.022 & 0.034 & 0.040 & 0.023  \\ \hline
Median & 0.791 & 0.816 & 0.789 & 0.845 & 0.822 & 0.804 & 0.778 & 0.924\\ \hline
Minimum & 0.733 & 0.776 & 0.744 & 0.792 & 0.775 & 0.777 & 0.760 & 0.912\\ \hline
			\end{tabular}%
		}
		\caption{Statistical Assessment on Accuracy For Compression Case on Dataset1.}
		\label{table2}
	\end{table}
\end{center}
\vspace{-5mm}

\subsubsection{Test Case 2:  Noise Scenario}
\begin{figure*}[!htbp]
	\vspace{-10mm}
\includegraphics[height=5cm,width=\textwidth]{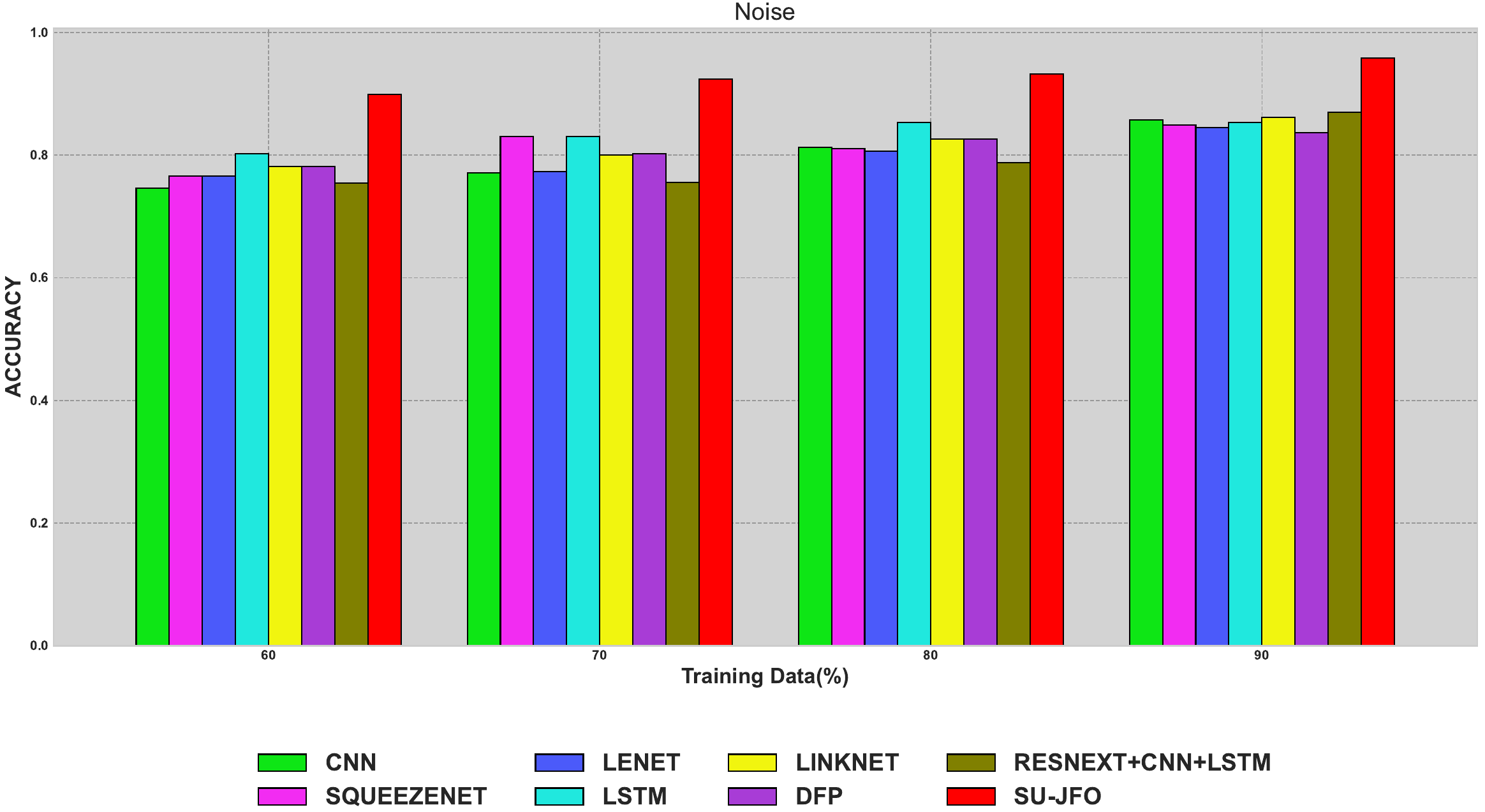}\hfill
\includegraphics[height=5cm,width=\textwidth]{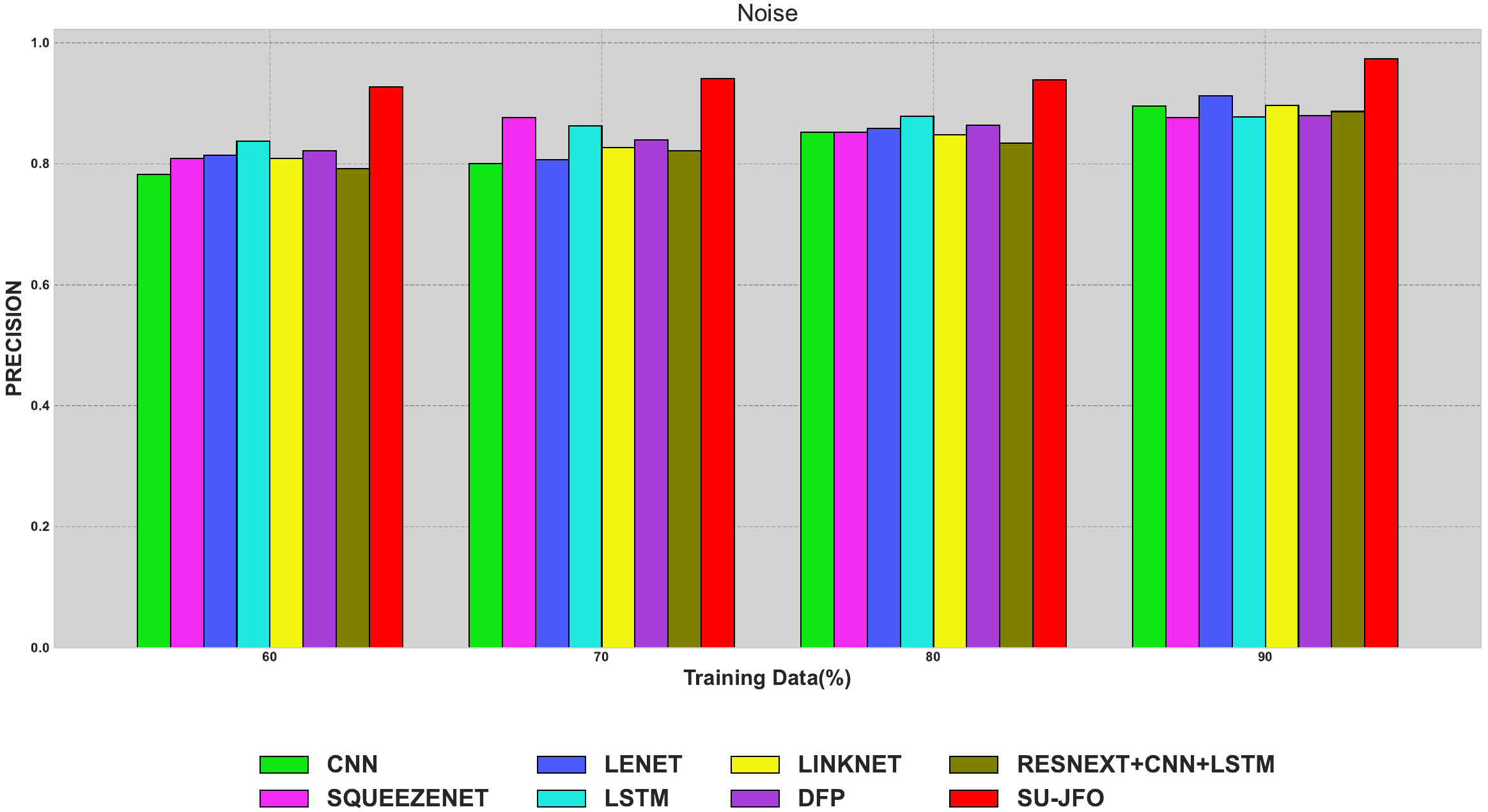}\hfill
\includegraphics[height=5cm,width=\textwidth]{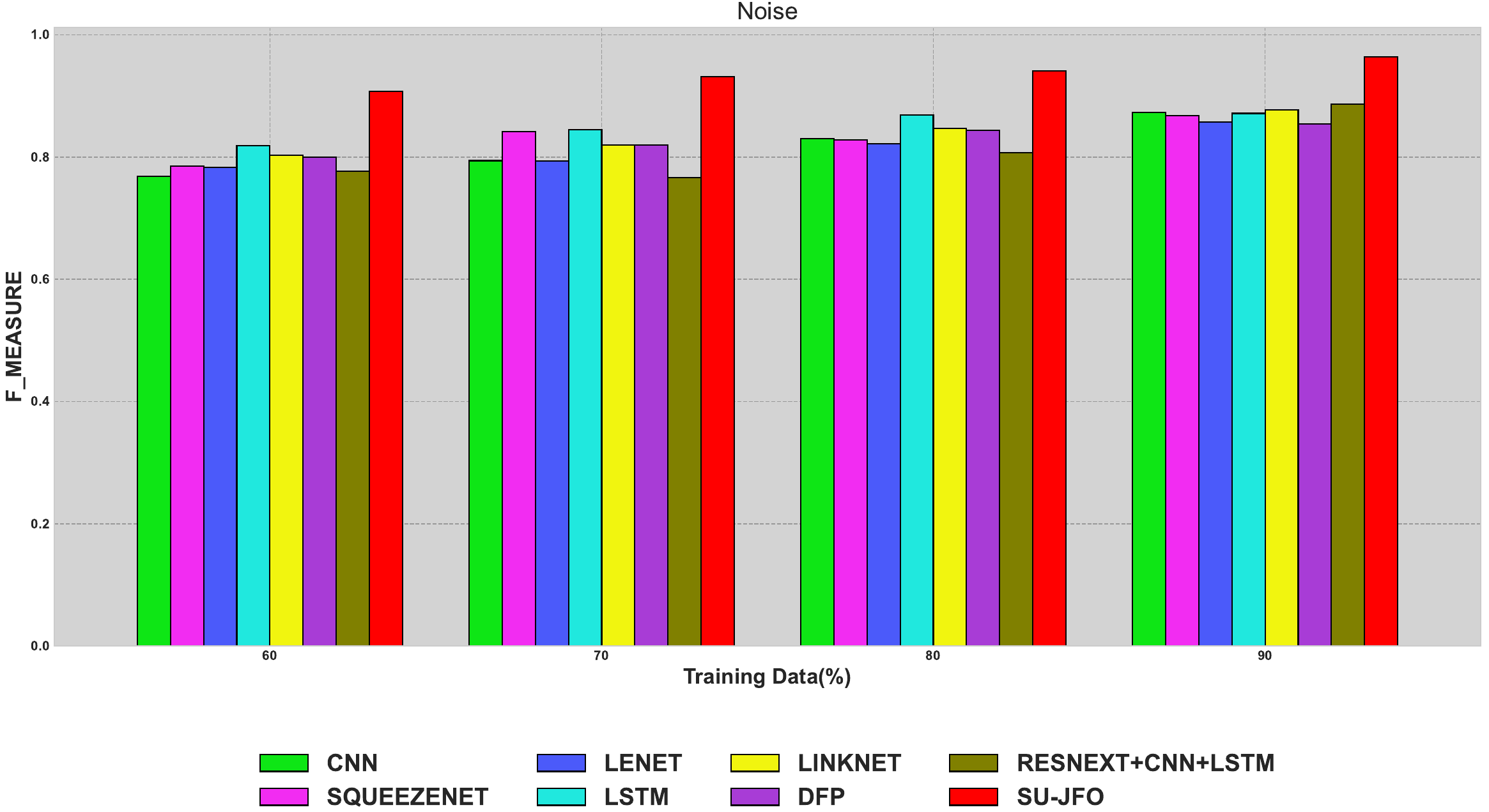}\hfill
\includegraphics[height=5cm,width=\textwidth]{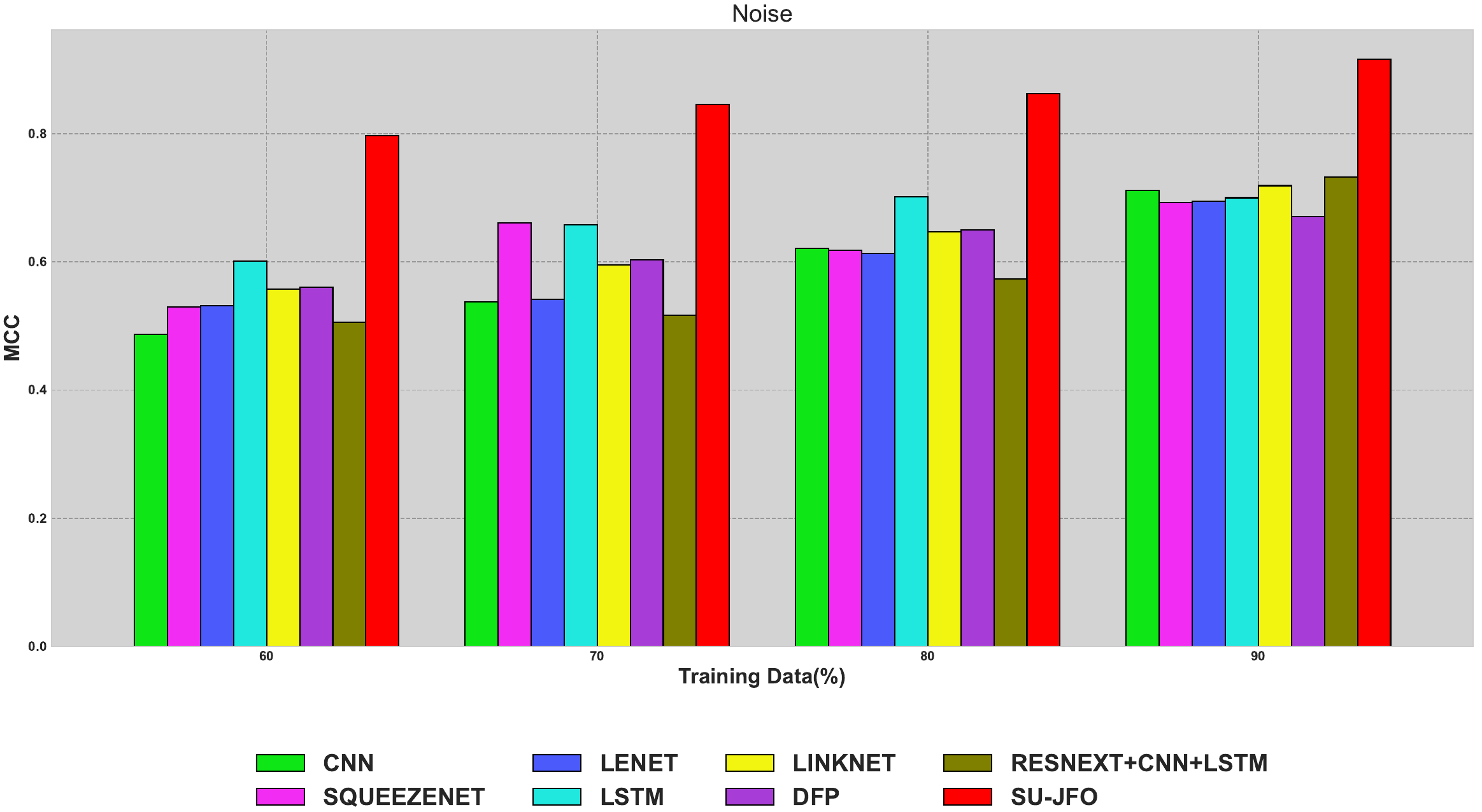}
	\caption{Assessment of SU-JFO and traditional schemes for Noise case on Dataset1 \\ a) Accuracy b) Precision c) F-measure and d) MCC.}\label{Fig13}
\end{figure*}

In this scenario, noise has been incorporated into both real and fake data and a comparative analysis has been carried out to assess the efficacy of SU-JFO compared to traditional methods. Figure \ref{Fig13} illustrates the evaluation of accuracy, precision, F-measure, and MCC for deepfake detection contrasting the SU-JFO approach with CNN, SqueezeNet, LeNet, LSTM, LinkNet, DFP \cite{raza2022novel}, and ResNext+CNN+LSTM \cite{vamsi2022deepfake}. Notably, the SU-JFO method demonstrated superior precision reaching 0.939 at the 80\% training data while CNN, SqueezeNet, LeNet, LSTM, LinkNet, DFP \cite{raza2022novel}, and ResNext+CNN+LSTM \cite{vamsi2022deepfake}  exhibited lower precision values. Additionally, the SU-JFO method consistently outperformed  other models in terms of MCC values across the entire training data surpassing the respective values of CNN, SqueezeNet, LeNet, LSTM, LinkNet, DFP \cite{raza2022novel}, and ResNext+CNN+LSTM \cite{vamsi2022deepfake}.\\

\begin{figure*}[!t]
	\includegraphics[height=9cm,width=\textwidth]{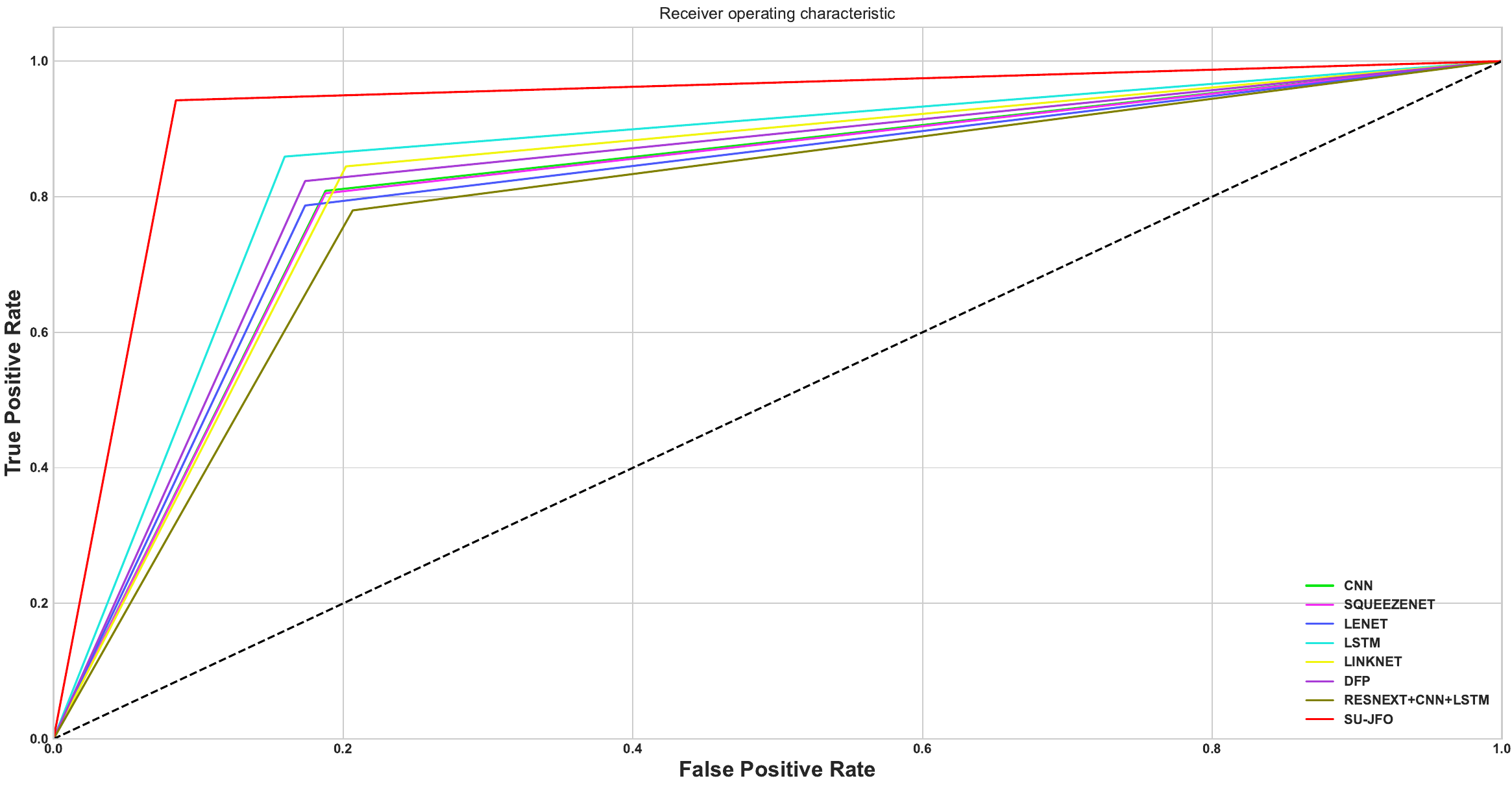}\hfill
	\caption{Evaluation on ROC for Noise Case on Dataset1.}\label{Fig14}
\end{figure*}

\noindent \textbf{ROC Analysis for Noise Case:}
In Figure \ref{Fig14}, the ROC evaluation illustrates a comparison between the SU-JFO method and several conventional techniques including CNN, SqueezeNet, LeNet, LSTM, LinkNet, DFP \cite{raza2022novel}, and ResNext+CNN+LSTM \cite{vamsi2022deepfake} for detecting deepfakes. From Figure \ref{Fig14}, it is evident that the SU-JFO approach consistently outperformed the traditional methods regarding TPR. Notably, when the FPR was set at 0.8, the SU-JFO method demonstrated the highest TPR of 0.835 surpassing the performance of CNN, SqueezeNet, LeNet, LSTM, LinkNet, DFP \cite{raza2022novel}, and ResNext+CNN+LSTM \cite{vamsi2022deepfake} which exhibited comparatively lower TPR in this particular scenario.\\

\noindent \textbf{Statistical Analysis on Accuracy For Noise Case:}
Table \ref{table3} presents a statistical analysis comparing the deepfake detection performance of the SU-JFO method against traditional approaches such as CNN, SqueezeNet, LeNet, LSTM, LinkNet, DFP \cite{raza2022novel}, and ResNext+CNN+LSTM \cite{vamsi2022deepfake}. The model was evaluated over 25 trials. Notably, the SU-JFO method exhibited the highest accuracy rate reaching 0.959 according to the maximum statistical metric. In contrast, CNN, SqueezeNet, LeNet, LSTM, LinkNet DFP \cite{raza2022novel}, and ResNext+CNN+LSTM \cite{vamsi2022deepfake}  showed relatively lower accuracy values.
\begin{center}
	\begin{table}
		\resizebox{\textwidth}{!}{%
			\begin{tabular}{|l|l|l|l|l|l|l|l|l|}
				\hline
				\begin{tabular}[c]{@{}c@{}}\textbf{Statistical}\\ \textbf{Metrics}\end{tabular} & \textbf{CNN} & \textbf{SqueezeNet}  & \textbf{LeNet}  & \textbf{LSTM}  & \textbf{LinkNet}  & \textbf{DFP}  & \begin{tabular}[c]{@{}c@{}}\textbf{ResNext+CNN+}\\ \textbf{LSTM}\end{tabular}
   & \begin{tabular}[c]{@{}c@{}}\textbf{SU-JFO}\end{tabular}      \\ 
 \hline  
Mean & 0.797 & 0.814 & 0.797 & 0.834 & 0.817 & 0.812 & 0.792 & 0.929\\ \hline
Maximum & 0.857 & 0.849 & 0.845 & 0.853 & 0.861 & 0.837 & 0.869 & 0.959\\ \hline 
Standard Deviation  & 0.042 & 0.031 & 0.031 & 0.021 & 0.030 & 0.021 & 0.047 & 0.021  \\ \hline
Median & 0.792 & 0.820 & 0.789 & 0.841 & 0.813 & 0.815 & 0.771 & 0.928\\ \hline
Minimum & 0.746 & 0.766 & 0.766 & 0.802 & 0.781 & 0.781 & 0.755 & 0.899\\ \hline
			\end{tabular}%
		}
		\caption{Statistical Assessment on Accuracy For Noise Case on Dataset1.}
		\label{table3}
	\end{table}
\end{center}

\vspace{-10mm}

\subsubsection{Test Case 3: Pose Illumination Scenario}
\begin{figure*}[!htbp]
	\vspace{-10mm}
\includegraphics[height=5cm,width=\textwidth]{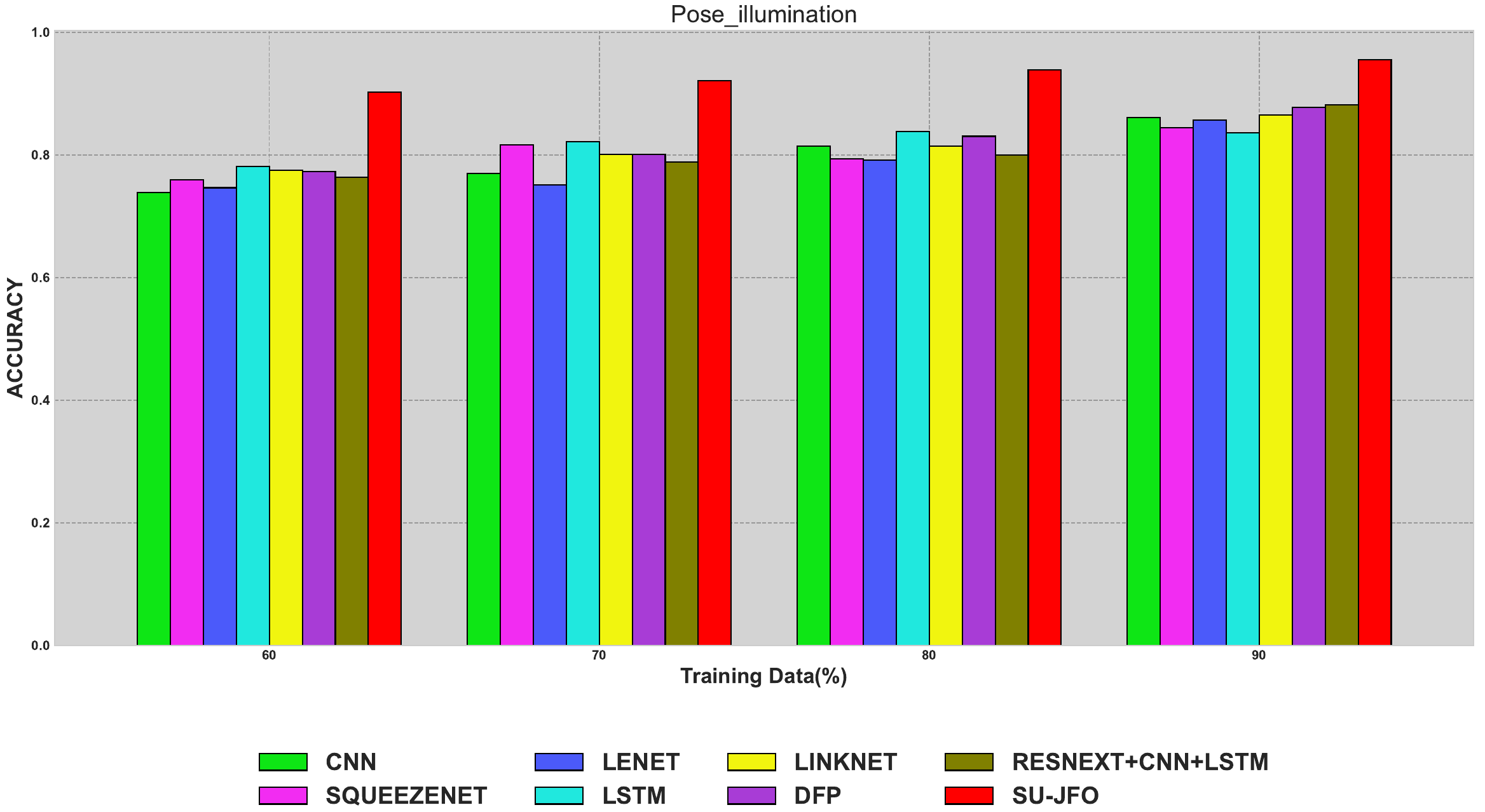}\hfill
\includegraphics[height=5cm,width=\textwidth]{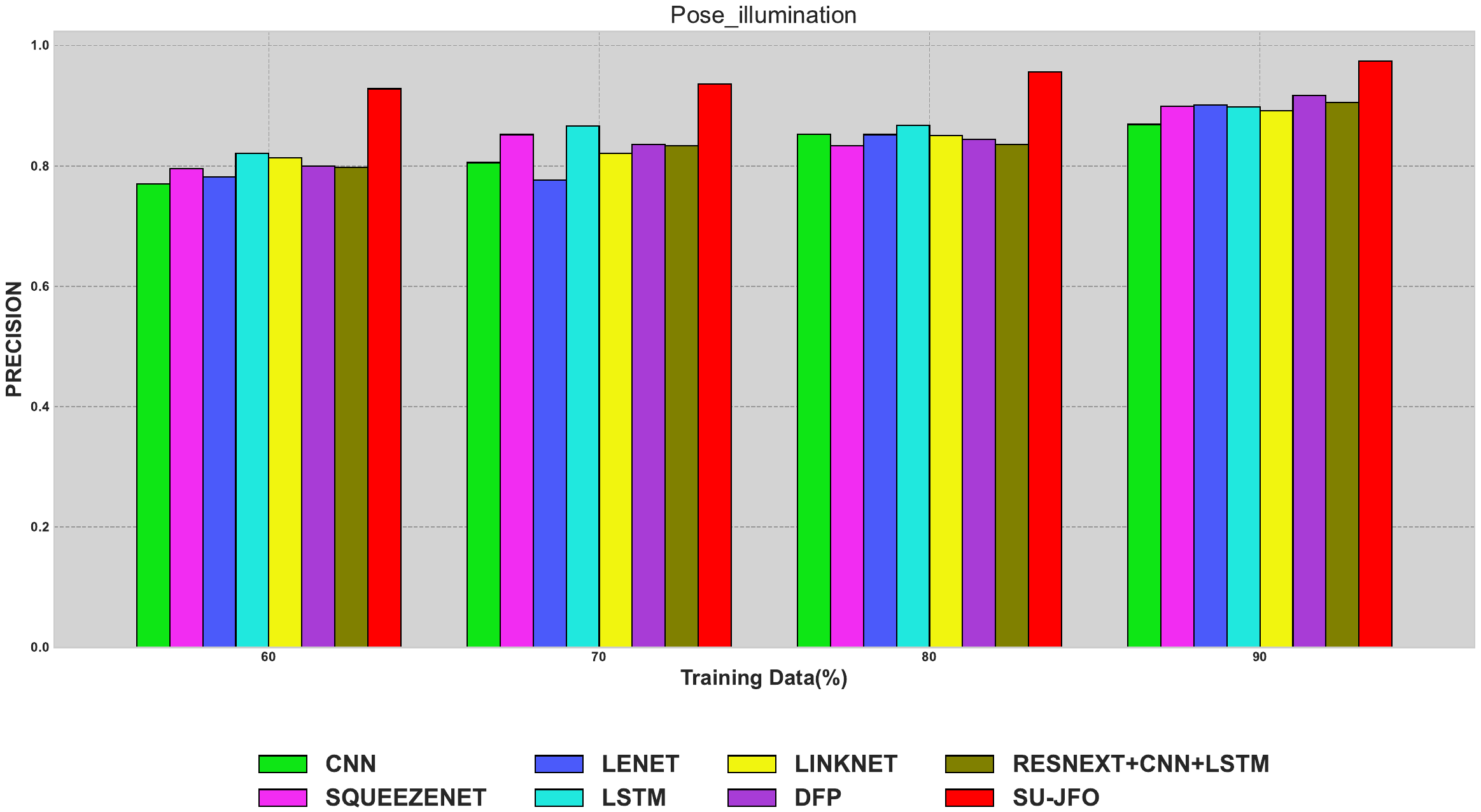}\hfill
\includegraphics[height=5cm,width=\textwidth]{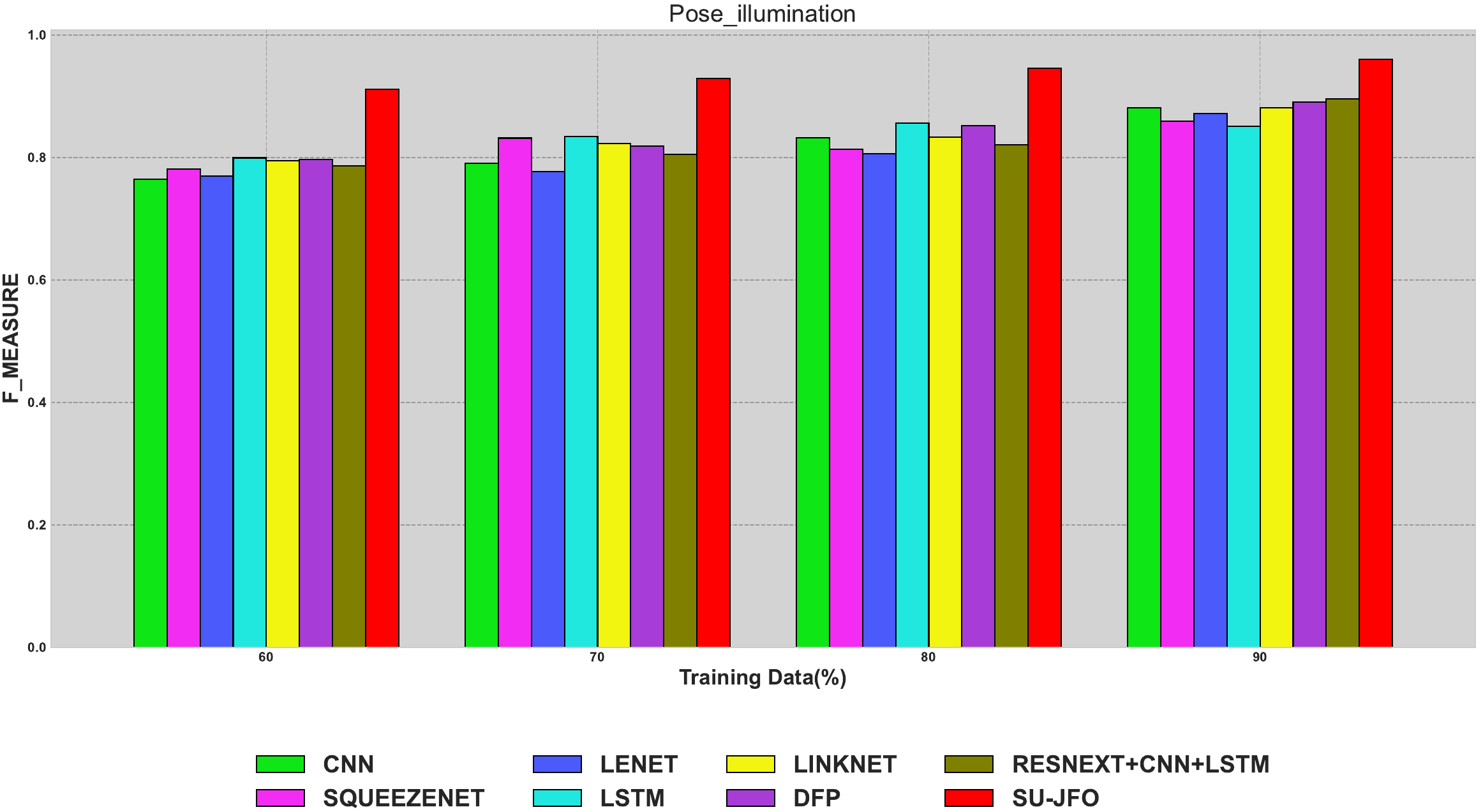}\hfill
\includegraphics[height=5cm,width=\textwidth]{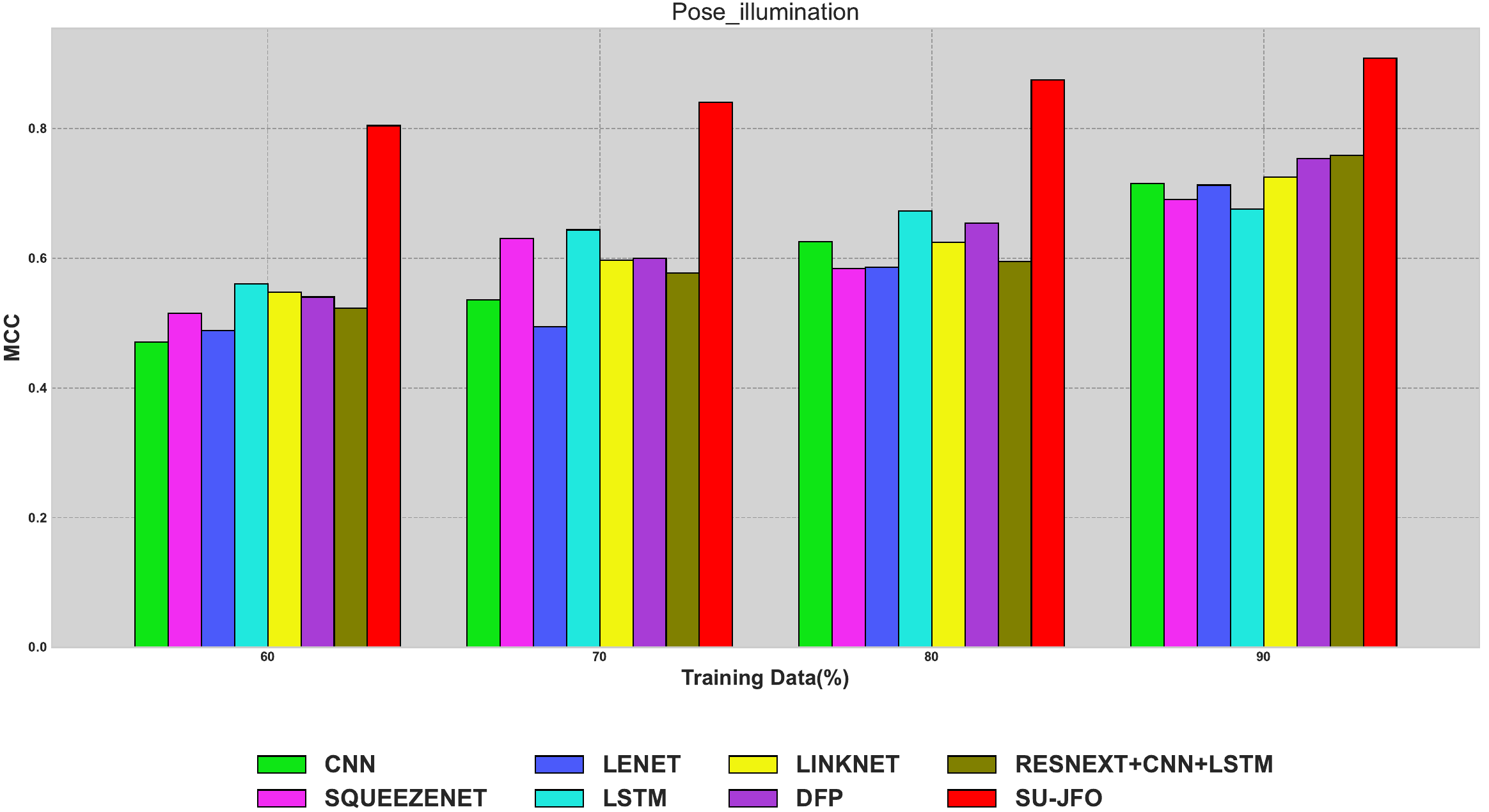}
	\caption{Assessment of SU-JFO and traditional schemes for Pose Illumination case on Dataset1 a) Accuracy b) Precision c) F-measure and d) MCC.}\label{Fig15}
\end{figure*}

In this case, pose illumination has been applied to both real and fake data and a comprehensive evaluation has been conducted to gauge the effectiveness of the SU-JFO technique when compared to conventional methods such as CNN, SqueezeNet, LeNet, LSTM, LinkNet, DFP \cite{raza2022novel}, and ResNext+CNN+LSTM \cite{vamsi2022deepfake}. The comparative analysis depicted in Figure \ref{Fig15} highlights that the SU-JFO methodology demonstrated superior performance in identifying deepfakes in contrast to conventional approaches. The SU-JFO approach demonstrated higher accuracy rates effectively discerning deepfake content. Remarkably, at a training data of 70\% the SU-JFO method showcased a F-measure of 0.928 surpassing the comparatively lower F-measure ratings of CNN, SqueezeNet, LeNet, LSTM, LinkNet, DFP \cite{raza2022novel}, and ResNext+CNN+LSTM \cite{vamsi2022deepfake}\\.   
\vspace{-2mm}

\begin{figure*}[!t]
	\includegraphics[height=9cm,width=\textwidth]{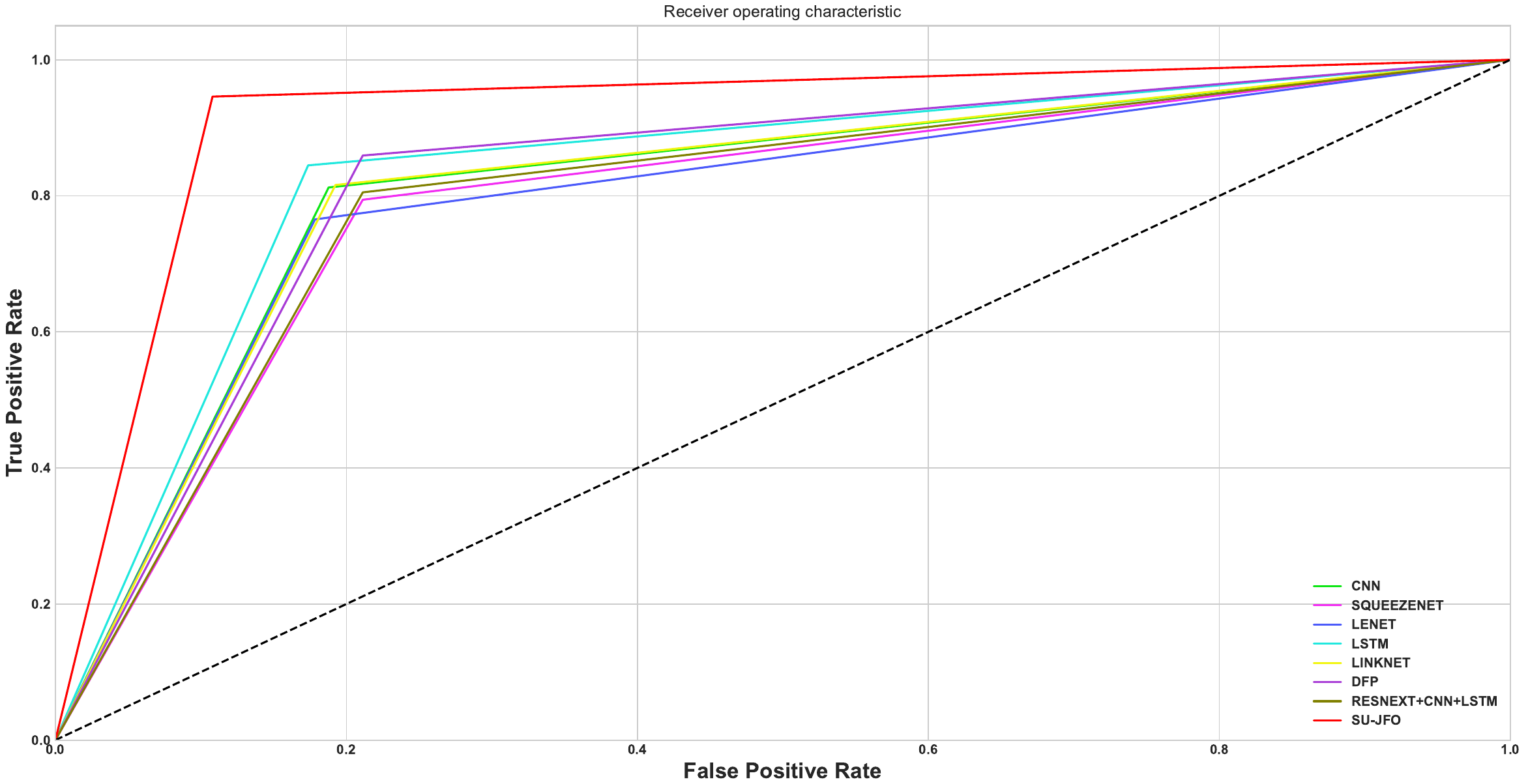}\hfill
	\caption{Evaluation on ROC for Pose Illumination Case on Dataset1.}\label{Fig16}
\end{figure*}

\noindent \textbf{ROC Analysis on Pose Illumination case:}
Figure \ref{Fig16} displays the ROC assessment for both the SU-JFO and traditional methods of deepfake detection. A ROC curve visually illustrates the performance highlighting TPR and FPR. Notably, the TPR achieved with the SU-JFO scheme markedly exceeds that of all other methods.\\   

\noindent \textbf{Statistical Evaluation on Accuracy for Pose Illumination Case:}\\
Table \ref{table4} provides the statistical assessment for both the SU-JFO and traditional approaches to deepfake detection. The model was trained and evaluated 25 times with accuracy measures including mean, maximum, standard deviation, and median calculated.  According to the mean statistical metric, the SU-JFO method attained the highest accuracy rate of 0.929 whereas CNN, SqueezeNet, LeNet, LSTM, LinkNet, DFP \cite{raza2022novel}, and ResNext+CNN+LSTM \cite{vamsi2022deepfake} displayed comparatively lower accuracy values. 
\vspace{-3mm}
\begin{center}
	\begin{table}[!htbp]
		\resizebox{\textwidth}{!}{%
			\begin{tabular}{|l|l|l|l|l|l|l|l|l|}
				\hline
				\begin{tabular}[c]{@{}c@{}}\textbf{Statistical}\\ \textbf{Metrics}\end{tabular} & \textbf{CNN} & \textbf{SqueezeNet}  & \textbf{LeNet}  & \textbf{LSTM}  & \textbf{LinkNet}  & \textbf{DFP}  & \begin{tabular}[c]{@{}c@{}}\textbf{ResNext+CNN+}\\ \textbf{LSTM}\end{tabular}
   & \begin{tabular}[c]{@{}c@{}}\textbf{SU-JFO}\end{tabular}      \\ 
 \hline  
Mean & 0.796 & 0.804 & 0.787 & 0.820 & 0.814 & 0.821 & 0.809 & 0.929\\ \hline
Maximum & 0.861 & 0.845 & 0.857 & 0.839 & 0.865 & 0.878 & 0.882 & 0.955\\ \hline 
Standard Deviation  & 0.046 & 0.031 & 0.044 & 0.023 & 0.033 & 0.039 & 0.044 & 0.019  \\ \hline
Median & 0.792 & 0.805 & 0.771 & 0.829 & 0.808 & 0.816 & 0.795 & 0.930\\ \hline
Minimum & 0.739 & 0.760 & 0.747 & 0.781 & 0.775 & 0.773 & 0.764 & 0.903\\ \hline
			\end{tabular}%
		}
		\caption{Statistical Assessment on Accuracy For Pose Illumination Case on Dataset1.}
		\label{table4}
	\end{table}
\end{center}
\vspace{-10mm}
\subsubsection{Test Case 4: Rotation Case}
\begin{figure*}[!htbp]
	\vspace{-10mm}
\includegraphics[height=5cm,width=\textwidth]{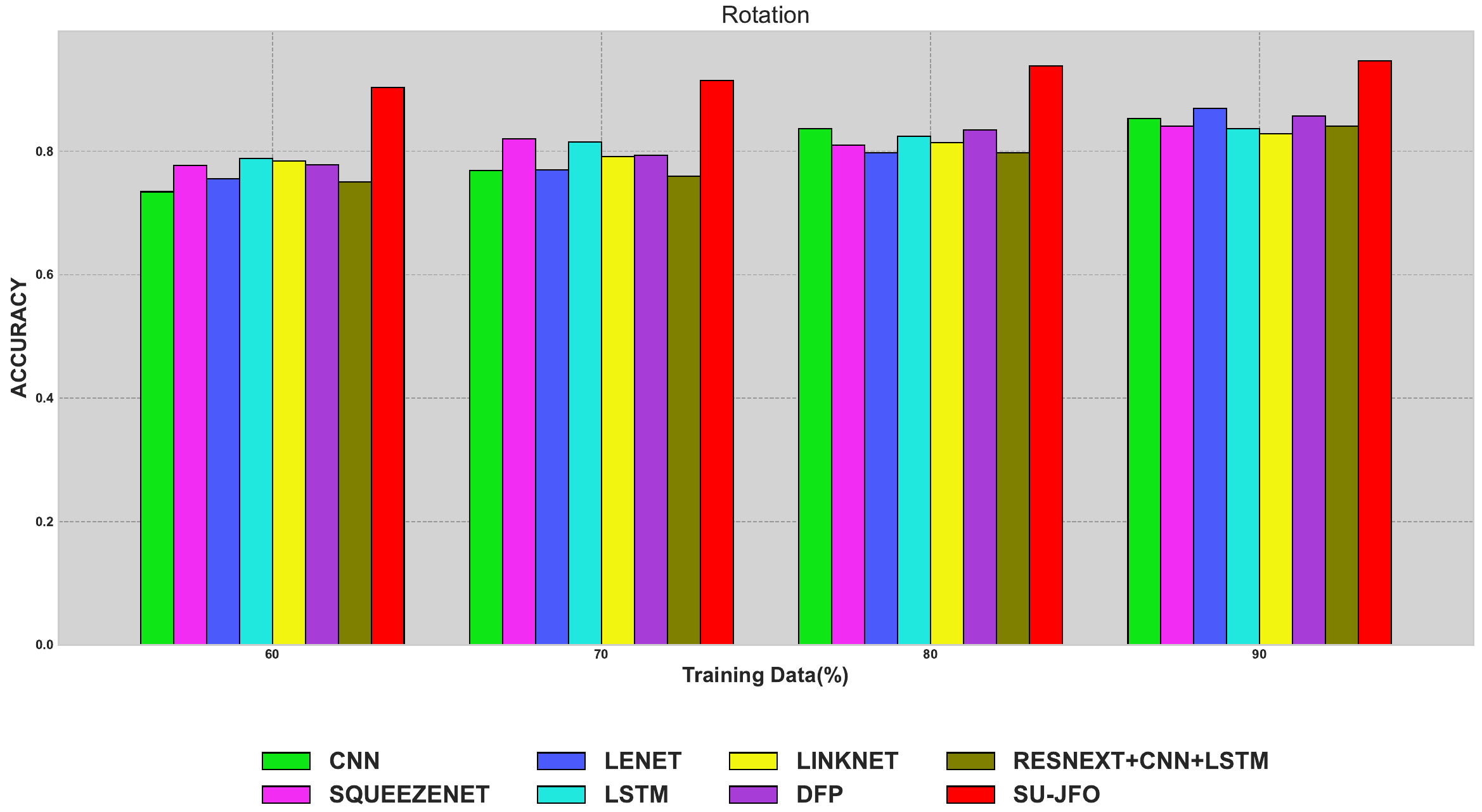}\hfill
\includegraphics[height=5cm,width=\textwidth]{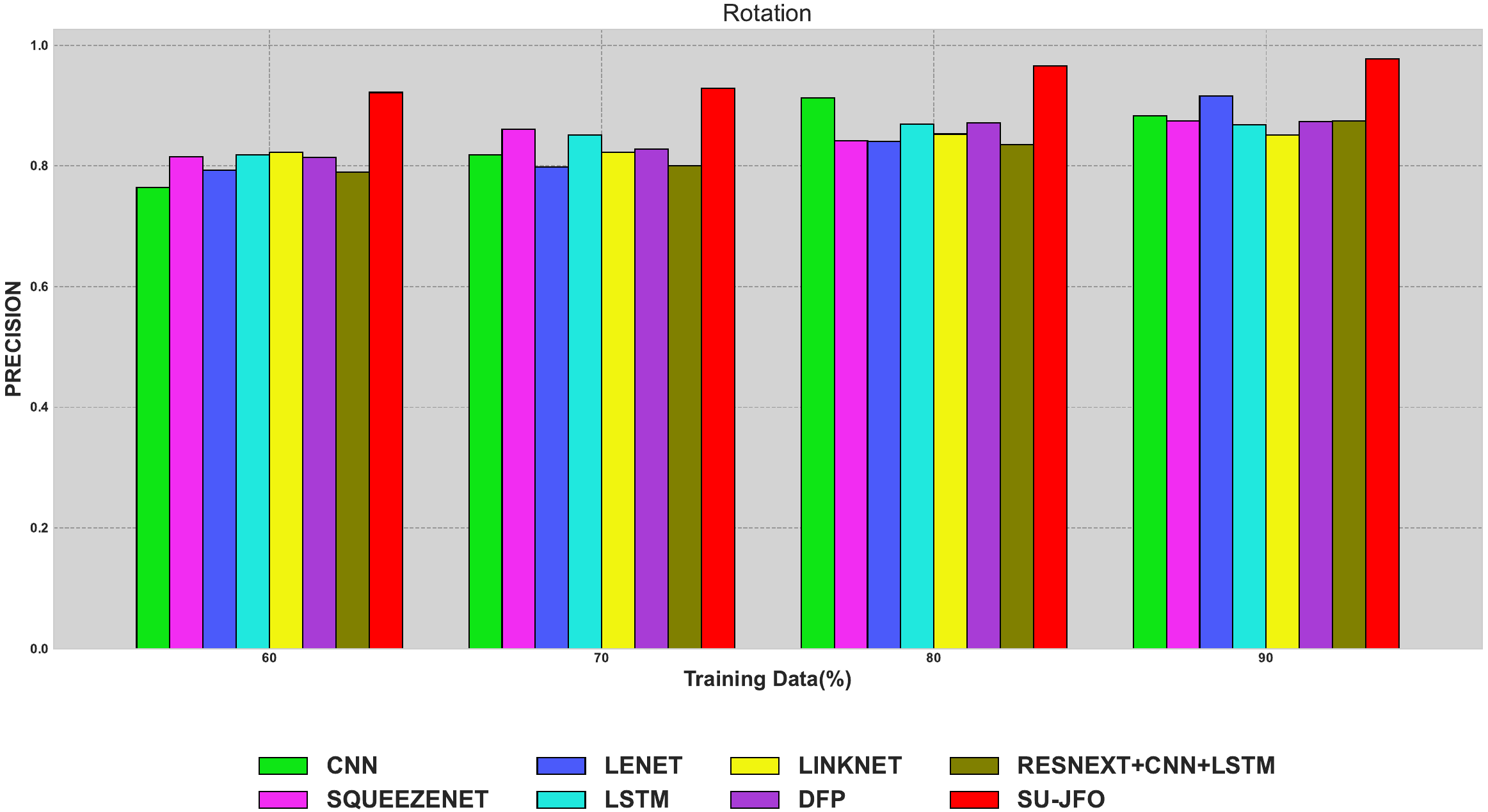}\hfill
\includegraphics[height=5cm,width=\textwidth]{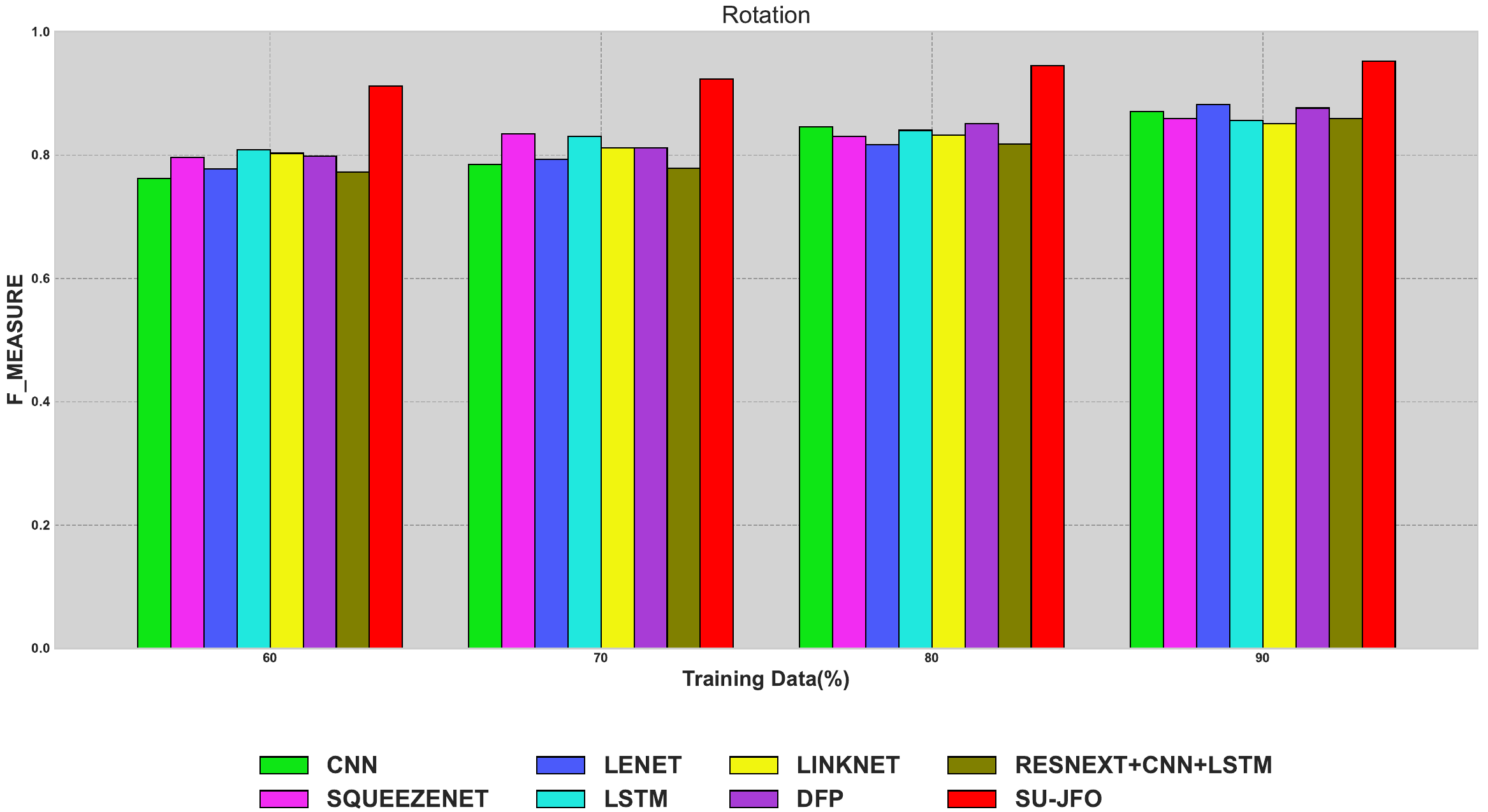}\hfill
\includegraphics[height=5cm,width=\textwidth]{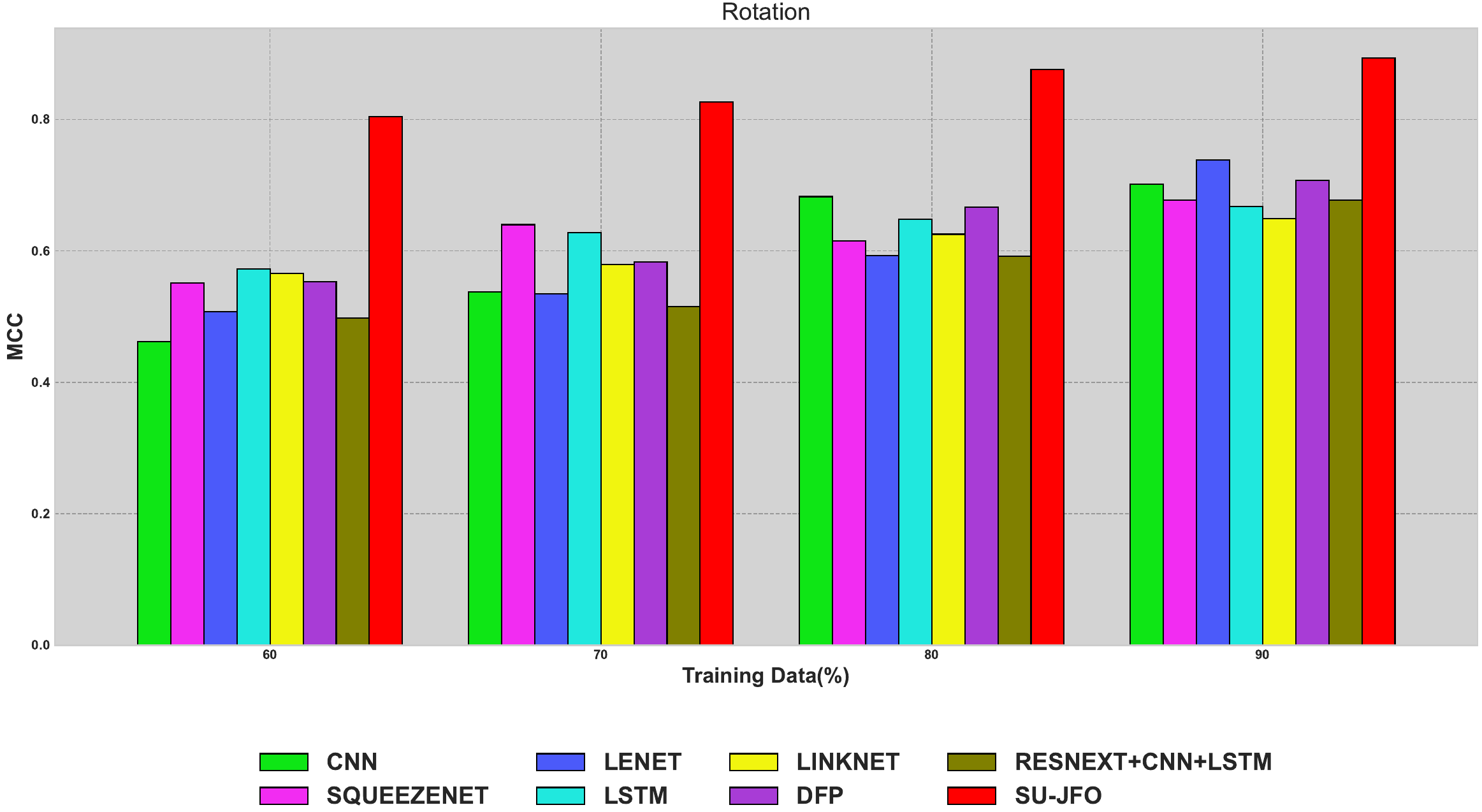}
	\caption{Assessment of SU-JFO and traditional schemes for Rotation case on Dataset1 \\ a) Accuracy b) Precision c) F-measure and d) MCC.}\label{Fig17}
\end{figure*}

In this case, both genuine and synthetic datasets are tested against rotation for analysis. Figure \ref{Fig17} illustrates a comparative investigation between the proposed approach and conventional methods for effective deepfake detection. The evaluation is based on metrics such as accuracy, precision, F-measure, and MCC. From figure \ref{Fig17}, it becomes apparent that our SU-JFO scheme consistently attains higher values compared to existing methods ensuring accurate identification of deepfakes. Notably, all traditional techniques fall below 70\% of MCC whereas our SU-JFO scheme consistently achieves an MCC exceeding 80\%. \\

\begin{figure*}[!t]
	\includegraphics[height=9cm,width=\textwidth]{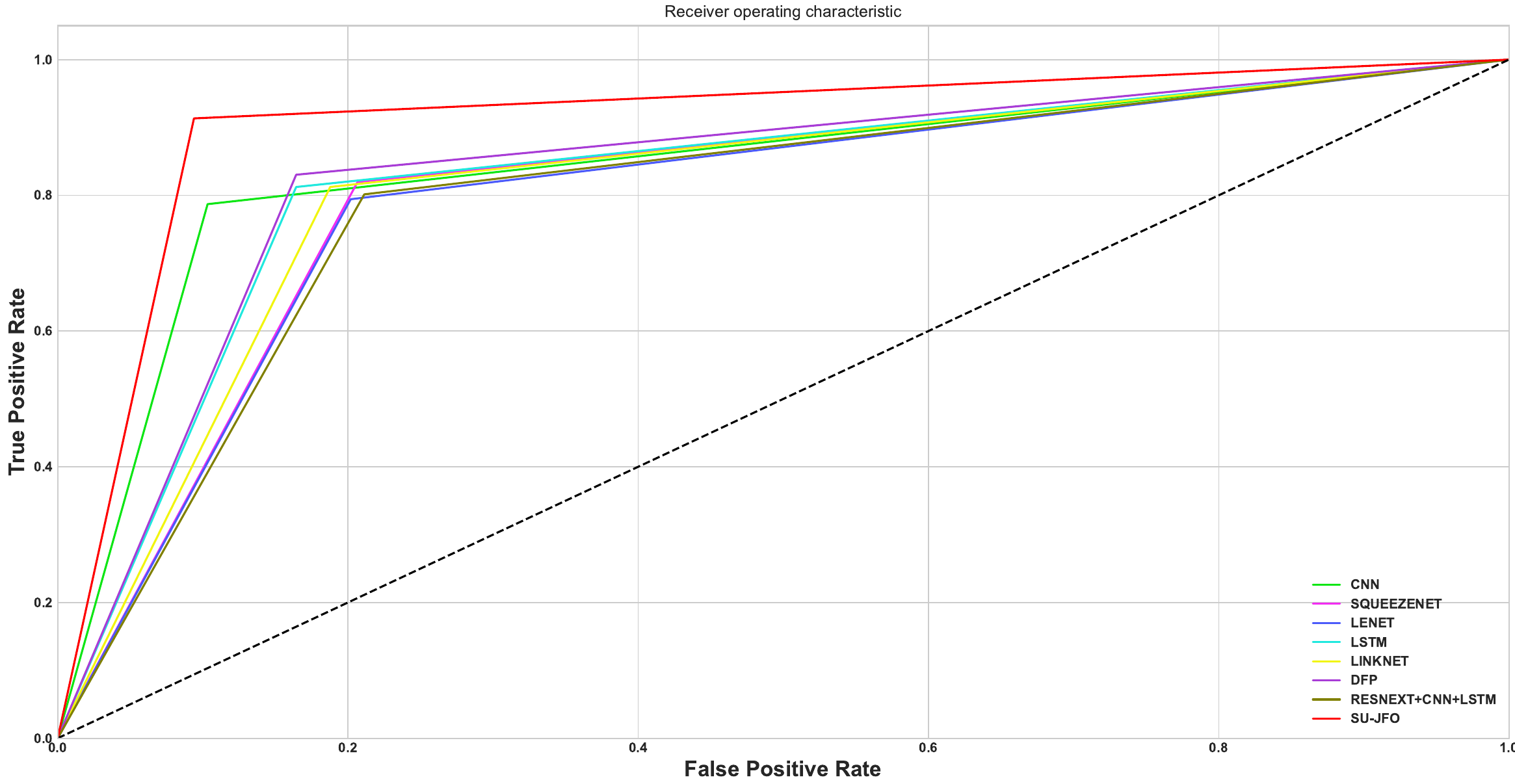}\hfill
	\caption{Evaluation on ROC for Rotation Case on Dataset1.}\label{Fig18}
\end{figure*}

\noindent \textbf{ROC Analysis on Rotation Case:}
For accurate deepfake detection, a ROC curve examination was carried out for SU-JFO and traditional strategies as shown in Figure \ref{Fig18}. In this assessment, both the SU-JFO and conventional approaches achieved true positive rates over 95\%. However, our SU-JFO strategy demonstrated superior performance attaining the highest true positive rate at 0.987 compared to other traditional methods.\\

\noindent \textbf{Statistical evaluation on Accuracy for Rotation Case:}
Table \ref{table5} presents a comparative statistical analysis of the SU-JFO method with conventional strategies for deepfake detection. Across nearly all statistical metrics, the SU-JFO approach achieved higher accuracy ratings when compared to CNN, SqueezeNet, LeNet, LSTM, LinkNet, DFP \cite{raza2022novel}, and ResNext+CNN+LSTM \cite{vamsi2022deepfake}.

\begin{center}
	\begin{table}[!htbp]
		\resizebox{\textwidth}{!}{%
			\begin{tabular}{|l|l|l|l|l|l|l|l|l|}
				\hline
				\begin{tabular}[c]{@{}c@{}}\textbf{Statistical}\\ \textbf{Metrics}\end{tabular} & \textbf{CNN} & \textbf{SqueezeNet}  & \textbf{LeNet}  & \textbf{LSTM}  & \textbf{LinkNet}  & \textbf{DFP}  & \begin{tabular}[c]{@{}c@{}}\textbf{ResNext+CNN+}\\ \textbf{LSTM}\end{tabular}
   & \begin{tabular}[c]{@{}c@{}}\textbf{SU-JFO}\end{tabular}      \\ 
 \hline  
Mean & 0.798 & 0.812 & 0.798 & 0.816 & 0.805 & 0.816 & 0.787 & 0.926\\ \hline
Maximum & 0.853 & 0.841 & 0.869 & 0.837 & 0.829 & 0.857 & 0.841 & 0.947\\ \hline 
Standard Deviation  & 0.049 & 0.023 & 0.044 & 0.018 & 0.018 & 0.032 & 0.036 & 0.018  \\ \hline
Median & 0.803 & 0.815 & 0.784 & 0.820 & 0.803 & 0.814 & 0.779 & 0.926\\ \hline
Minimum & 0.734 & 0.777 & 0.756 & 0.789 & 0.784 & 0.778 & 0.751 & 0.903\\ \hline
			\end{tabular}%
		}
		\caption{Statistical Assessment on Accuracy For Rotation Case on Dataset1}
		\label{table5}
	\end{table}
\end{center}
\vspace{-9mm}

\subsection{Comparative Analysis on Dataset2}\label{section4.7}
\subsubsection{Test case 1: Compression Scenario}
\begin{figure*}[!htbp]
	\vspace{-10mm}
\includegraphics[height=5cm,width=\textwidth]{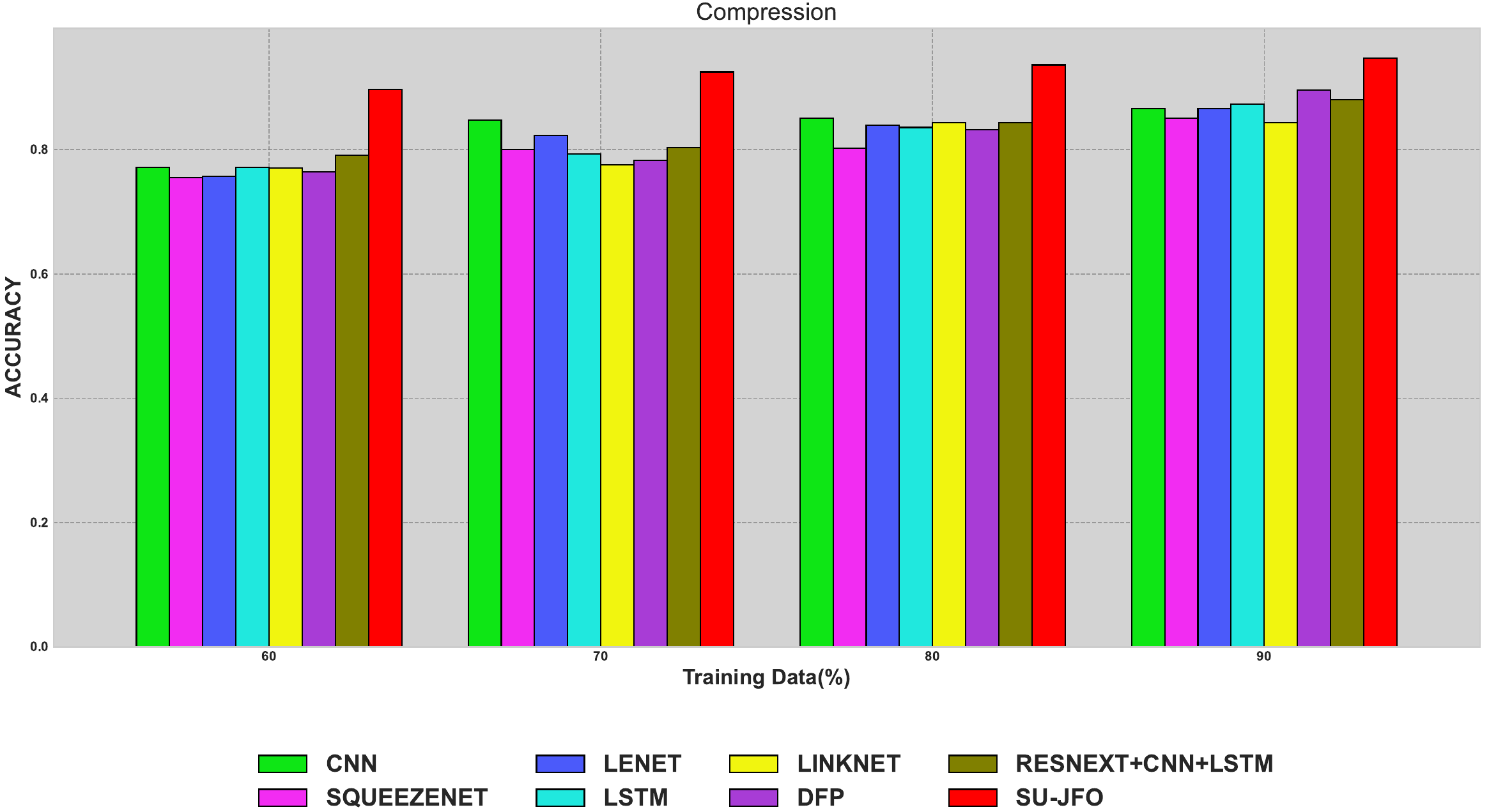}\hfill
\includegraphics[height=5cm,width=\textwidth]{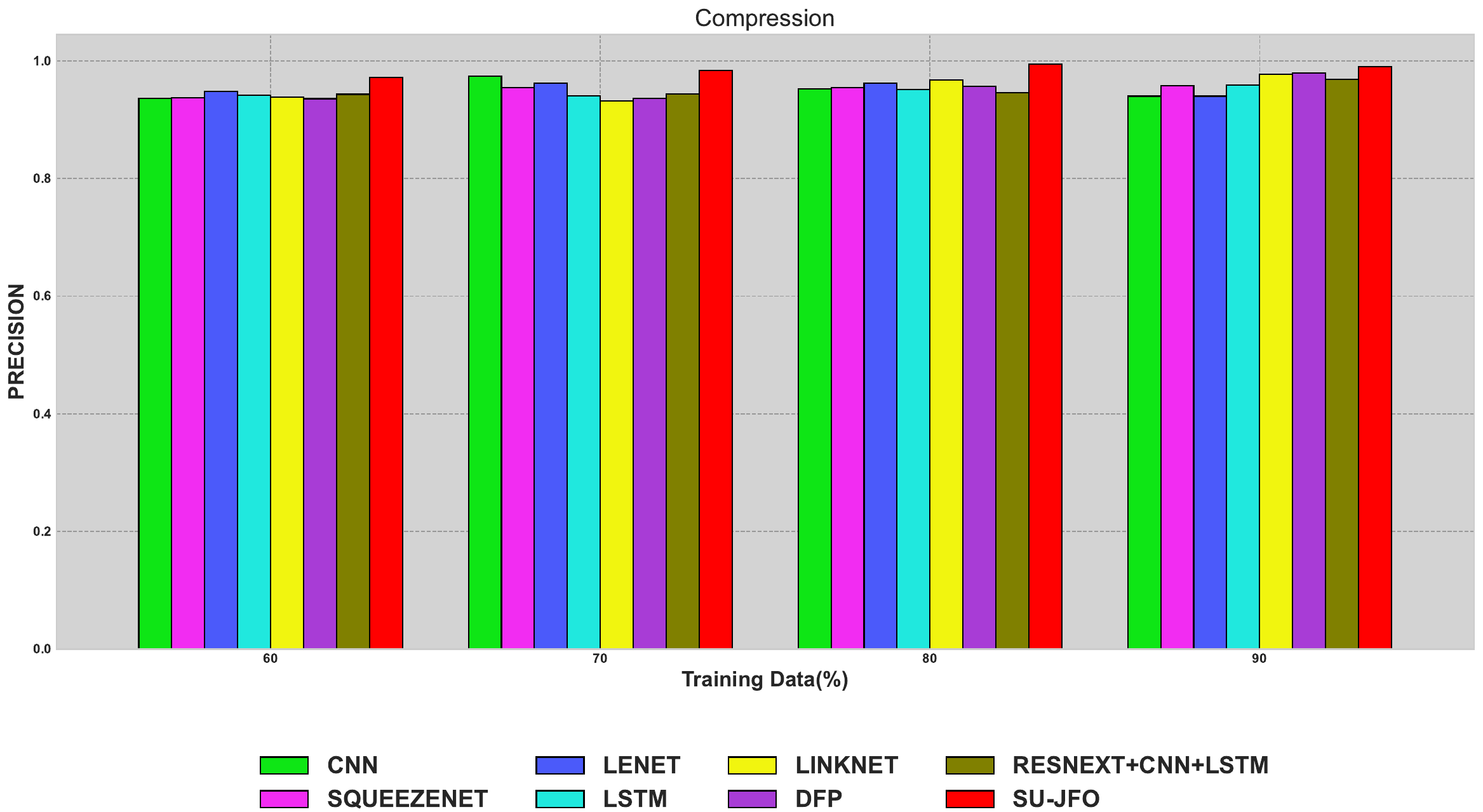}\hfill
\includegraphics[height=5cm,width=\textwidth]{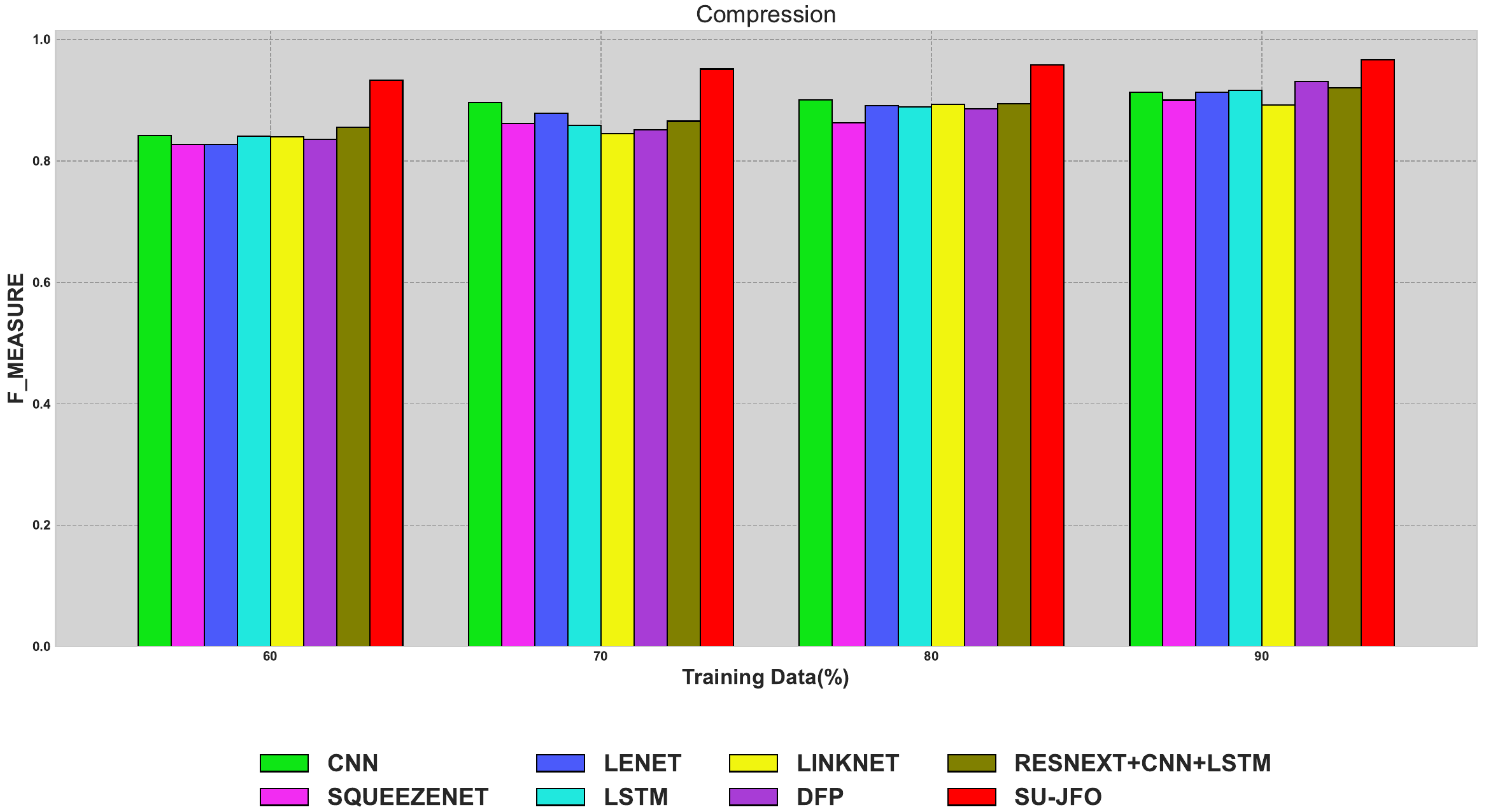}\hfill
\includegraphics[height=5cm,width=\textwidth]{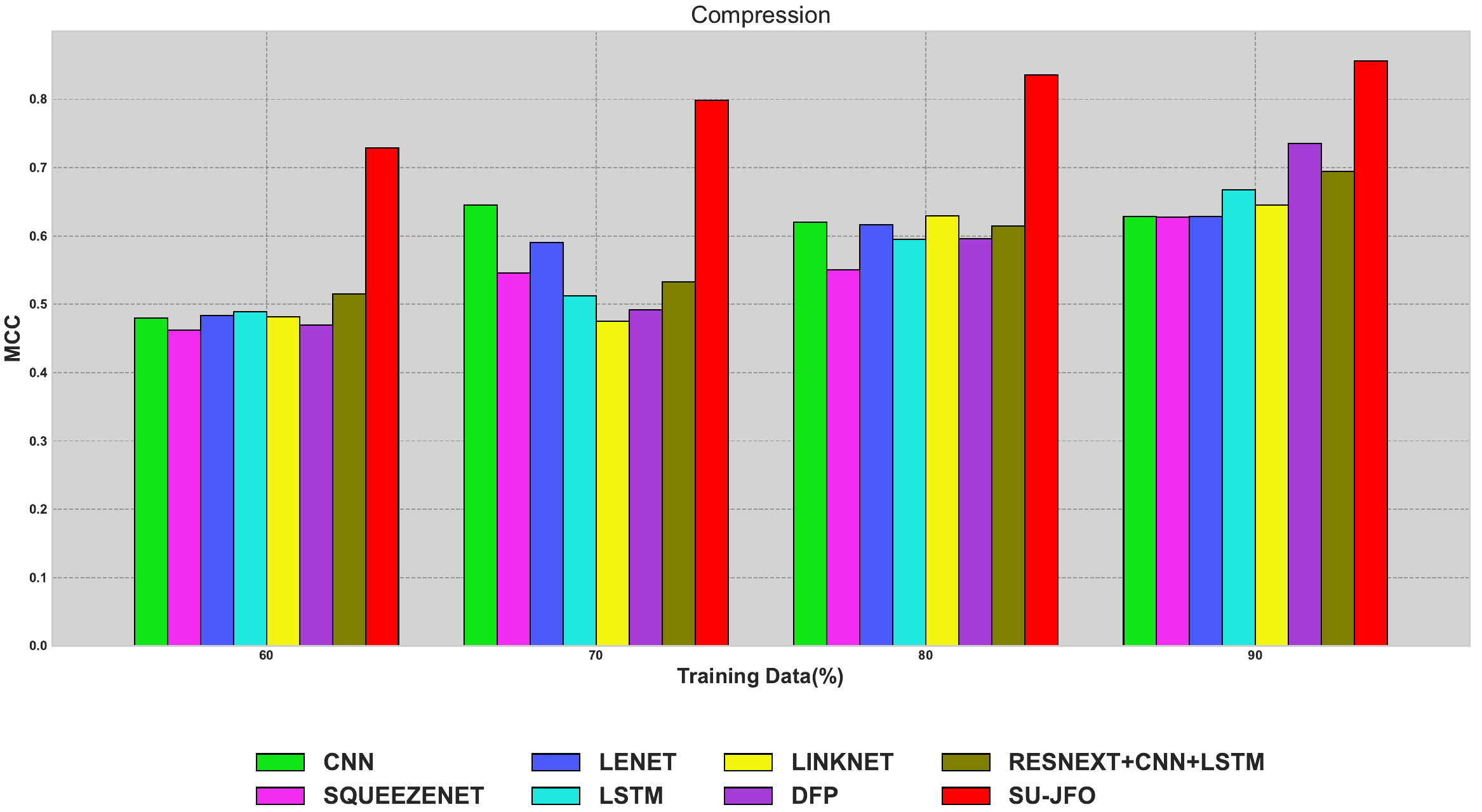}
	\caption{Assessment of SU-JFO and traditional schemes for Compression case on Dataset2 a) Accuracy b) Precision c) F-measure and d) MCC.}\label{Fig19}
\end{figure*}

The assessment of the SU-JFO is compared with evaluations of CNN, SqueezeNet, LeNet, LSTM, LinkNet, DFP \cite{raza2022novel}, and ResNext+CNN+LSTM \cite{vamsi2022deepfake} for deepfake detection as depicted in Figure \ref{Fig19}. Precision, MCC, Accuracy, and F-measure are the key metrics considered in this evaluation. It can be observed that SU-JFO method consistently outperformed prior approaches. Notably, both the SU-JFO and traditional methods exhibited precision values exceeding 90\%. However, the SU-JFO strategy outperformed with a precision value of 0.971 significantly than that of the traditional methods. 
\vspace{2mm}

\begin{figure*}[!t]
	\includegraphics[height=9cm,width=\textwidth]{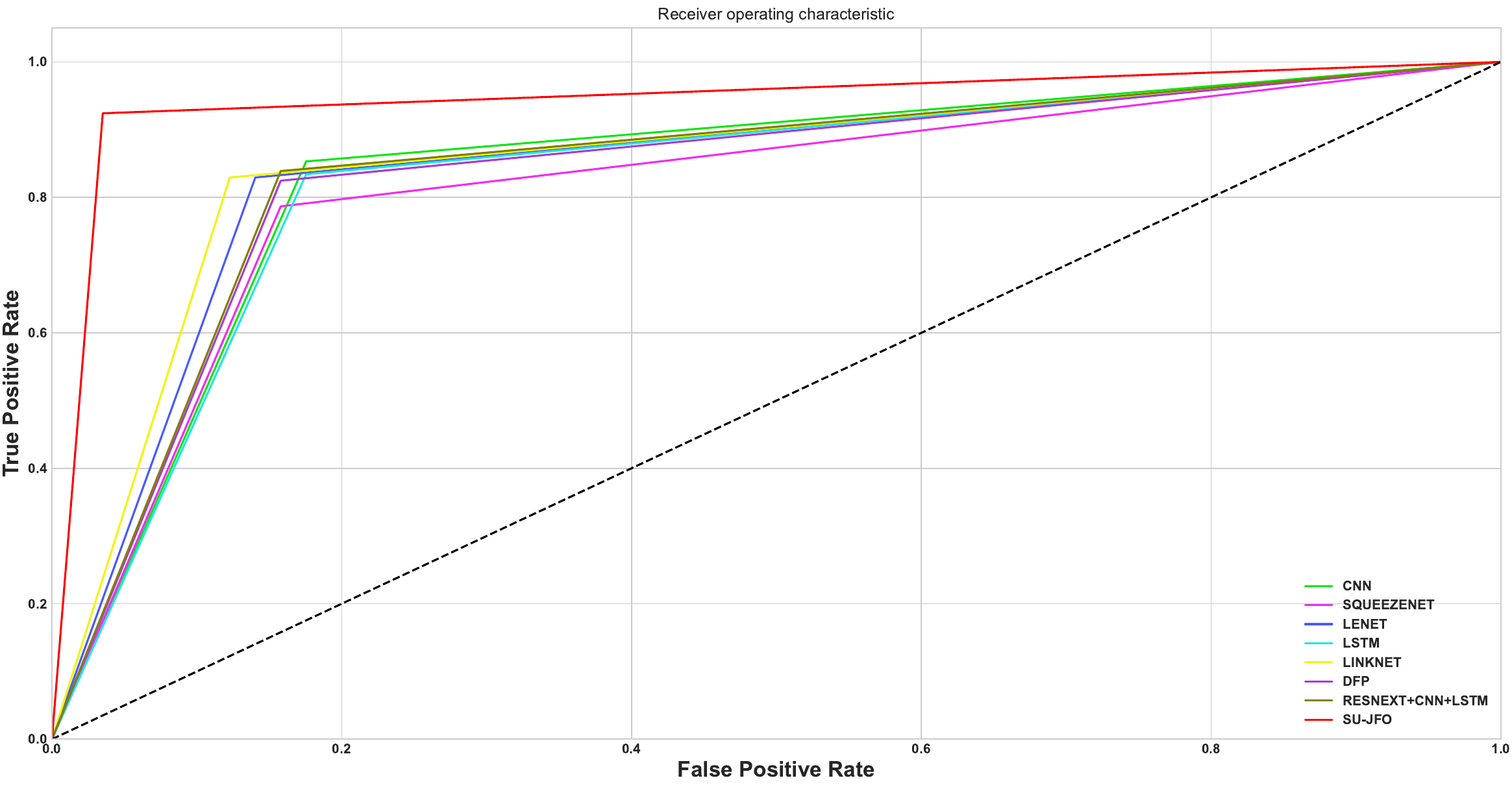}\hfill
	\caption{Evaluation on ROC for Compression Case on Dataset2.}\label{Fig20}
\end{figure*}

\noindent \textbf{ROC Analysis on Compression Case:}
In Figure \ref{Fig20}, the ROC assessment compares the SU-JFO method with CNN, SqueezeNet, LeNet, LSTM, LinkNet, DFP \cite{raza2022novel}, and ResNext+CNN+LSTM \cite{vamsi2022deepfake} for deepfake detection. Achieving ROC values greater than 95\% is crucial for effective deepfake detection. In this analysis, all algorithms surpassed the 95\% threshold when reaching a false positive rate of 0.8. Notably, our SU-JFO scheme outperforms DFP \cite{raza2022novel} and ResNext+CNN+LSTM \cite{vamsi2022deepfake} demonstrating the highest ROC values in this context.\\ 

\noindent \textbf{Statistical Analysis on Accuracy for Compression Case:}
Table \ref{table6} summarizes the statistical comparison between the SU-JFO and conventional strategies for deepfake detection. The model was trained and evaluated 25 times and the SU-JFO method achieved an accuracy score of 0.897 surpassing the minimum accuracy values obtained by CNN (0.772), SqueezeNet (0.755), LeNet (0.757), LSTM (0.772), LinkNet (0.770), DFP \cite{raza2022novel} (0.764), and ResNext+CNN+LSTM \cite{vamsi2022deepfake}  (0.791).

\begin{center}
	\begin{table}[!htbp]
		\resizebox{\textwidth}{!}{%
			\begin{tabular}{|l|l|l|l|l|l|l|l|l|}
				\hline
				\begin{tabular}[c]{@{}c@{}}\textbf{Statistical}\\ \textbf{Metrics}\end{tabular} & \textbf{CNN} & \textbf{SqueezeNet}  & \textbf{LeNet}  & \textbf{LSTM}  & \textbf{LinkNet}  & \textbf{DFP}  & \begin{tabular}[c]{@{}c@{}}\textbf{ResNext+CNN+}\\ \textbf{LSTM}\end{tabular}
   & \begin{tabular}[c]{@{}c@{}}\textbf{SU-JFO}\end{tabular}      \\ 
 \hline  
Mean & 0.834 & 0.802 & 0.821 & 0.818 & 0.808 & 0.819 & 0.829 & 0.927\\ \hline
Maximum & 0.866 & 0.851 & 0.866 & 0.873 & 0.843 & 0.896 & 0.881 & 0.948\\ \hline 
Standard Deviation  & 0.036 & 0.034 & 0.040 & 0.039 & 0.035 & 0.051 & 0.035 & 0.019  \\ \hline
Median & 0.849 & 0.801 & 0.831 & 0.814 & 0.809 & 0.808 & 0.823 & 0.931\\ \hline
Minimum & 0.772 & 0.755 & 0.757 & 0.772 & 0.770 & 0.764 & 0.791 & 0.897\\ \hline
			\end{tabular}%
		}
		\caption{Statistical Assessment on Accuracy For Compression Case on Dataset2.}
		\label{table6}
	\end{table}
\end{center}
\vspace{-8mm}

\subsubsection{Test Case 2: Noise Scenario}
\begin{figure*}[!htbp]
	\vspace{-10mm}
\includegraphics[height=5cm,width=\textwidth]{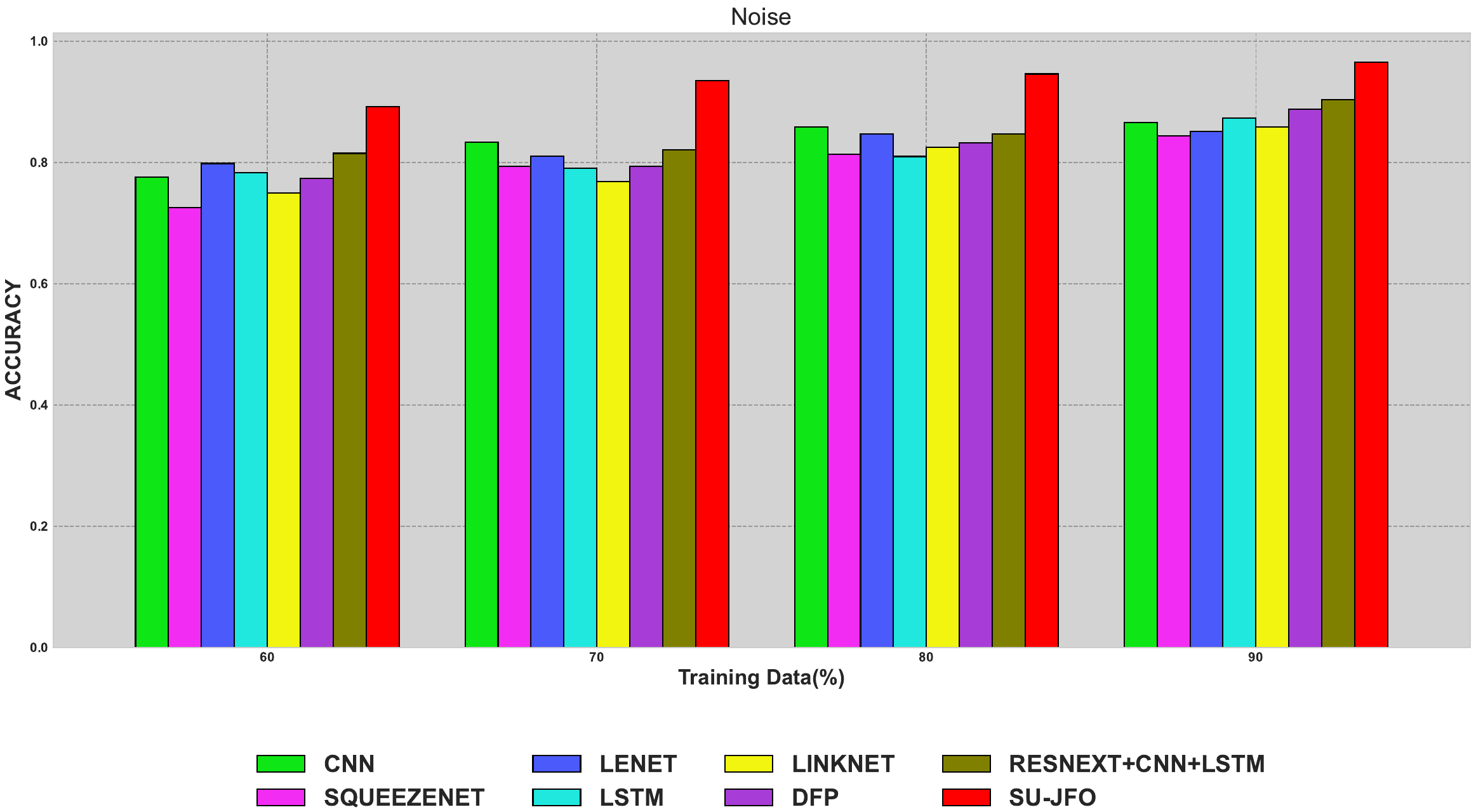}\hfill
\includegraphics[height=5cm,width=\textwidth]{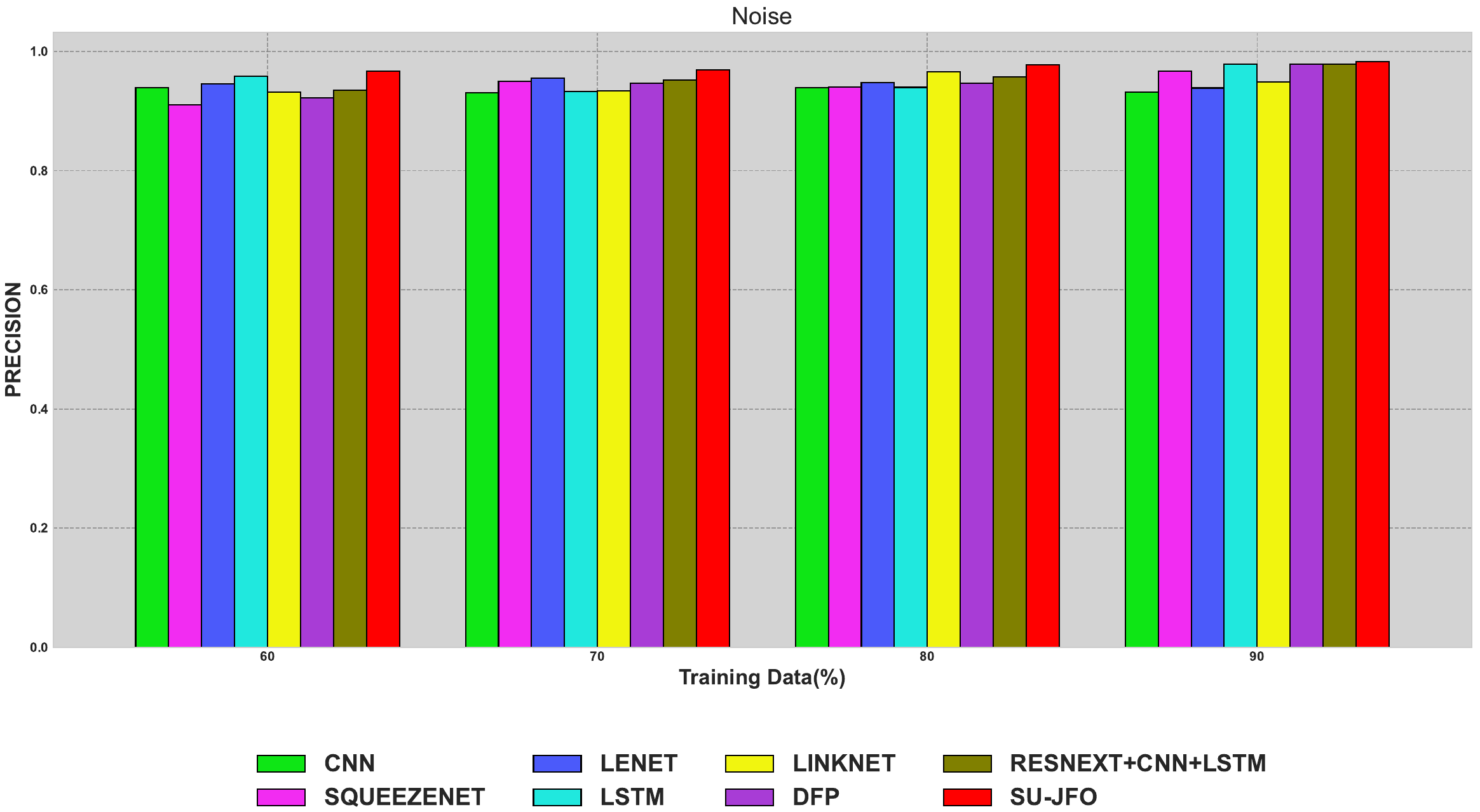}\hfill
\includegraphics[height=5cm,width=\textwidth]{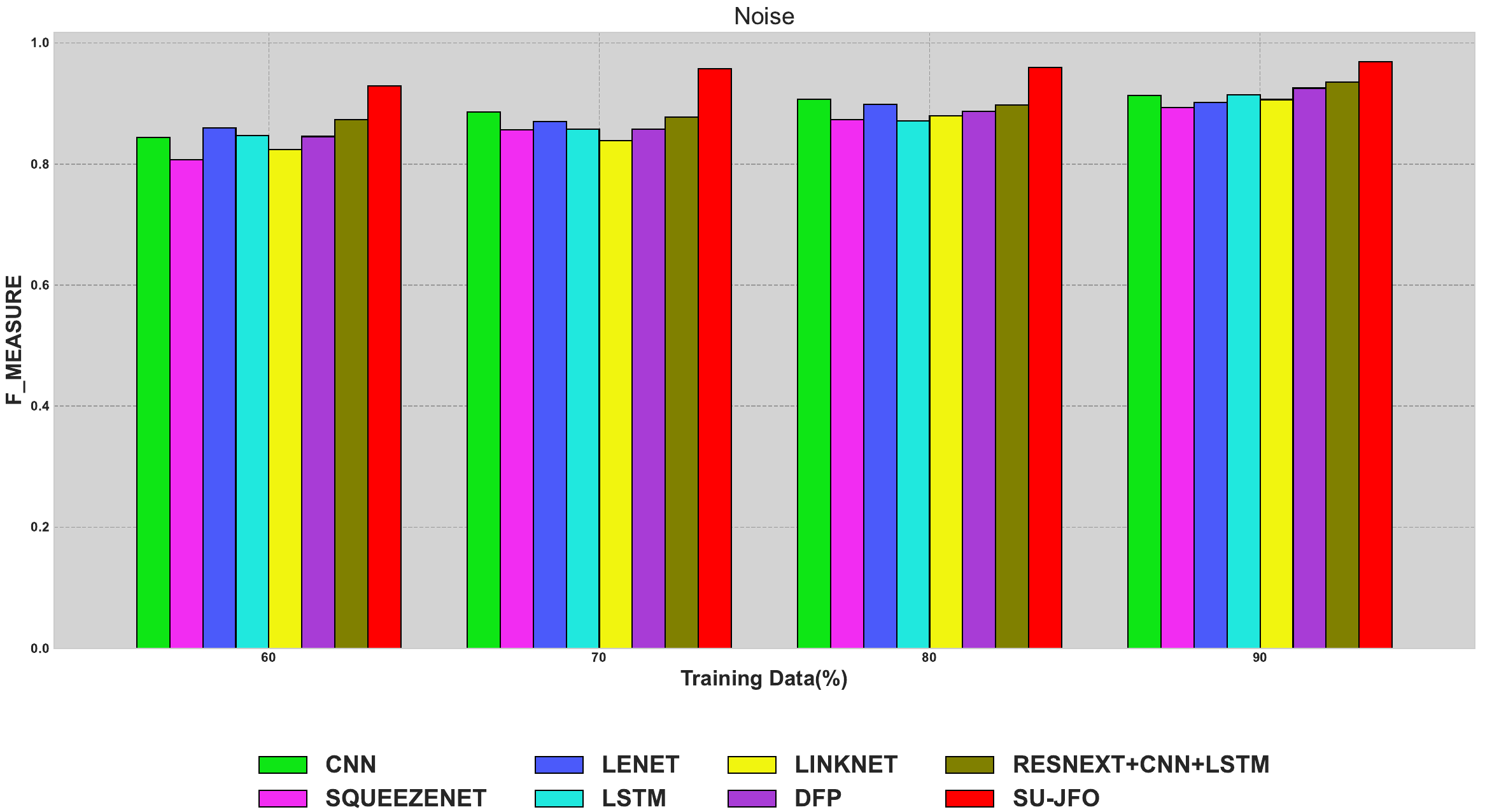}\hfill
\includegraphics[height=5cm,width=\textwidth]{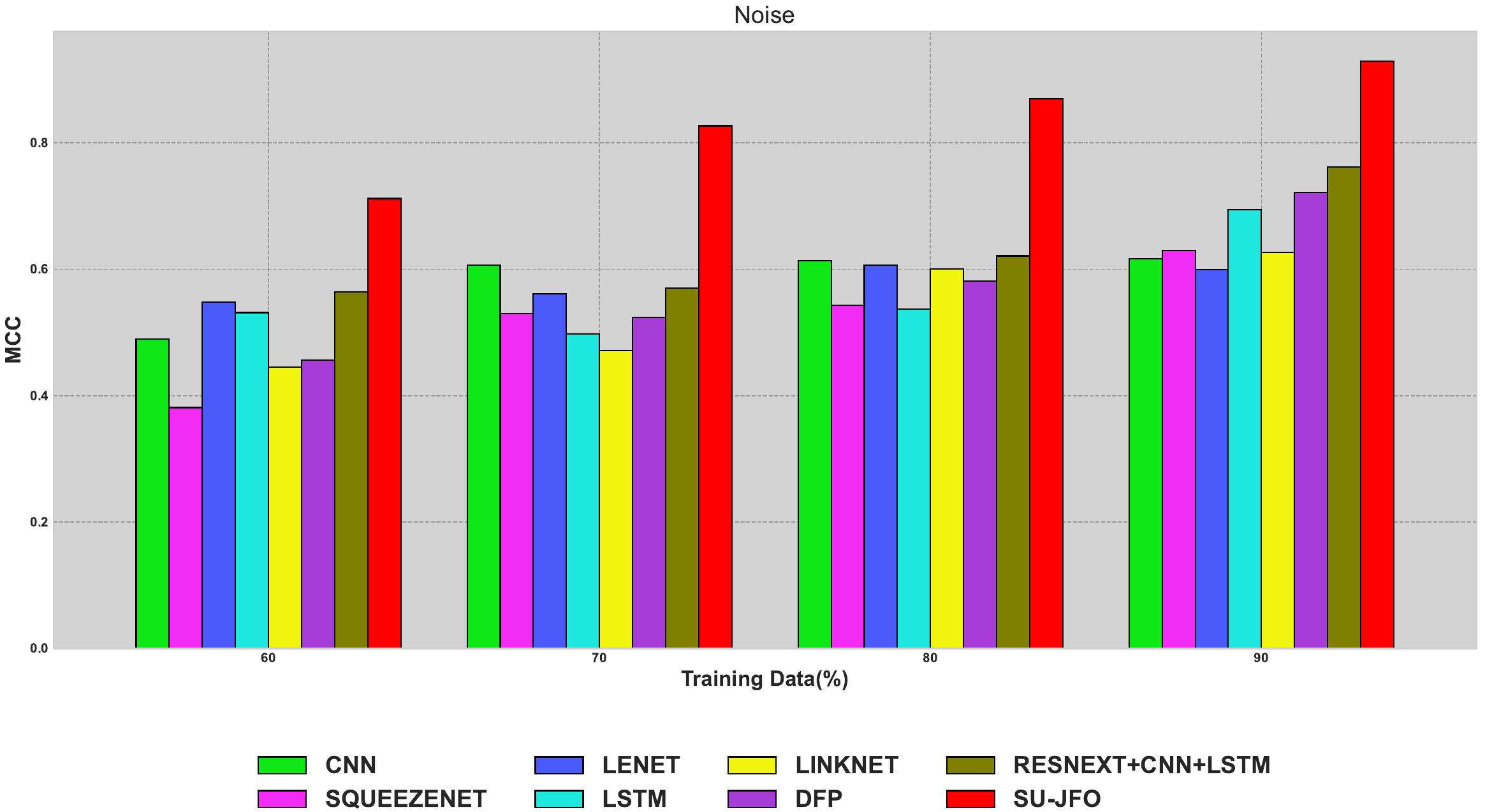}
	\caption{Assessment of SU-JFO and traditional schemes for Noise case on Dataset2 \\a) Accuracy b) Precision c) F-measure and d) MCC.}\label{Fig21}
\end{figure*}

To estimate the efficacy of the SU-JFO-based deepfake detection approach, a thorough evaluation was conducted encompassing the analysis of various metrics including Accuracy, F-measure, MCC, and Precision across diverse sets of training data. The outcomes of this assessment are illustrated in Figure \ref{Fig21}. Furthermore, the suggested model undergoes a comparative examination alongside other models such as CNN, SqueezeNet, LeNet, LSTM, LinkNet, DFP \cite{raza2022novel}, and ResNext+CNN+LSTM \cite{vamsi2022deepfake}. In this scenario, the SU-JFO method demonstrated superior outcomes in contrast to conventional strategies. As evident from Figure 20(c), the SU-JFO surpassed the existing approaches by accomplishing a maximal F-measure with a training data of 60\%. 

\begin{figure*}[!t]
	\includegraphics[height=9cm,width=\textwidth]{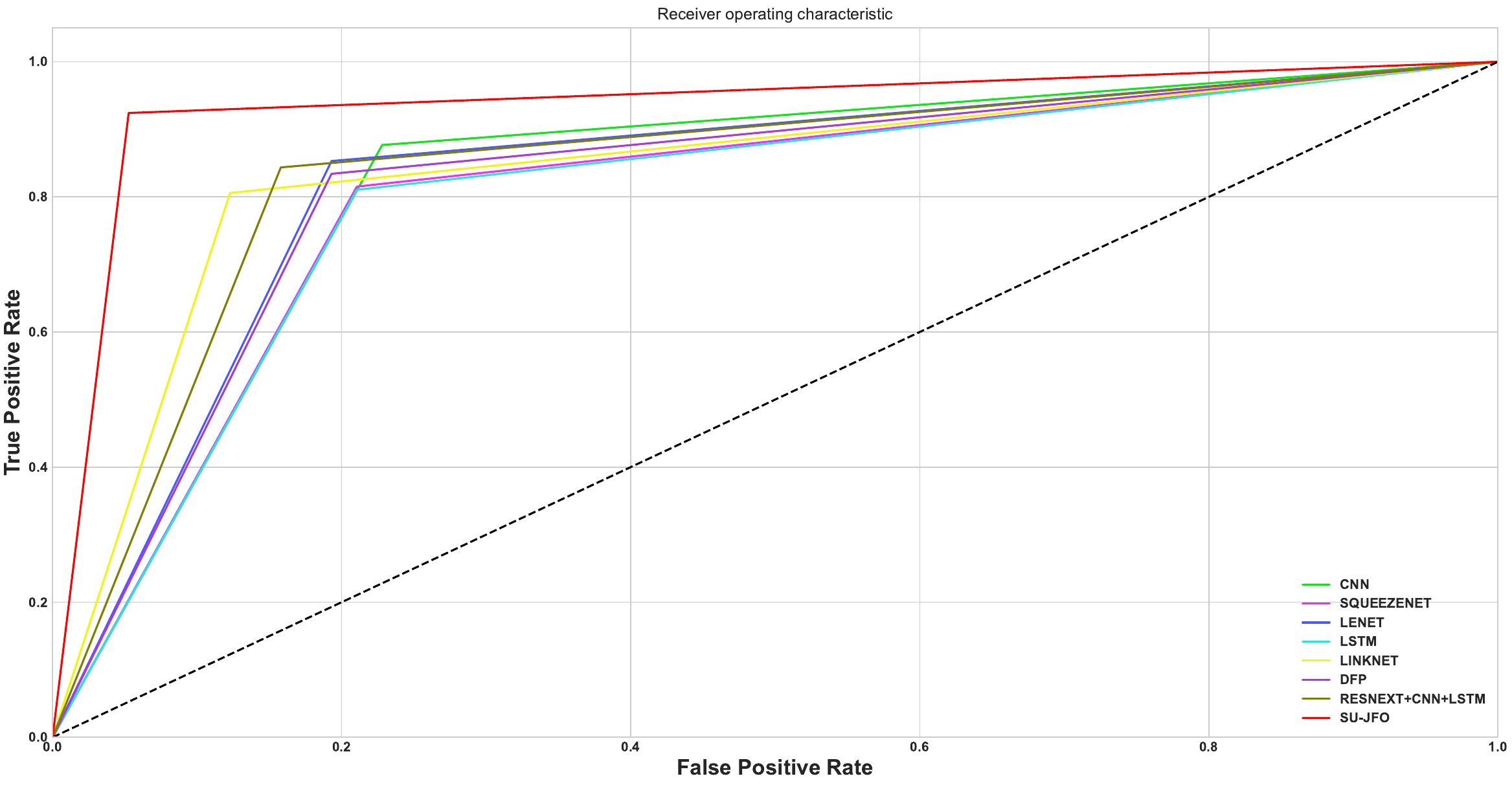}\hfill
	\caption{Evaluation on ROC for Noise case on Dataset2.}\label{Fig22}
\end{figure*}

\noindent \textbf{ROC Analysis on Noise Case:}
Figure \ref{Fig22} presents the ROC curve analysis contrasting the SU-JFO method with conventional approaches like CNN, SqueezeNet, LeNet, LSTM, LinkNet, DFP \cite{raza2022novel}, and ResNext+CNN+LSTM \cite{vamsi2022deepfake}. The ROC curve was created by plotting the TPR against FPR using a dataset that included 70\% of the training data. Achieving an ROC area of 95\% or higher is crucial to ensure the precision of deepfake detection. In line with this objective, our SU-JFO achieved TPR surpassing 0.992 demonstrating a notable superiority over true positive rates attained by traditional methodologies. 

\vspace{2mm}
\noindent \textbf{Statistical Analysis on Accuracy on Noise Case:}
Table \ref{table7} presents a statistical evaluation comparing the effectiveness of the SU-JFO method with traditional approaches in deepfake detection. Upon examination, the SU-JFO  method consistently demonstrated higher accuracy scores across most of the statistical metrics whereas the conventional methods consistently showed lower accuracy values. 

\begin{center}
	\begin{table}[!htbp]
		\resizebox{\textwidth}{!}{%
			\begin{tabular}{|l|l|l|l|l|l|l|l|l|}
				\hline
				\begin{tabular}[c]{@{}c@{}}\textbf{Statistical}\\ \textbf{Metrics}\end{tabular} & \textbf{CNN} & \textbf{SqueezeNet}  & \textbf{LeNet}  & \textbf{LSTM}  & \textbf{LinkNet}  & \textbf{DFP}  & \begin{tabular}[c]{@{}c@{}}\textbf{ResNext+CNN+}\\ \textbf{LSTM}\end{tabular}
   & \begin{tabular}[c]{@{}c@{}}\textbf{SU-JFO}\end{tabular}      \\ 
 \hline  
Mean & 0.833 & 0.794 & 0.827 & 0.814 & 0.800 & 0.822 & 0.846 & 0.934\\ \hline
Maximum & 0.866 & 0.843 & 0.851 & 0.873 & 0.858 & 0.888 & 0.903 & 0.965\\ \hline 
Standard Deviation  & 0.035 & 0.043 & 0.023 & 0.035 & 0.043 & 0.044 & 0.035 & 0.027  \\ \hline
Median & 0.846 & 0.803 & 0.829 & 0.800 & 0.796 & 0.813 & 0.834 & 0.940\\ \hline
Minimum & 0.776 & 0.725 & 0.798 & 0.783 & 0.750 & 0.774 & 0.815 & 0.892\\ \hline
			\end{tabular}%
		}
		\caption{Statistical Assessment on Accuracy For Noise Case on Dataset2.}
		\label{table7}
	\end{table}
\end{center}
\vspace{-8mm}

\subsubsection{Test Case 3: Pose Illumination Scenario}
\begin{figure*}[!htbp]
	\vspace{-10mm}
\includegraphics[height=5cm,width=\textwidth]{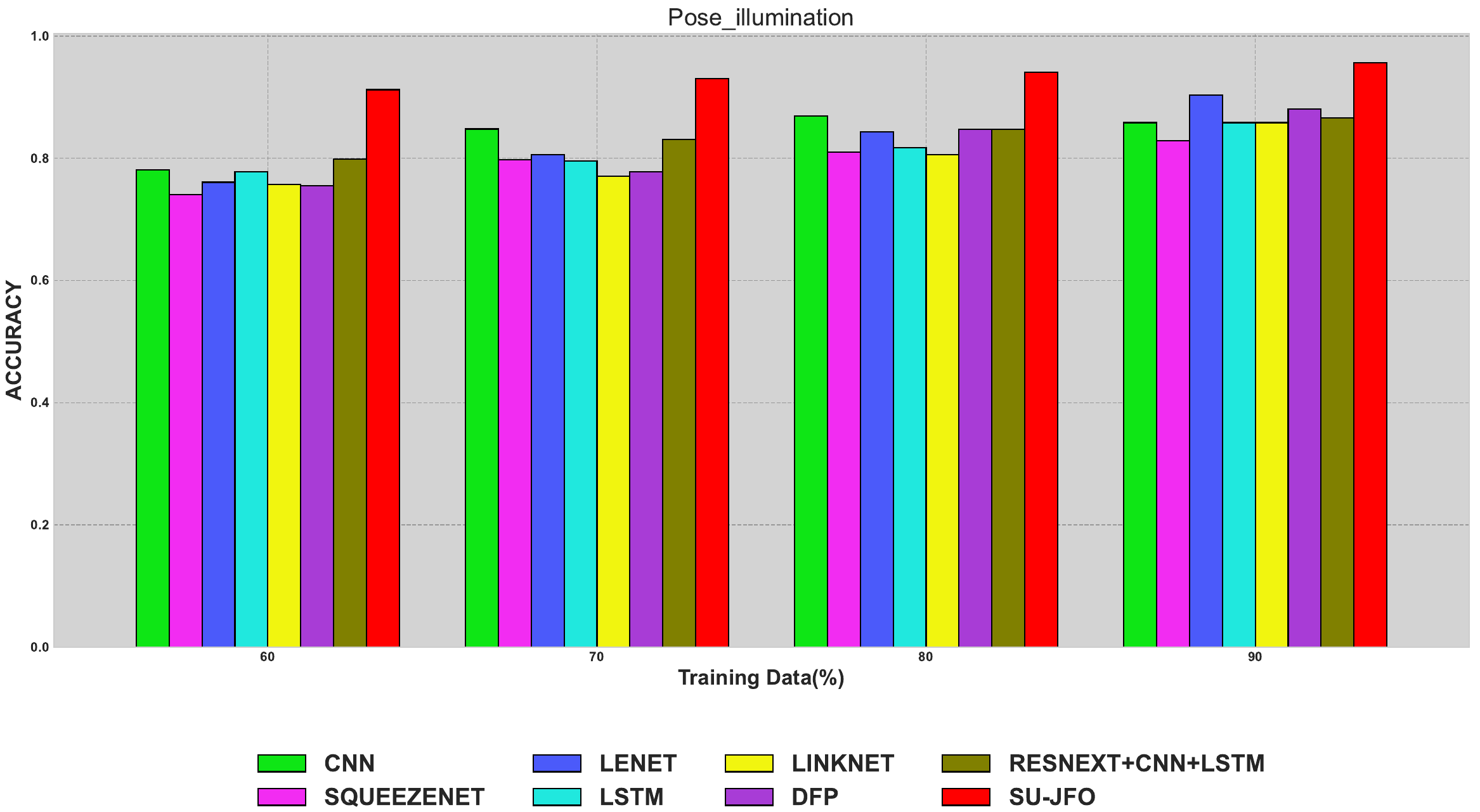}\hfill
\includegraphics[height=5cm,width=\textwidth]{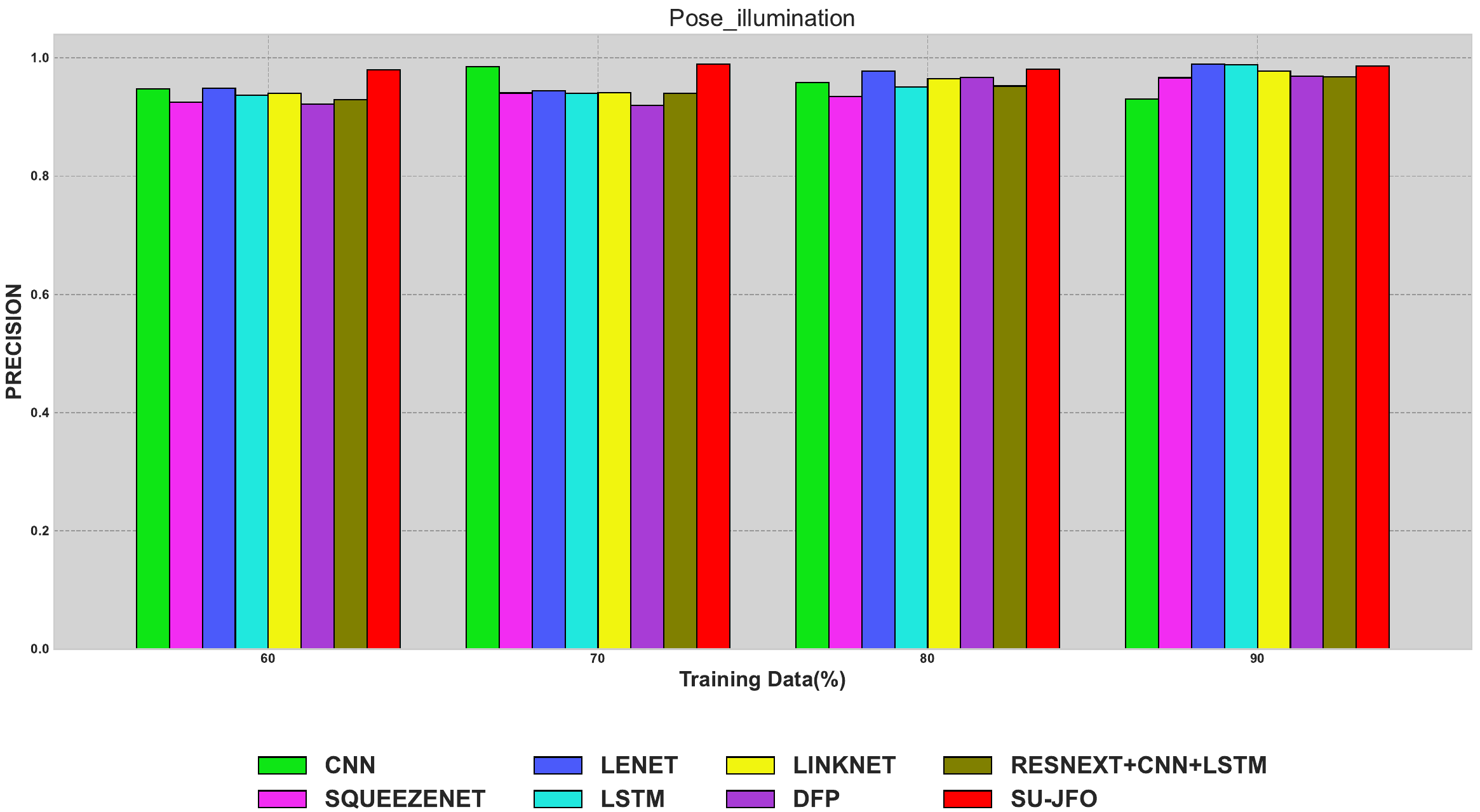}\hfill
\includegraphics[height=5cm,width=\textwidth]{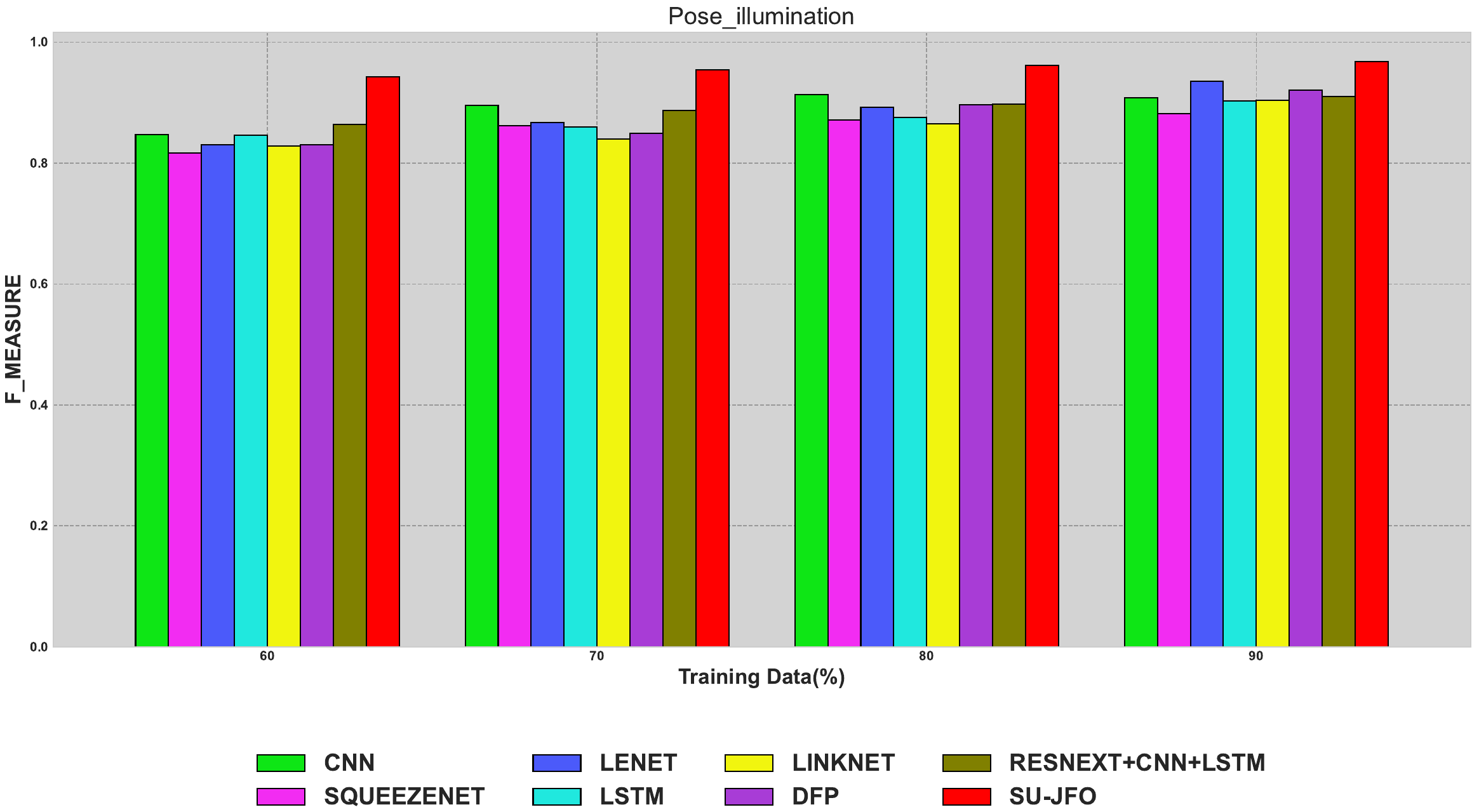}\hfill
\includegraphics[height=5cm,width=\textwidth]{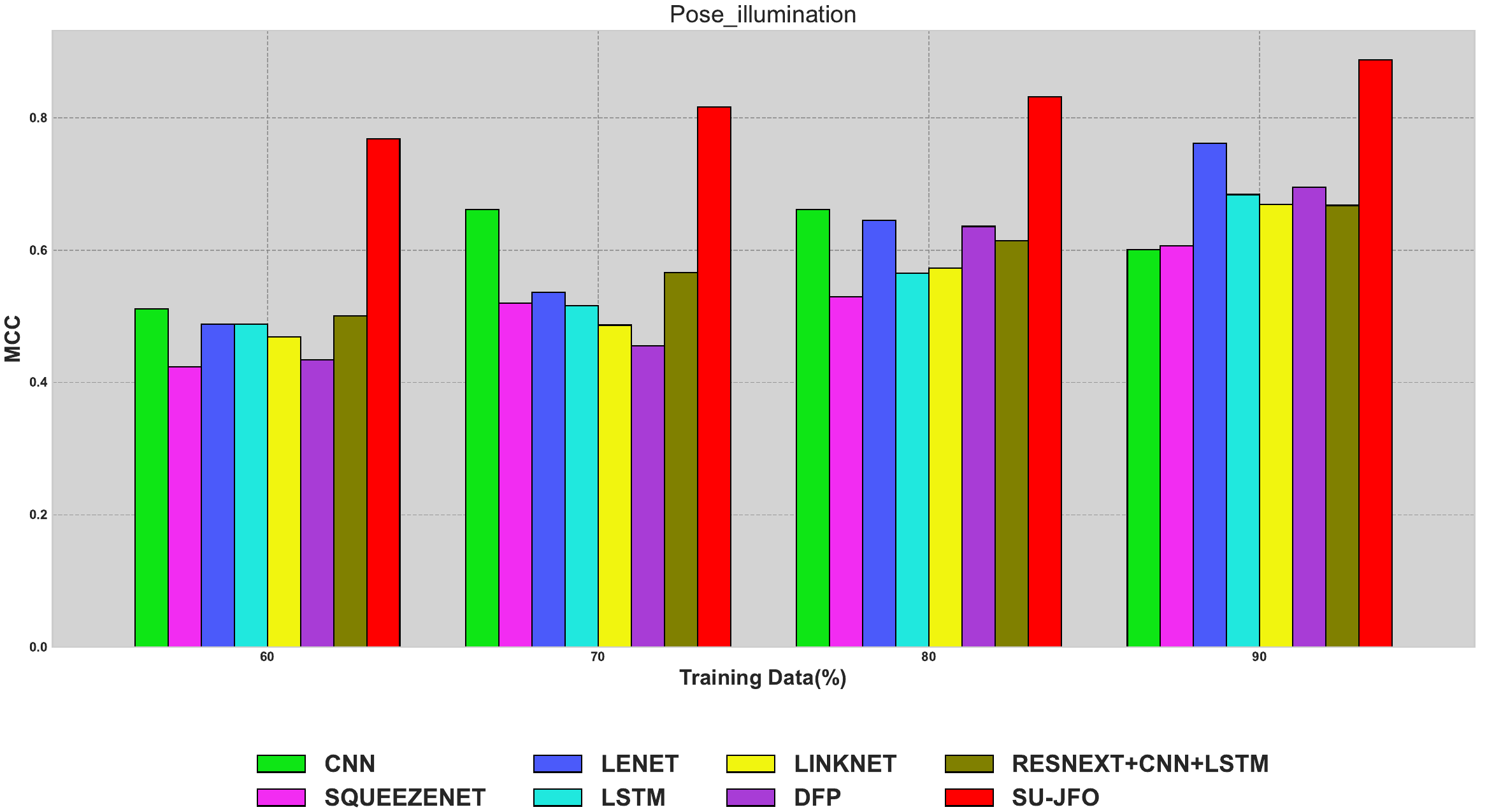}
	\caption{Assessment of SU-JFO and traditional schemes for Pose Illumination case on Dataset2 a) Accuracy b) Precision c) F-measure and d) MCC.}\label{Fig23}
\end{figure*}

Figure \ref{Fig23} illustrates the graphical depiction of the efficacy of the SU-JFO methodology in detecting deepfakes using diverse training data. The findings indicate that when 90\% of the training data is utilized the model demonstrates enhanced performance in deepfake detection achieving an F-measure rate surpassing 94\%. Conversely, traditional models face challenges in reaching F-measure levels surpassing 85\%. Moreover, across all training data scenarios the SU-JFO approach consistently outperforms conventional methods reflected in higher metric values. 
\vspace{2mm}
\\

\begin{figure*}[!t]
	\includegraphics[height=9cm,width=\textwidth]{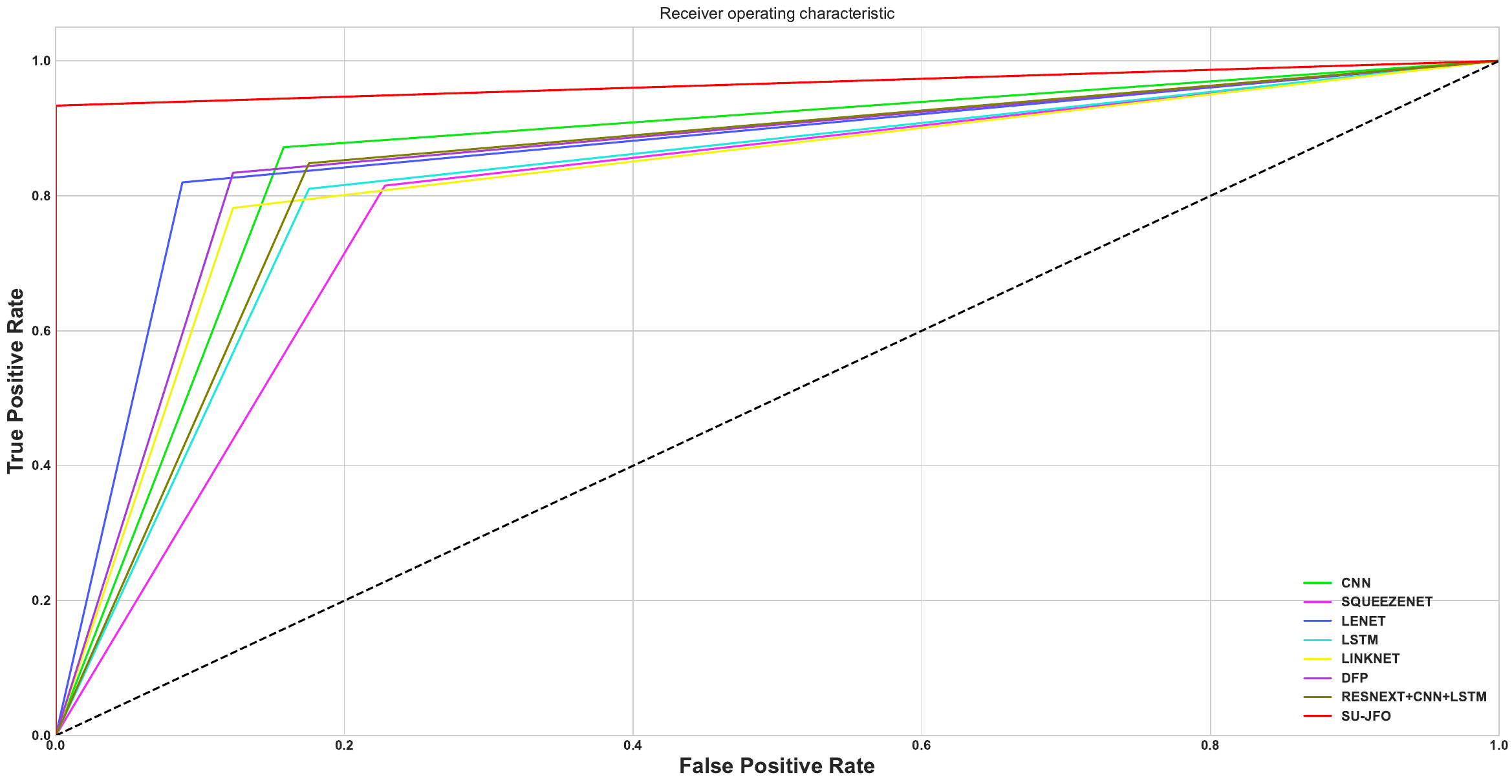}\hfill
	\caption{Evaluation on ROC for Pose Illumination case on Dataset2.}\label{Fig24}
\end{figure*}

\noindent \textbf{ROC Analysis on Pose Illumination Case:}
Figure \ref{Fig24} makes it clear that the SU-JFO model surpasses conventional methods in terms of TPR for detecting deepfakes. The graphical representation clearly illustrates that as the false positive rate increases there is a gradual improvement in the true positive rate across all algorithms. Nevertheless, the SU-JFO model achieves a higher TPR specifically excelling in the precise detection of deepfakes.\\

\noindent \textbf{Statistical Analysis on Accuracy for Pose Illumination Case:}
Table \ref{table8} displays a statistical examination comparing the efficacy of SU-JFO and conventional approaches in the context of deepfake detection. The results reveal that the SU-JFO method consistently outperformed the traditional approaches in terms of accuracy across various statistical metrics. Particularly, for the median statistical metric the SU-JFO approach demonstrated a remarkable accuracy of 0.935 surpassing the lower accuracy values recorded by CNN, SqueezeNet, LeNet, LSTM, LinkNet, DFP \cite{raza2022novel}, and ResNext+CNN+LSTM \cite{vamsi2022deepfake}.
\begin{center}
	\begin{table}[!htbp]
		\resizebox{\textwidth}{!}{%
			\begin{tabular}{|l|l|l|l|l|l|l|l|l|}
				\hline
				\begin{tabular}[c]{@{}c@{}}\textbf{Statistical}\\ \textbf{Metrics}\end{tabular} & \textbf{CNN} & \textbf{SqueezeNet}  & \textbf{LeNet}  & \textbf{LSTM}  & \textbf{LinkNet}  & \textbf{DFP}  & \begin{tabular}[c]{@{}c@{}}\textbf{ResNext+CNN+}\\ \textbf{LSTM}\end{tabular}
   & \begin{tabular}[c]{@{}c@{}}\textbf{SU-JFO}\end{tabular}      \\ 
 \hline  
Mean & 0.839 & 0.794 & 0.828 & 0.812 & 0.798 & 0.815 & 0.835 & 0.935\\ \hline
Maximum & 0.869 & 0.828 & 0.903 & 0.858 & 0.858 & 0.881 & 0.866 & 0.956\\ \hline 
Standard Deviation  & 0.034 & 0.033 & 0.052 & 0.030 & 0.039 & 0.051 & 0.025 & 0.016  \\ \hline
Median & 0.853 & 0.804 & 0.824 & 0.806 & 0.788 & 0.813 & 0.839 & 0.935\\ \hline
Minimum & 0.781 & 0.740 & 0.761 & 0.778 & 0.757 & 0.755 & 0.798 & 0.912\\ \hline
			\end{tabular}%
		}
		\caption{Statistical Assessment on Accuracy For Pose Illumination case on Dataset2.}
		\label{table8}
	\end{table}
\end{center}
\vspace{-5mm}

\subsubsection{Test Case 4: Rotation Scenario}
\begin{figure*}[!htbp]
	\vspace{-10mm}
\includegraphics[height=5cm,width=\textwidth]{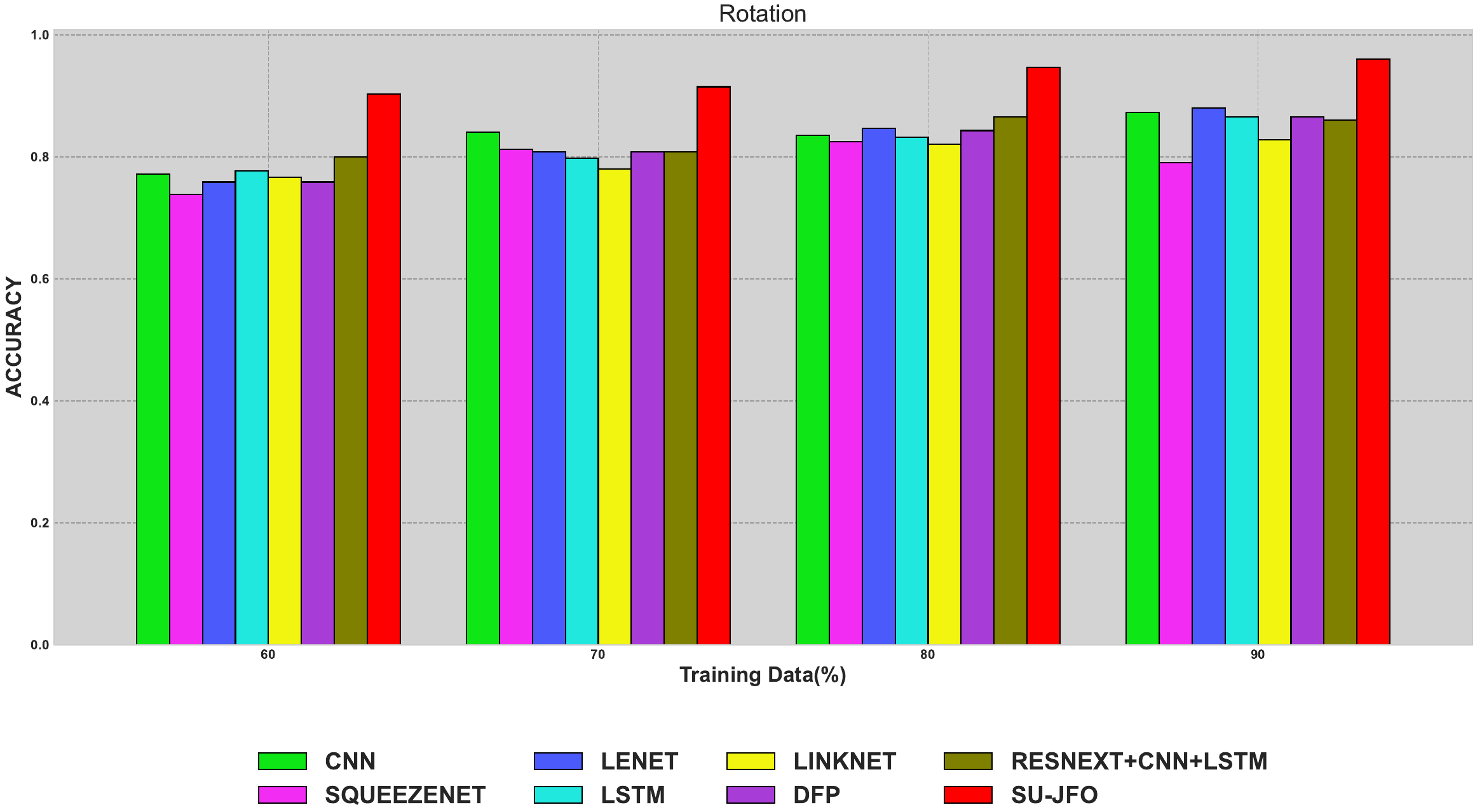}\hfill
\includegraphics[height=5cm,width=\textwidth]{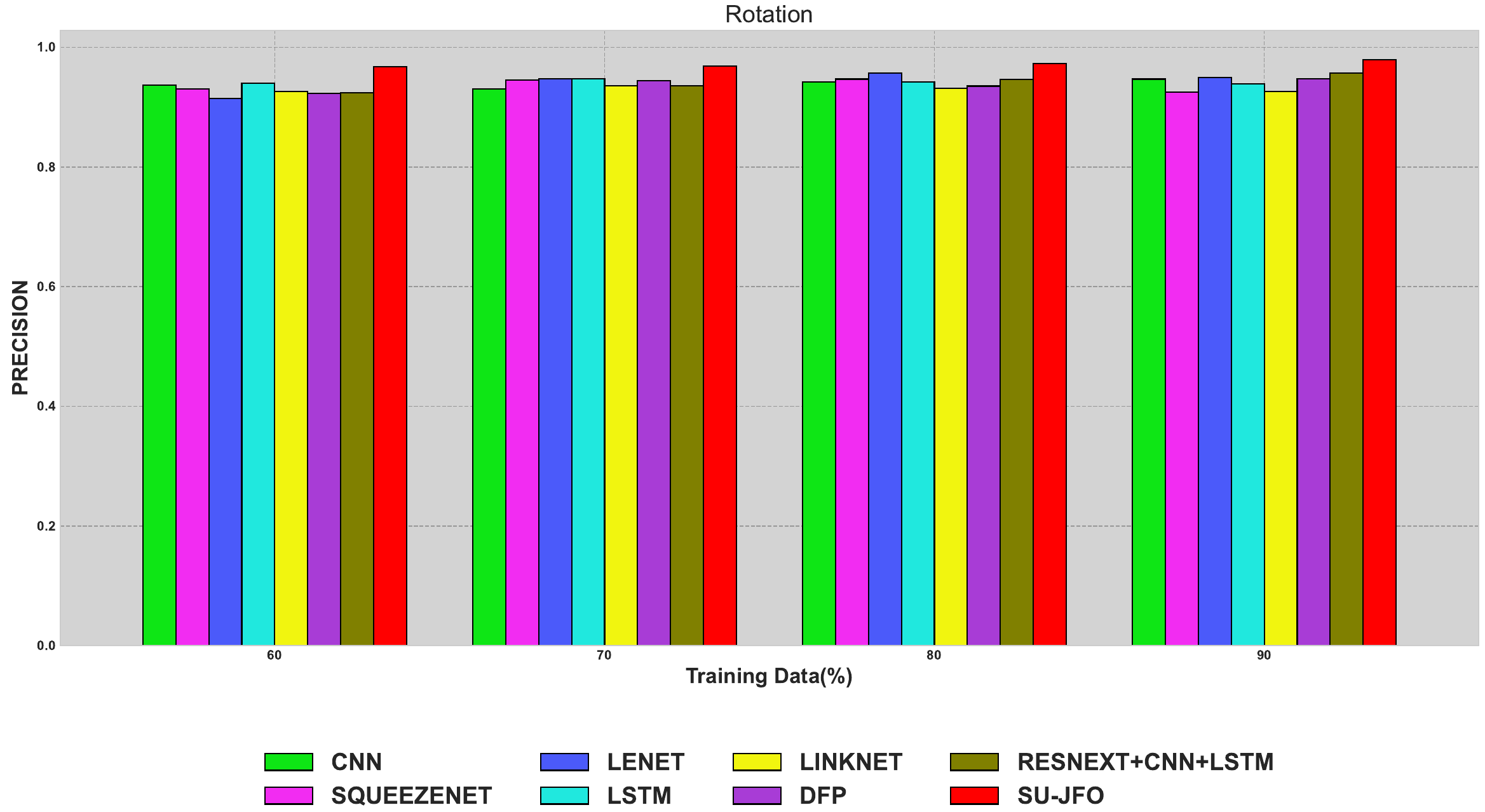}\hfill
\includegraphics[height=5cm,width=\textwidth]{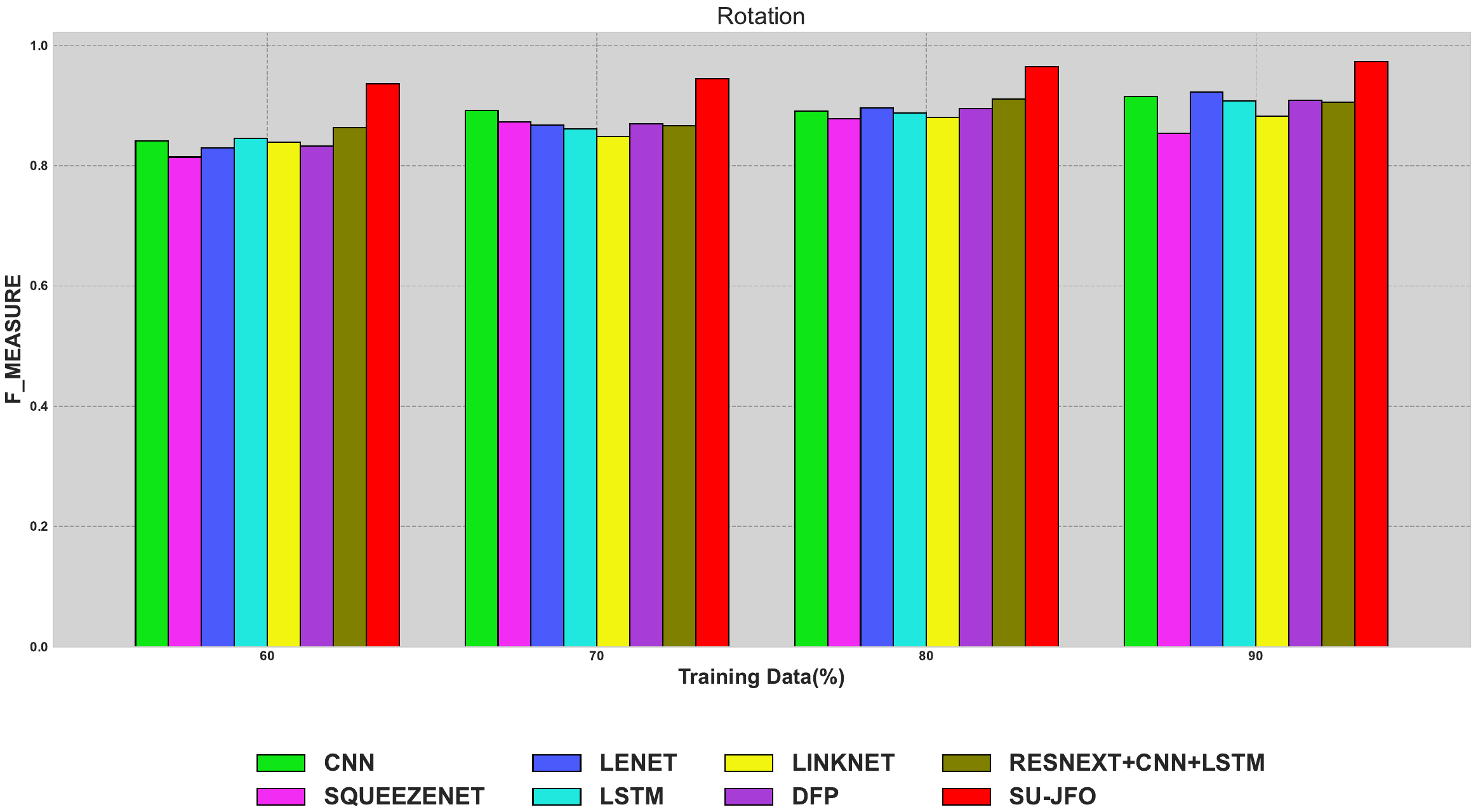}\hfill
\includegraphics[height=5cm,width=\textwidth]{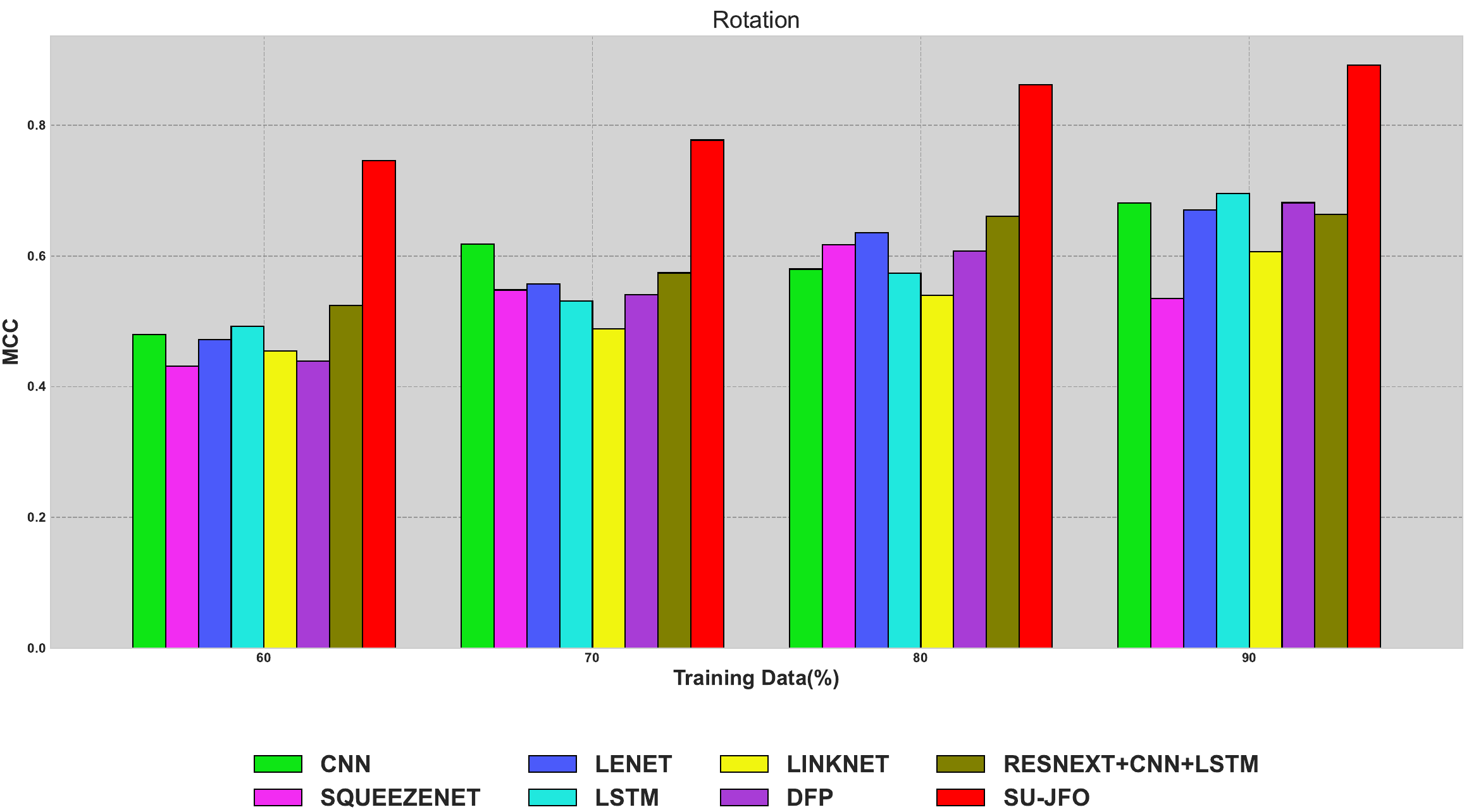}
	\caption{Assessment of SU-JFO and traditional schemes for Rotation case on Dataset2 \\a) Accuracy b) Precision c) F-measure and d) MCC.}\label{Fig25}
\end{figure*}

Figure \ref{Fig25} showcases the performance analysis of the SU-JFO method alongside conventional methodologies for deepfake detection. In the context of the SU-JFO strategy attaining superior values compared to the proceeding methods. This suggests that the SU-JFO approach possesses the capability to more accurately detect deepfakes.\\ 

\begin{figure*}[!t]
	\includegraphics[height=9cm,width=\textwidth]{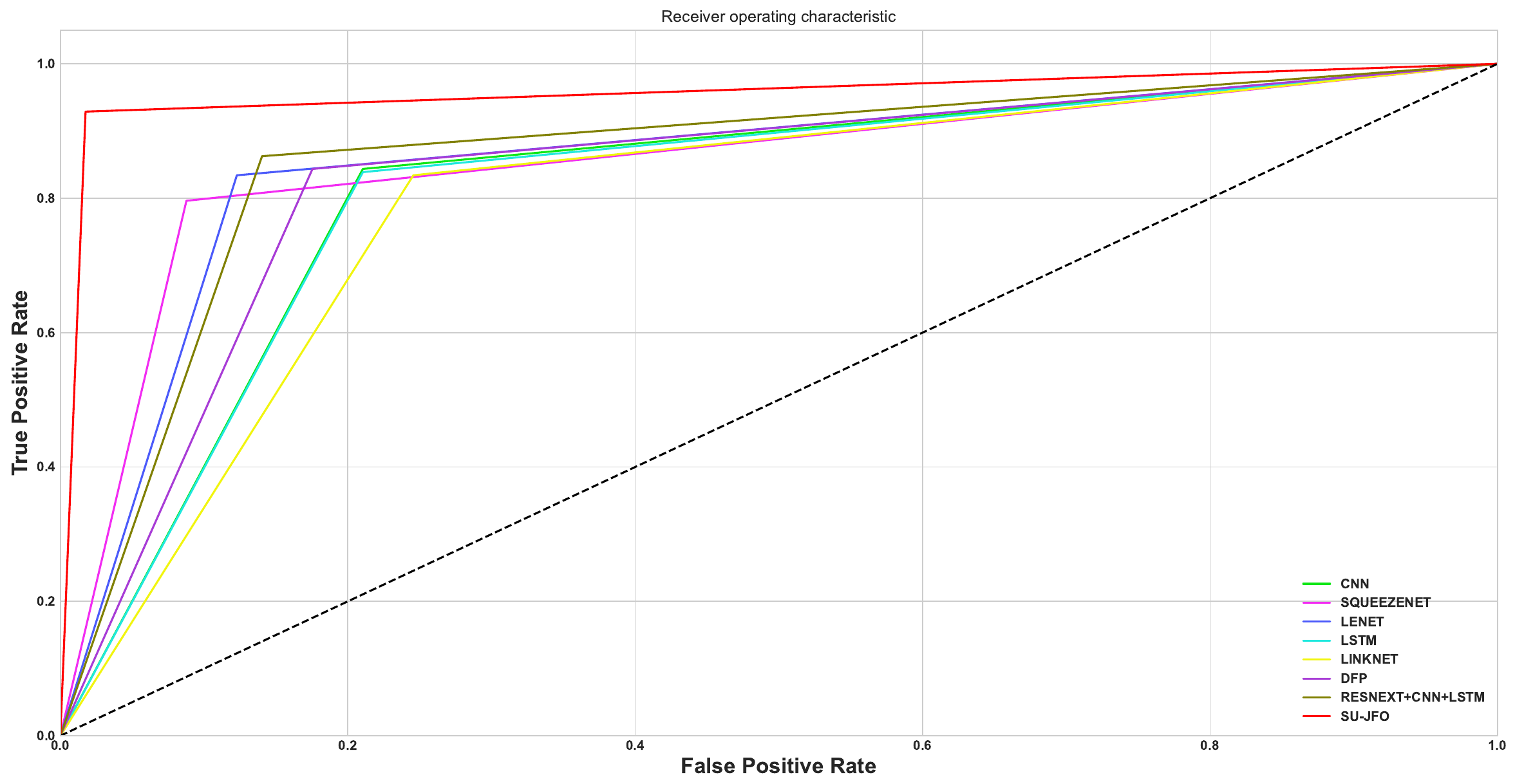}\hfill
	\caption{Evaluation on ROC for Rotation case on Dataset2.}\label{Fig26}
\end{figure*}

\noindent \textbf{ROC Analysis on Rotation Case:}
In Figure \ref{Fig26}, the ROC assessment illustrates the performance of the SU-JFO method compared to conventional approaches in deepfake detection. The SU-JFO approach achieves a TPR exceeding 99\%  whereas CNN, SqueezeNet, LeNet, LSTM, LinkNet, DFP \cite{raza2022novel}, and ResNext+CNN+LSTM \cite{vamsi2022deepfake} exhibit lower TPR.\\

\noindent \textbf{Statistical Analysis on Accuracy for Rotation Case:}
Table \ref{table9} presents a statistical comparison of the SU-JFO method with established models such as CNN, SqueezeNet, LeNet, LSTM, LinkNet, DFP \cite{raza2022novel}, and ResNext+CNN+LSTM \cite{vamsi2022deepfake} for deepfake detection. Specifically when considering mean and median statistical metrics, the SU-JFO approach demonstrates an accuracy of 0.931 which is consistently superior to conventional methods.
\begin{center}
	\begin{table}[!htbp]
		\resizebox{\textwidth}{!}{%
			\begin{tabular}{|l|l|l|l|l|l|l|l|l|}
				\hline
				\begin{tabular}[c]{@{}c@{}}\textbf{Statistical}\\ \textbf{Metrics}\end{tabular} & \textbf{CNN} & \textbf{SqueezeNet}  & \textbf{LeNet}  & \textbf{LSTM}  & \textbf{LinkNet}  & \textbf{DFP}  & \begin{tabular}[c]{@{}c@{}}\textbf{ResNext+CNN+}\\ \textbf{LSTM}\end{tabular}
   & \begin{tabular}[c]{@{}c@{}}\textbf{SU-JFO}\end{tabular}      \\ 
 \hline  
Mean & 0.830 & 0.792 & 0.824 & 0.818 & 0.799 & 0.819 & 0.833 & 0.931\\ \hline
Maximum & 0.873 & 0.825 & 0.881 & 0.866 & 0.828 & 0.866 & 0.866 & 0.960\\ \hline 
Standard Deviation  & 0.037 & 0.033 & 0.045 & 0.034 & 0.026 & 0.040 & 0.030 & 0.023  \\ \hline
Median & 0.838 & 0.802 & 0.827 & 0.815 & 0.801 & 0.826 & 0.834 & 0.931\\ \hline
Minimum & 0.772 & 0.738 & 0.759 & 0.778 & 0.766 & 0.759 & 0.800 & 0.903\\ \hline
			\end{tabular}%
		}
		\caption{Statistical Assessment on Accuracy For Rotation case on Dataset2.}
		\label{table9}
	\end{table}
\end{center}
\vspace{-6mm}

\subsection{Comparative Analysis on Dataset3}\label{section4.8}
\subsubsection{Testcase 1: Compression Scenario}
\begin{figure*}[!htbp]
	\vspace{-10mm}
\includegraphics[height=5cm,width=\textwidth]{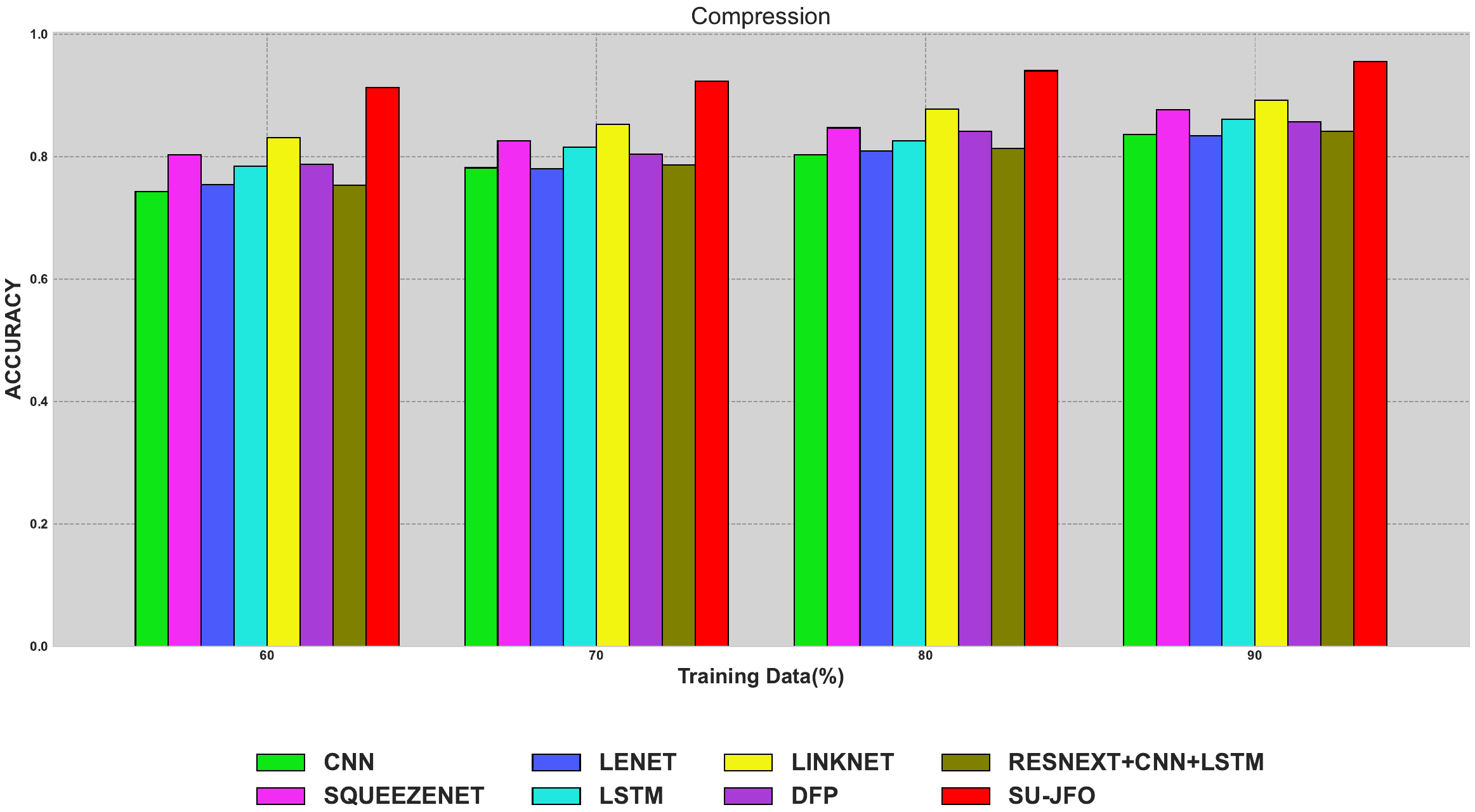}\hfill
\includegraphics[height=5cm,width=\textwidth]{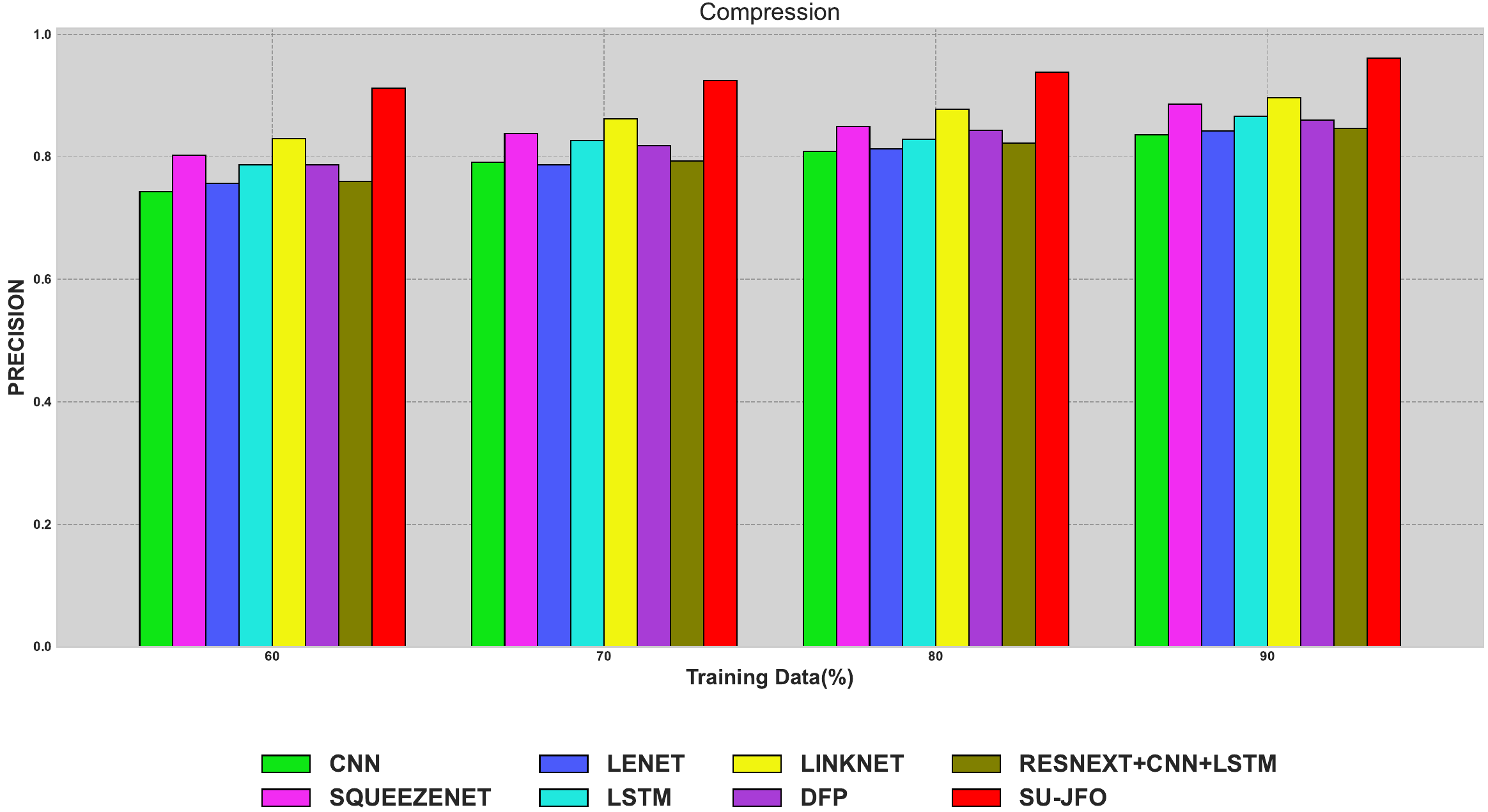}\hfill
\includegraphics[height=5cm,width=\textwidth]{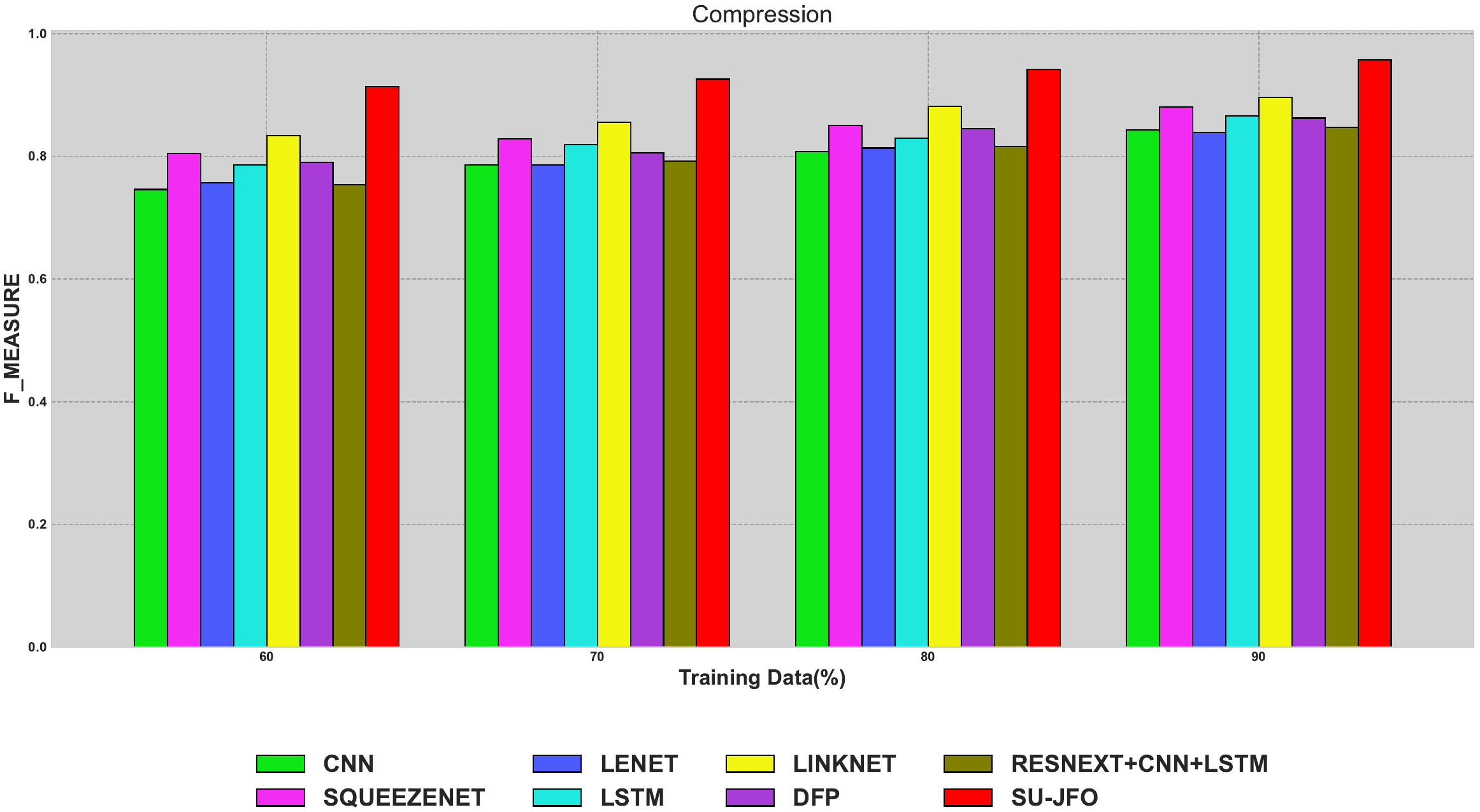}\hfill
\includegraphics[height=5cm,width=\textwidth]{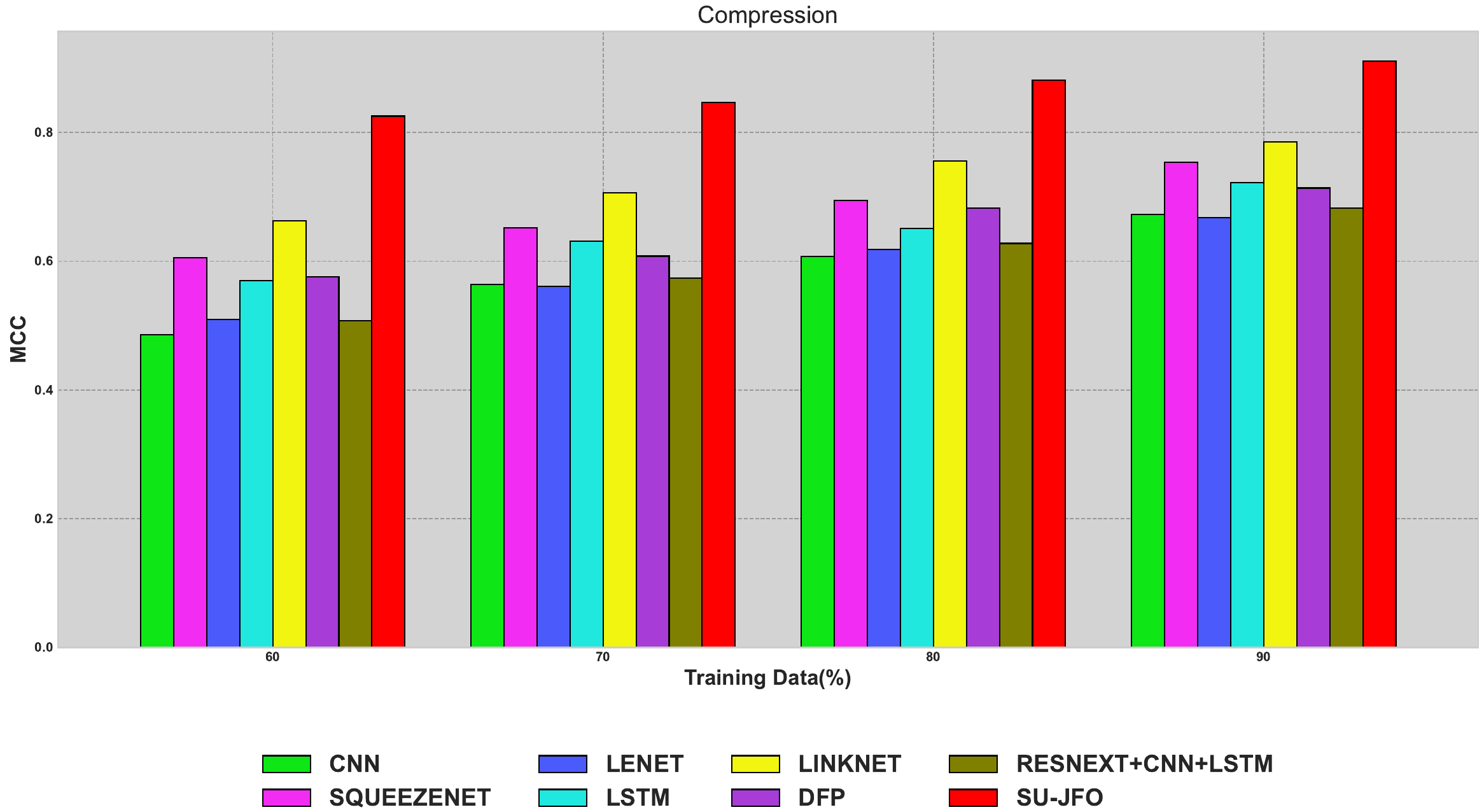}
	\caption{Assessment of SU-JFO and traditional schemes for Compression case on Dataset3 \\ a) Accuracy b) Precision c) F-measure, and d) MCC.}\label{Fig27}
\end{figure*}

The performance of the proposed SU-JFO method is compared against several deepfake detection approaches including CNN, SqueezeNet, LeNet, LSTM, LinkNet, DFP \cite{raza2022novel}, and ResNext+CNN+LSTM \cite{vamsi2022deepfake}  as illustrated in Figure \ref{Fig27}. The evaluation focuses on key metrics such as Precision, MCC, Accuracy, and F-measure. Consistent with previous findings on Dataset 3, the SU-JFO method demonstrates superior performance. For instance, at a 90\% training data percentage the SU-JFO achieves an accuracy of 0.955629 surpassing traditional methods: CNN (0.836), SqueezeNet (0.876), LeNet (0.834), LSTM (0.861), LinkNet (0.892), DFP \cite{raza2022novel} (0.857), and ResNext+CNN+LSTM \cite{vamsi2022deepfake} (0.841).\\

\begin{figure*}[!t]
	\includegraphics[height=9cm,width=\textwidth]{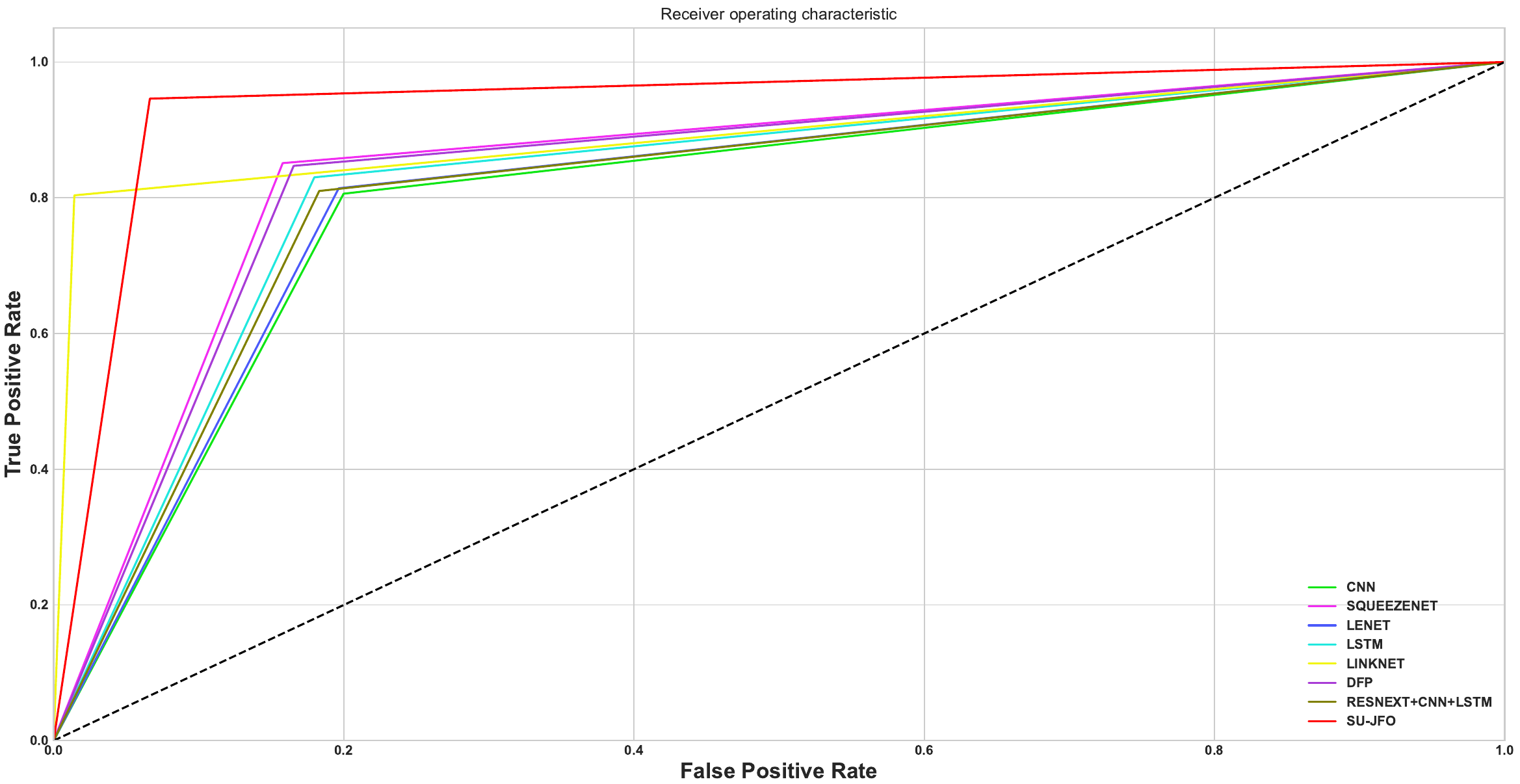}\hfill
	\caption{Evaluation on ROC for Compression Case  on Dataset3.}\label{Fig28}
\end{figure*}

\noindent \textbf{ROC Analysis on Compression Case:}
In Figure \ref{Fig28}, the ROC analysis compares the performance of the SU-JFO method against CNN, SqueezeNet, LeNet, LSTM, LinkNet, DFP \cite{raza2022novel}, and ResNext+CNN+LSTM \cite{vamsi2022deepfake}  for deepfake detection. Achieving ROC values above 97.6\% is essential for robust detection of deepfakes. In this evaluation, all the algorithms exceeded a 94\% threshold at a false positive rate of 0.8. However, the SU-JFO method stands out delivering the highest ROC values surpassing the performance of the other approaches in this comparison. \\

\noindent \textbf{Statistical Analysis on Accuracy for Compression Case:}\\
Table \ref{table10} presents a statistical comparison of the effectiveness of the SU-JFO method versus conventional approaches for deepfake detection. The findings highlight that SU-JFO consistently outperforms traditional methods in terms of accuracy across various statistical measures. Notably, for the highest statistical metric the SU-JFO achieved an impressive accuracy of 0.956 significantly higher than the accuracy rates of other approaches such as CNN (0.836), SqueezeNet (0.876), LeNet (0.834), LSTM (0.861), LinkNet (0.892), DFP (0.857) \cite{raza2022novel} , and ResNext+CNN+LSTM  (0.841) \cite{vamsi2022deepfake}. \\

\begin{center}
	\begin{table}[!htbp]
		\resizebox{\textwidth}{!}{%
			\begin{tabular}{|l|l|l|l|l|l|l|l|l|}
				\hline
				\begin{tabular}[c]{@{}c@{}}\textbf{Statistical}\\ \textbf{Metrics}\end{tabular} & \textbf{CNN} & \textbf{SqueezeNet}  & \textbf{LeNet}  & \textbf{LSTM}  & \textbf{LinkNet}  & \textbf{DFP}  & \begin{tabular}[c]{@{}c@{}}\textbf{ResNext+CNN+}\\ \textbf{LSTM}\end{tabular}
   & \begin{tabular}[c]{@{}c@{}}\textbf{SU-JFO}\end{tabular}      \\ 
 \hline  
Mean & 0.791 & 0.838 & 0.795 & 0.822 & 0.864 & 0.823 & 0.799 & 0.933\\ \hline
Maximum & 0.836 & 0.877 & 0.834 & 0.861 & 0.893 & 0.857 & 0.841 & 0.956\\ \hline 
Standard Deviation  & 0.034 & 0.027 & 0.030 & 0.027 & 0.023 & 0.028 & 0.032 & 0.016  \\ \hline
Median & 0.793 & 0.837 & 0.795 & 0.821 & 0.866 & 0.823 & 0.800 & 0.932\\ \hline
Minimum & 0.743 & 0.803 & 0.755 & 0.785 & 0.831 & 0.788 & 0.754 & 0.913\\ \hline
			\end{tabular}%
		}
		\caption{Statistical Assessment on Accuracy For Compression Case on Dataset3.}
		\label{table10}
	\end{table}
\end{center}
\vspace{-14mm}

\subsubsection{Test Case 2: Noise Scenario}
\begin{figure*}[!htbp]
	\vspace{-10mm}
\includegraphics[height=5cm,width=\textwidth]{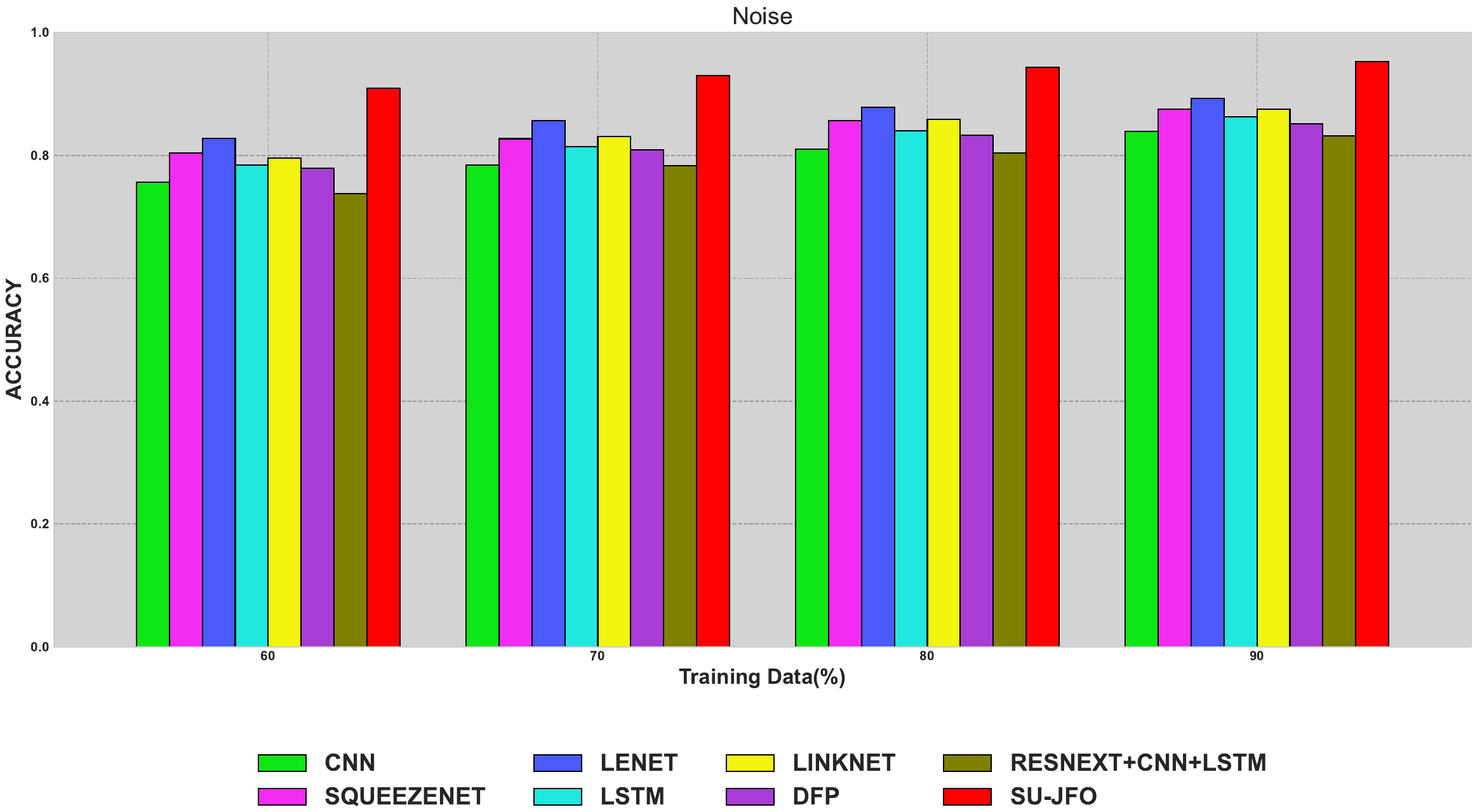}\hfill
\includegraphics[height=5cm,width=\textwidth]{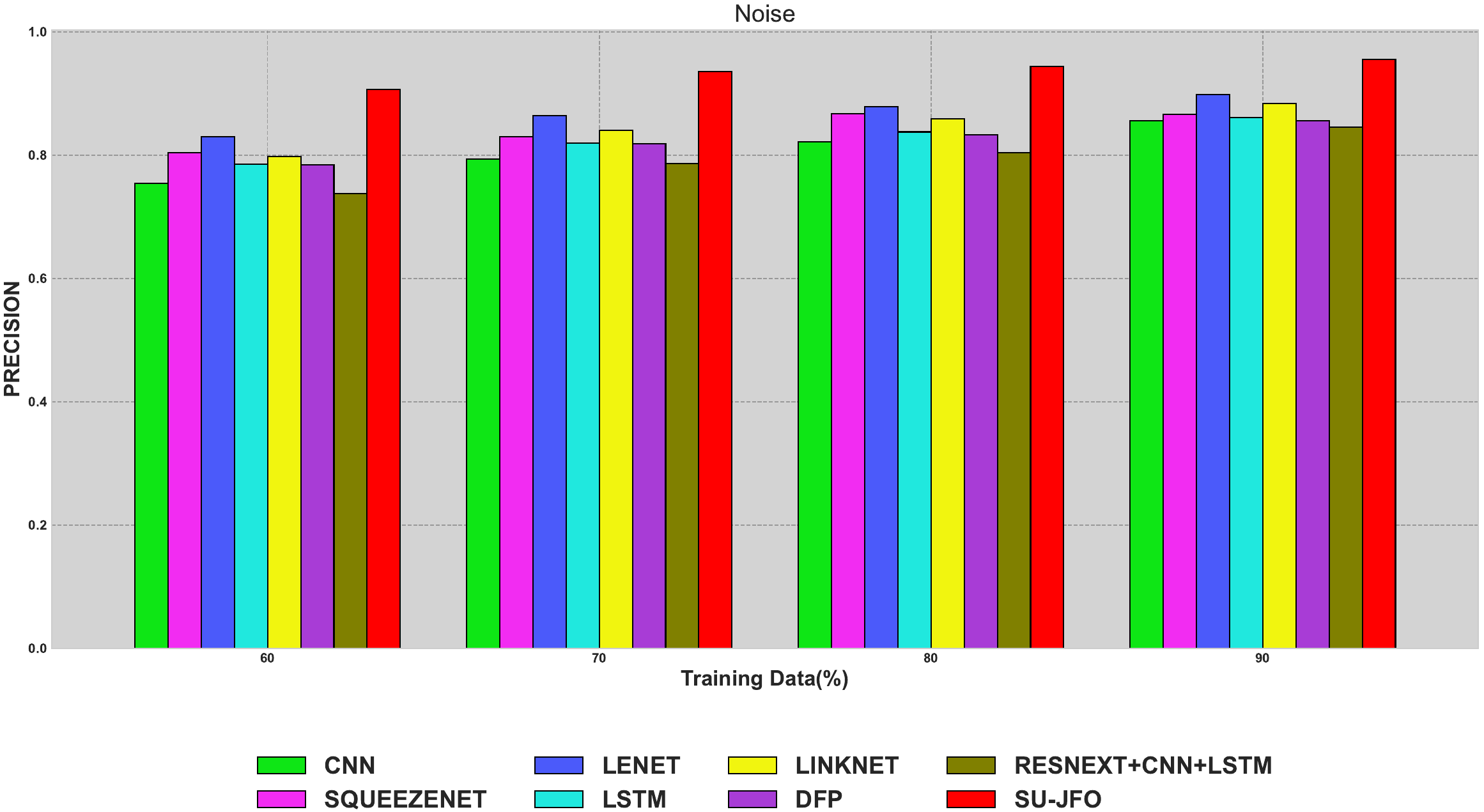}\hfill
\includegraphics[height=5cm,width=\textwidth]{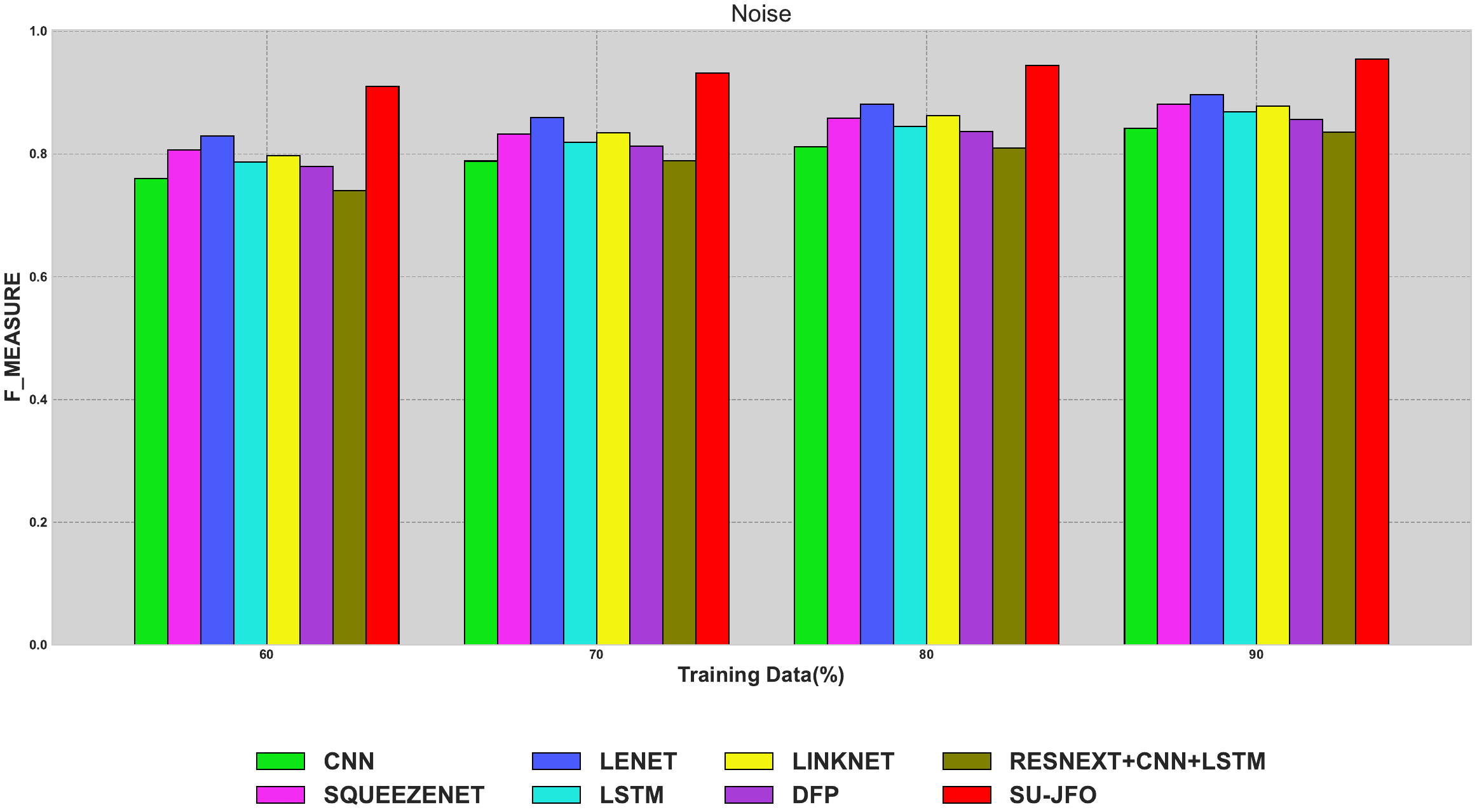}\hfill
\includegraphics[height=5cm,width=\textwidth]{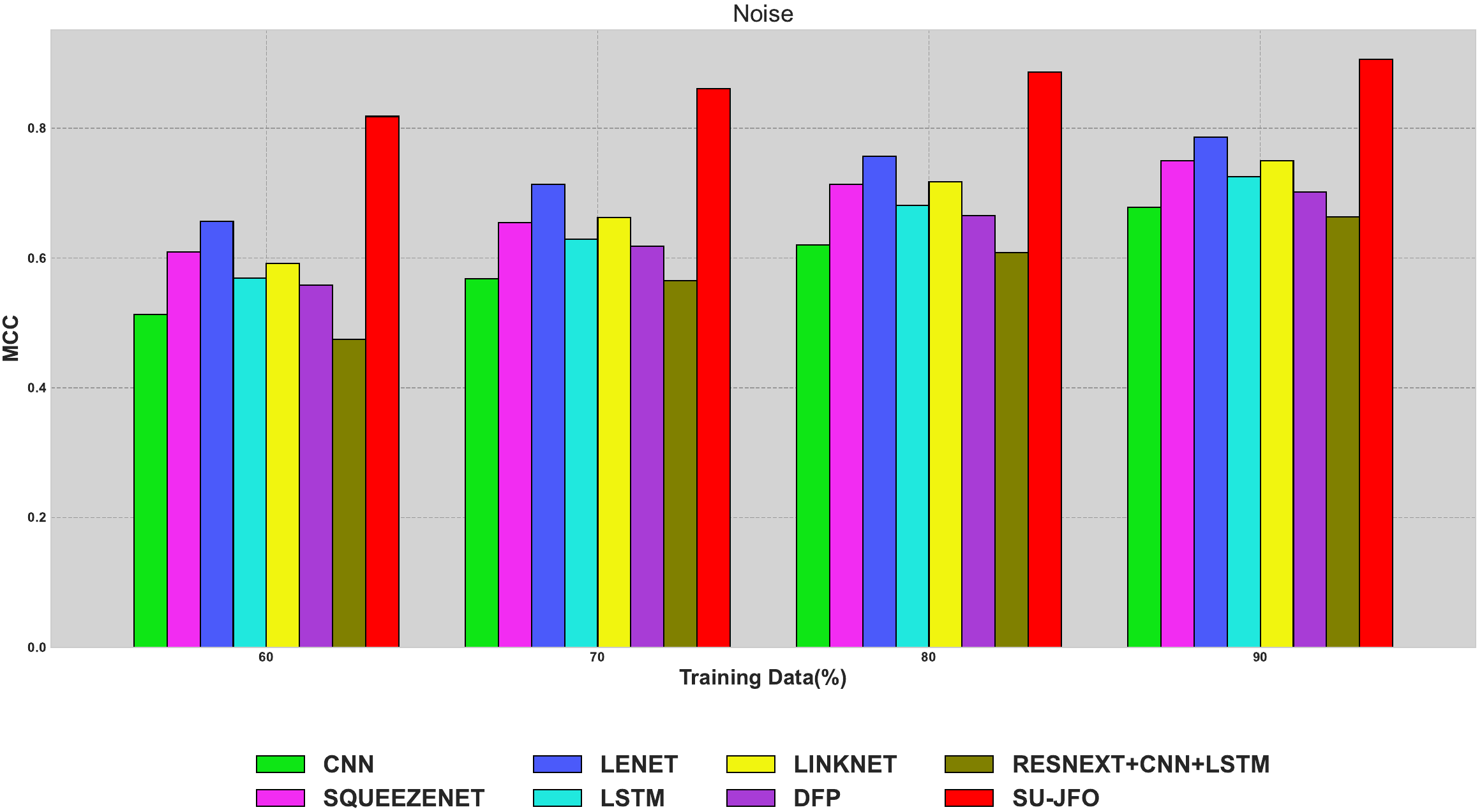}
	\caption{Assessment of SU-JFO and traditional schemes for Noise case  on Dataset3 \\ a) Accuracy b) Precision c) F-measure and d) MCC.}\label{Fig29}
\end{figure*}

To evaluate the effectiveness of the SU-JFO-based deepfake detection approach an extensive analysis was performed using multiple metrics including Accuracy, F-measure, MCC, and Precision across different sets of training data. The results of this evaluation are depicted in Figure \ref{Fig29}. Additionally, the proposed model was compared with several other models including CNN, SqueezeNet, LeNet, LSTM, LinkNet, DFP \cite{raza2022novel}, and ResNext+CNN+LSTM \cite{vamsi2022deepfake}. In this comparison, the SU-JFO method consistently outperformed the traditional approaches. As depicted in Figure \ref{Fig29} , the SU-JFO achieved highest accuracy across various training percentages compared to conventional methods. 
\vspace{2mm}

\begin{figure*}[!t]
	\includegraphics[height=9cm,width=\textwidth]{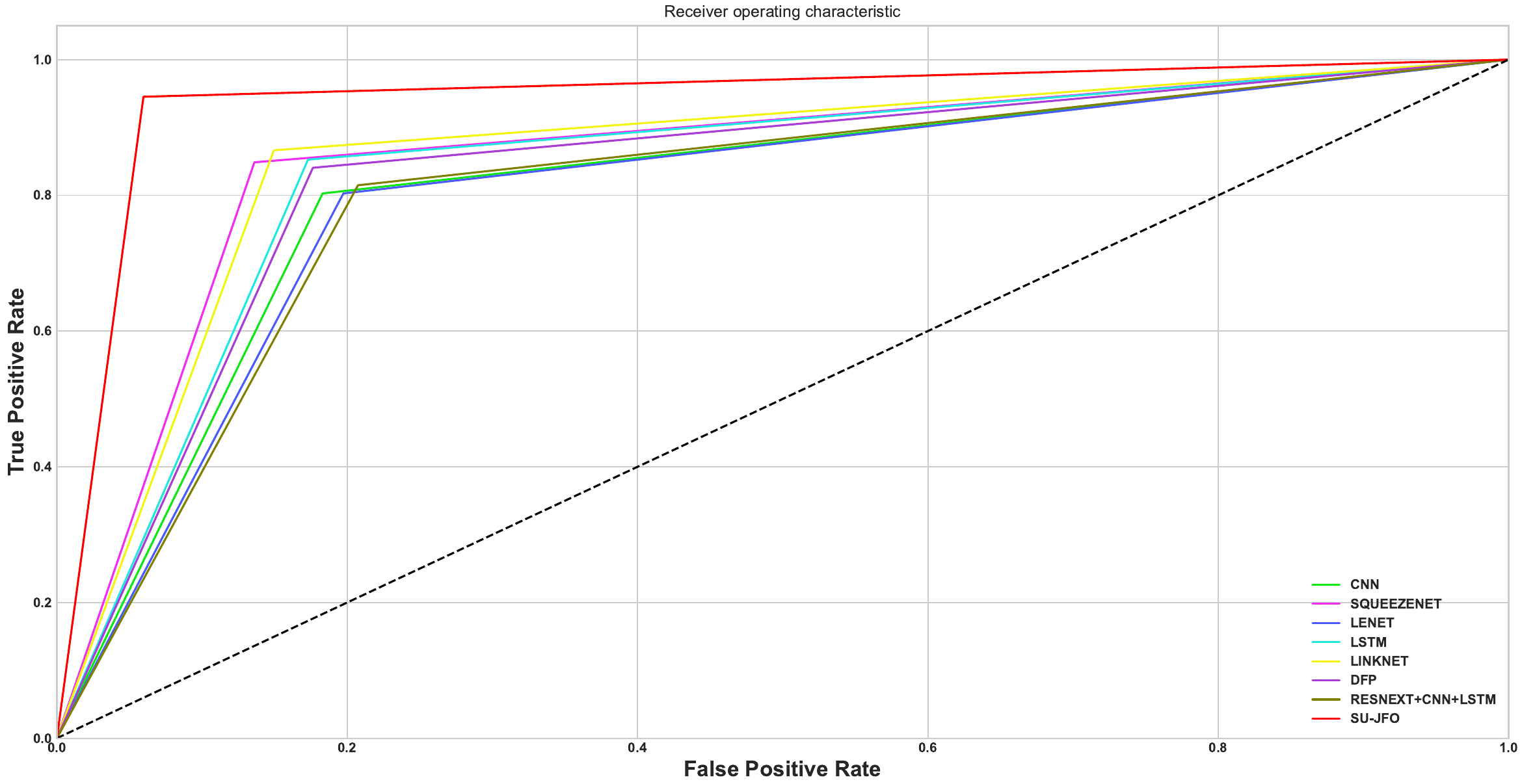}\hfill
	\caption{Evaluation on ROC for Noise Case  on Dataset3.}\label{Fig30}
\end{figure*}

\noindent \textbf{ROC Analysis on Noise Case:}
Figure \ref{Fig30} presents the ROC assessment showcasing the performance of the SU-JFO method in comparison to conventional approaches for deepfake detection. The SU-JFO method achieves a TPR exceeding 99\% while traditional methods such as CNN, SqueezeNet, LeNet, LSTM, LinkNet, DFP \cite{raza2022novel}, and ResNext+CNN+LSTM \cite{vamsi2022deepfake} display lower TPR.\\

\noindent \textbf{Statistical Analysis on Accuracy on Noise Case:}
Table \ref{table11} provides a statistical comparison of the effectiveness of the SU-JFO method against traditional approaches in deepfake detection. The analysis reveals that the SU-JFO method consistently achieved higher accuracy scores across the majority of statistical metrics while the conventional methods tended to exhibit lower accuracy values.

\begin{center}
	\begin{table}[!htbp]
		\resizebox{\textwidth}{!}{%
			\begin{tabular}{|l|l|l|l|l|l|l|l|l|}
				\hline
				\begin{tabular}[c]{@{}c@{}}\textbf{Statistical}\\ \textbf{Metrics}\end{tabular} & \textbf{CNN} & \textbf{SqueezeNet}  & \textbf{LeNet}  & \textbf{LSTM}  & \textbf{LinkNet}  & \textbf{DFP}  & \begin{tabular}[c]{@{}c@{}}\textbf{ResNext+CNN+}\\ \textbf{LSTM}\end{tabular}
   & \begin{tabular}[c]{@{}c@{}}\textbf{SU-JFO}\end{tabular}      \\ 
 \hline  
Mean & 0.798 & 0.841 & 0.864 & 0.826 & 0.840 & 0.818 & 0.789 & 0.934\\ \hline
Maximum & 0.839 & 0.875 & 0.893 & 0.863 & 0.875 & 0.851 & 0.832 & 0.953\\ \hline 
Standard Deviation  & 0.031 & 0.027 & 0.024 & 0.029 & 0.030 & 0.027 & 0.034 & 0.016  \\ \hline
Median & 0.797 & 0.842 & 0.868 & 0.828 & 0.845 & 0.821 & 0.794 & 0.937\\ \hline
Minimum & 0.757 & 0.805 & 0.828 & 0.785 & 0.796 & 0.779 & 0.738 & 0.909\\ \hline
			\end{tabular}%
		}
		\caption{Statistical Assessment on Accuracy For Noise Case  on Dataset3.}
		\label{table11}
	\end{table}
\end{center}

\subsubsection{Test Case 3: Pose Illumination Scenario}
\begin{figure*}[!htbp]
\includegraphics[height=5cm,width=\textwidth]{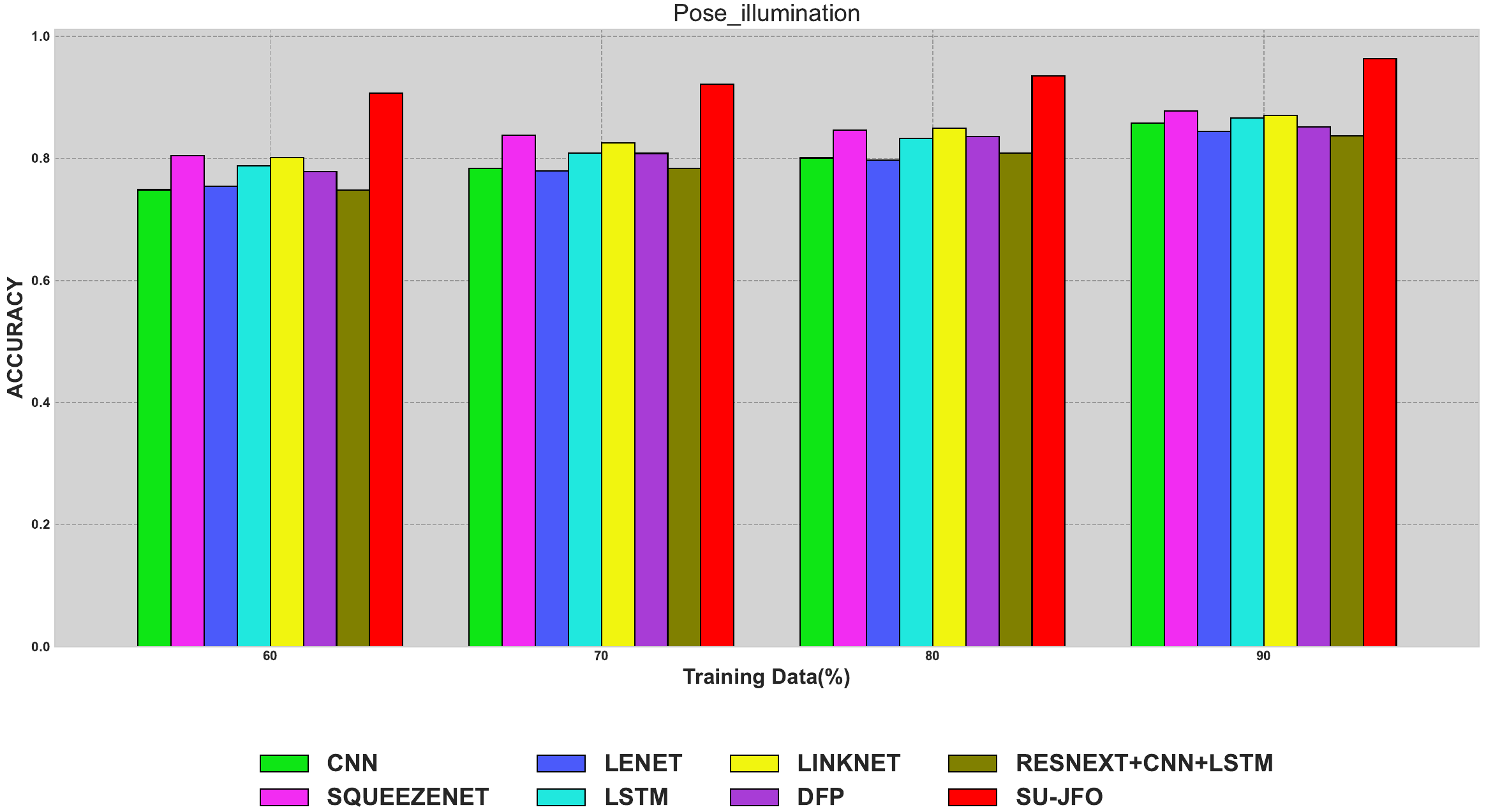}\hfill
\includegraphics[height=5cm,width=\textwidth]{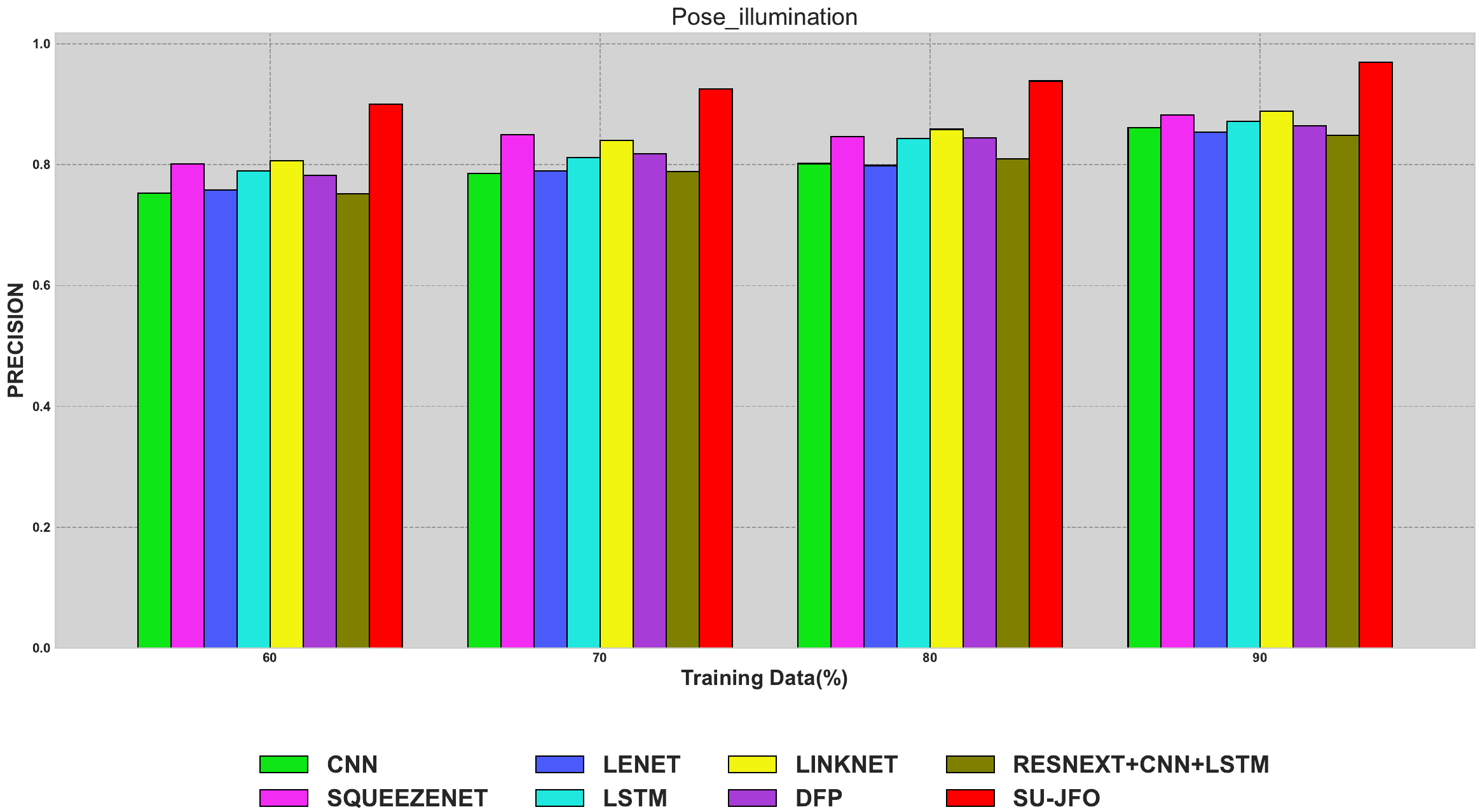}\hfill
\includegraphics[height=5cm,width=\textwidth]{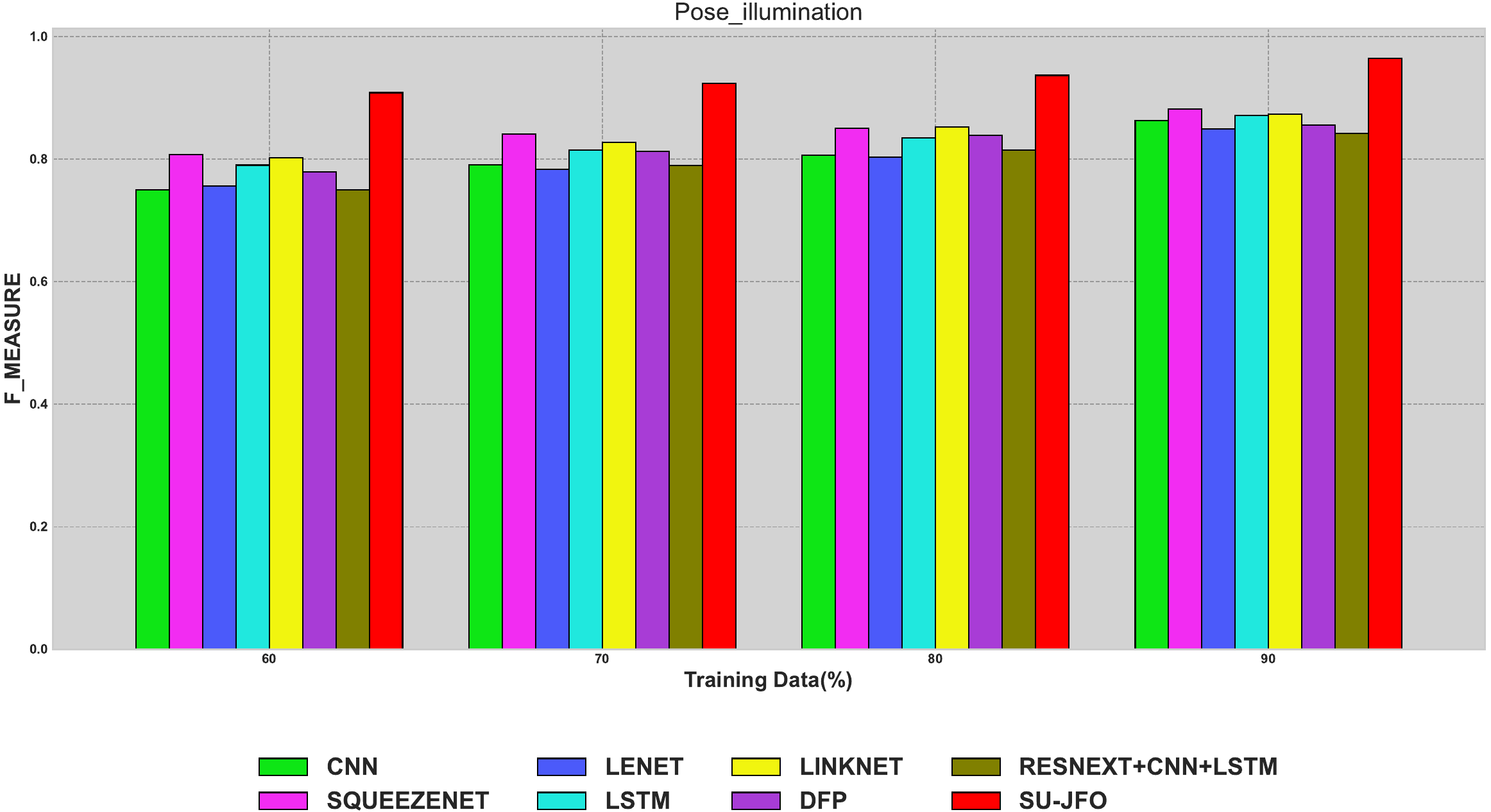}\hfill
\includegraphics[height=5cm,width=\textwidth]{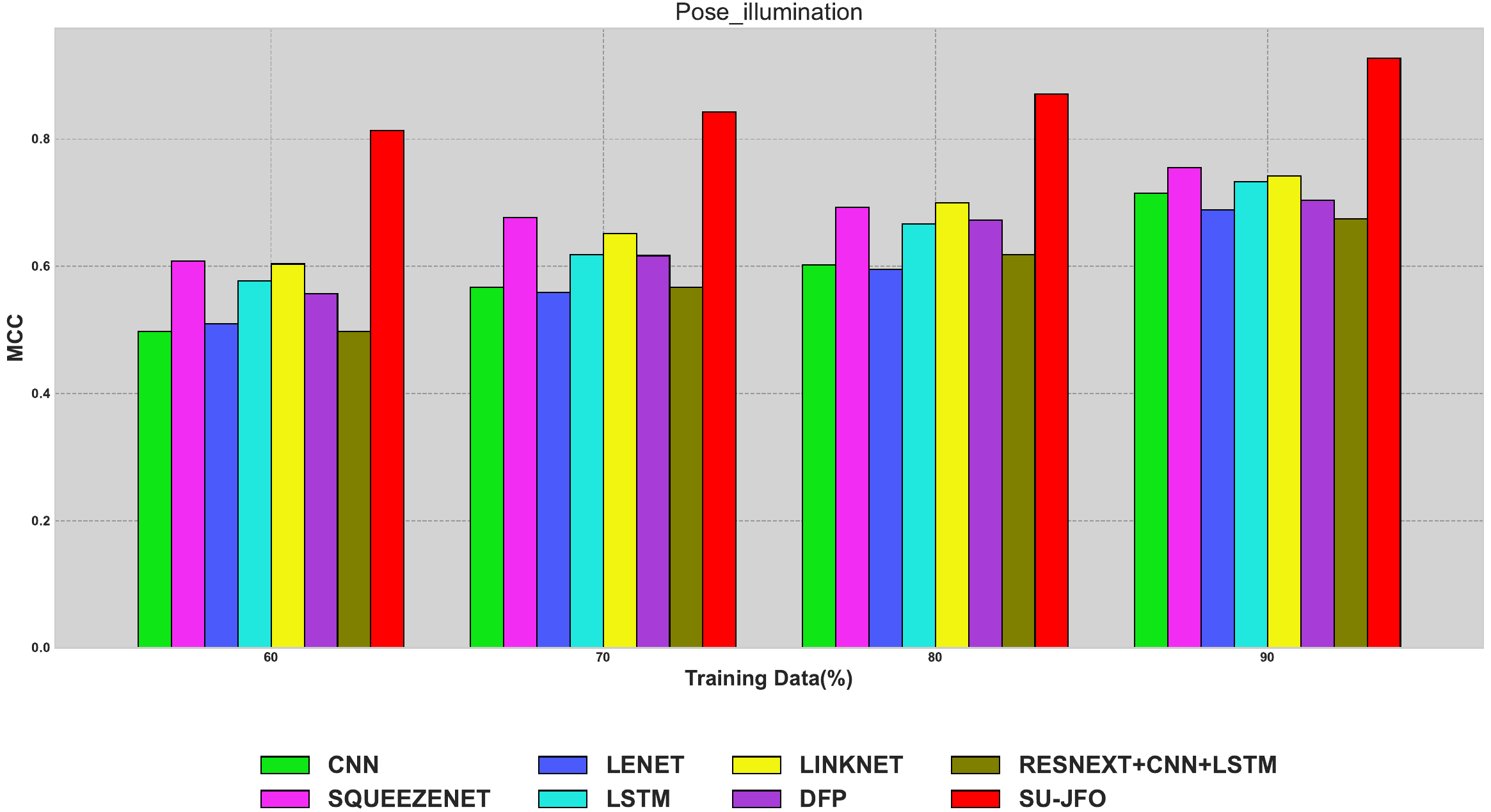}
	\caption{Assessment of SU-JFO and traditional schemes for Pose illumination case  on Dataset3 a) Accuracy b) Precision c) F-measure and d) MCC.}\label{Fig31}
\end{figure*}

Figure \ref{Fig31} presents a graphical representation of the SU-JFO methodology's effectiveness in detecting deepfakes using various training datasets. The results show that the model achieves an impressive precision rate of 0.970 when utilizing 90\% of the training data while traditional methods report lower precision rates: CNN at 0.861, SqueezeNet at 0.882, LeNet at 0.854, LSTM at 0.871, LinkNet at 0.888, DFP \cite{raza2022novel} at 0.864, and ResNext+CNN+LSTM \cite{vamsi2022deepfake} at 0.848. Furthermore, in all training data scenarios the SU-JFO approach consistently outperforms conventional methods as evidenced by its superior metric values. \\

\begin{figure*}[!t]
	\includegraphics[height=9cm,width=\textwidth]{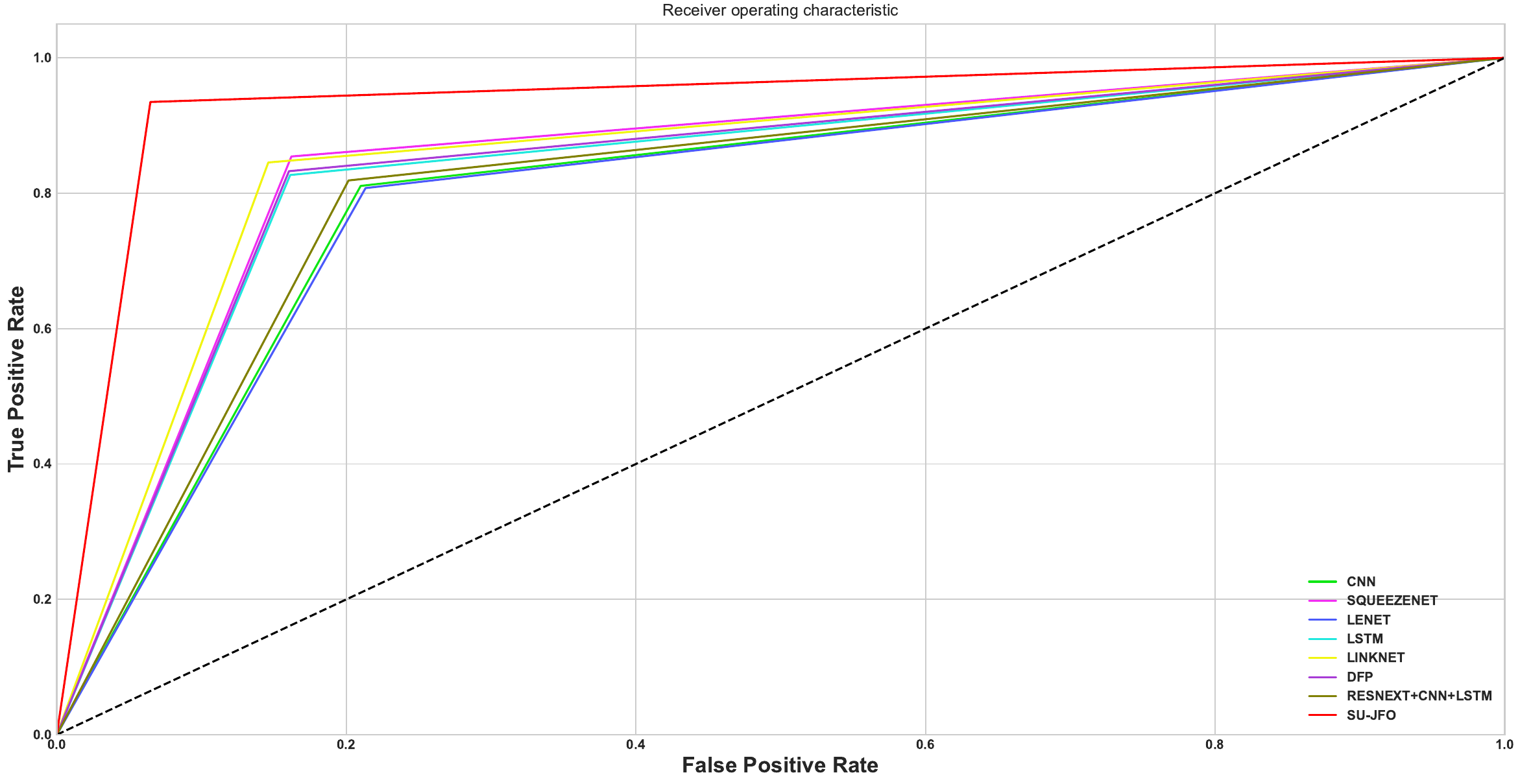}\hfill
	\caption{Evaluation on ROC for Pose illumination Case  on Dataset3.}\label{Fig32}
\end{figure*}

 \noindent \textbf{ROC Analysis on Pose Illumination Case:}
Figure \ref{Fig32} highlights the superiority of the SU-JFO model over traditional methods regarding the TPR in deepfake detection. The graph illustrates that as the false positive rate rises, the TPR improves gradually for all algorithms. However, the SU-JFO model distinctly achieves a higher TPR demonstrating its effectiveness in accurately identifying deepfakes. \\

\noindent \textbf{Statistical Analysis on Accuracy for Pose Illumination Case:}
Table \ref{table12} presents a statistical comparison of the effectiveness of the SU-JFO method against traditional approaches in deepfake detection. The findings indicate that the SU-JFO consistently achieved higher accuracy across various statistical metrics. Notably, for the mean accuracy metric the SU-JFO method reached an impressive accuracy of 0.932 significantly exceeding the lower accuracy values of CNN, SqueezeNet, LeNet, LSTM, LinkNet, DFP \cite{raza2022novel}, and ResNext+CNN+LSTM \cite{vamsi2022deepfake}. 

\begin{center}
	\begin{table}[!htbp]
		\resizebox{\textwidth}{!}{%
			\begin{tabular}{|l|l|l|l|l|l|l|l|l|}
				\hline
				\begin{tabular}[c]{@{}c@{}}\textbf{Statistical}\\ \textbf{Metrics}\end{tabular} & \textbf{CNN} & \textbf{SqueezeNet}  & \textbf{LeNet}  & \textbf{LSTM}  & \textbf{LinkNet}  & \textbf{DFP}  & \begin{tabular}[c]{@{}c@{}}\textbf{ResNext+CNN+}\\ \textbf{LSTM}\end{tabular}
   & \begin{tabular}[c]{@{}c@{}}\textbf{SU-JFO}\end{tabular}      \\ 
 \hline  
Mean & 0.798 & 0.842 & 0.794 & 0.824 & 0.837 & 0.819 & 0.795 & 0.932\\ \hline
Maximum & 0.858 & 0.878 & 0.845 & 0.867 & 0.871 & 0.852 & 0.837 & 0.964\\ \hline 
Standard Deviation  & 0.039 & 0.026 & 0.033 & 0.029 & 0.026 & 0.028 & 0.033 & 0.021  \\ \hline
Median & 0.792 & 0.843 & 0.789 & 0.821 & 0.838 & 0.822 & 0.796 & 0.929\\ \hline
Minimum & 0.749 & 0.804 & 0.755 & 0.788 & 0.802 & 0.779 & 0.749 & 0.907\\ \hline
			\end{tabular}%
		}
		\caption{Statistical Assessment on Accuracy For Pose Illumination case  on Dataset3.}
		\label{table12}
	\end{table}
\end{center}

\subsubsection{Test Case 4: Rotation Scenario}
\begin{figure*}[!htbp]
	\vspace{-10mm}
\includegraphics[height=5cm,width=\textwidth]{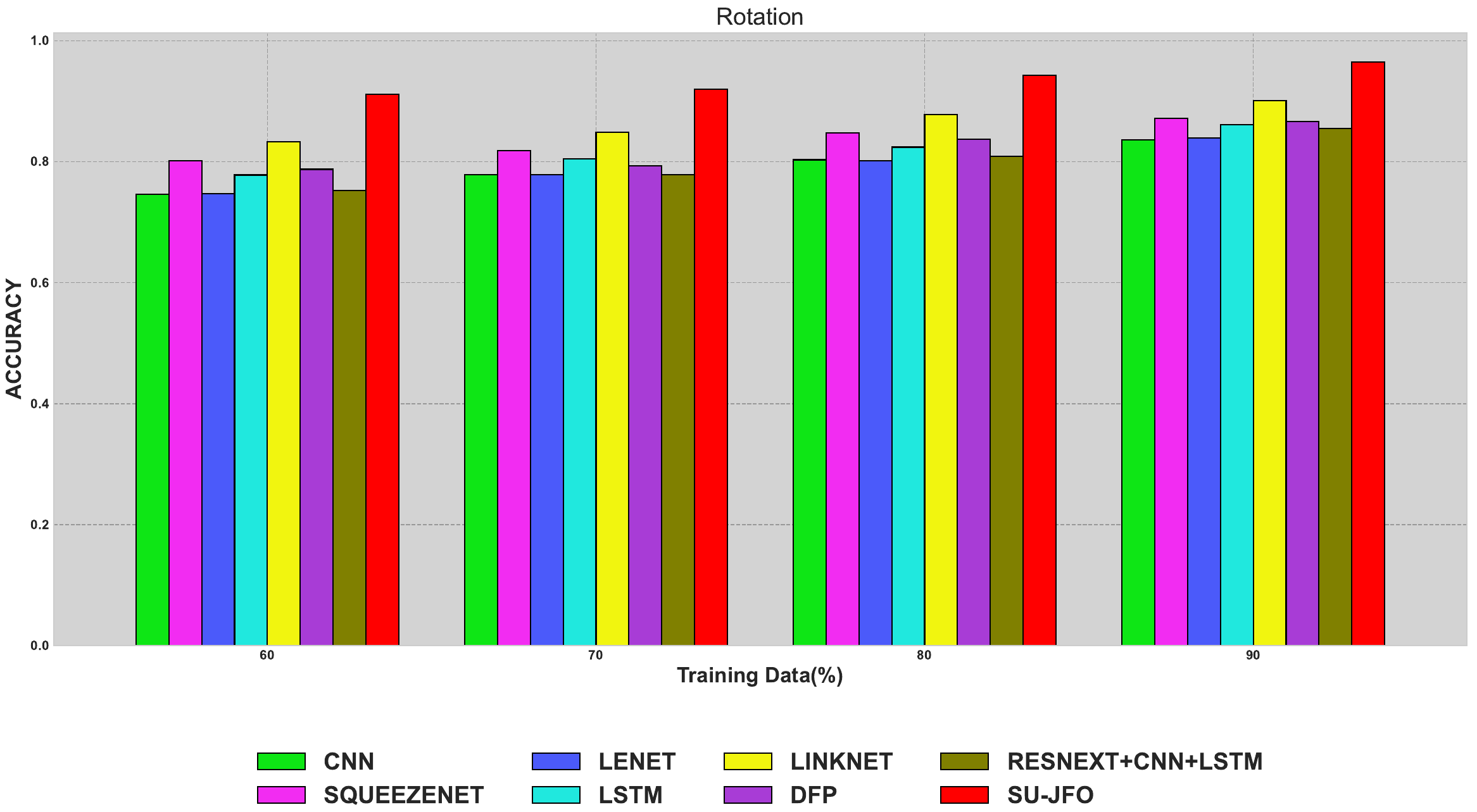}\hfill
\includegraphics[height=5cm,width=\textwidth]{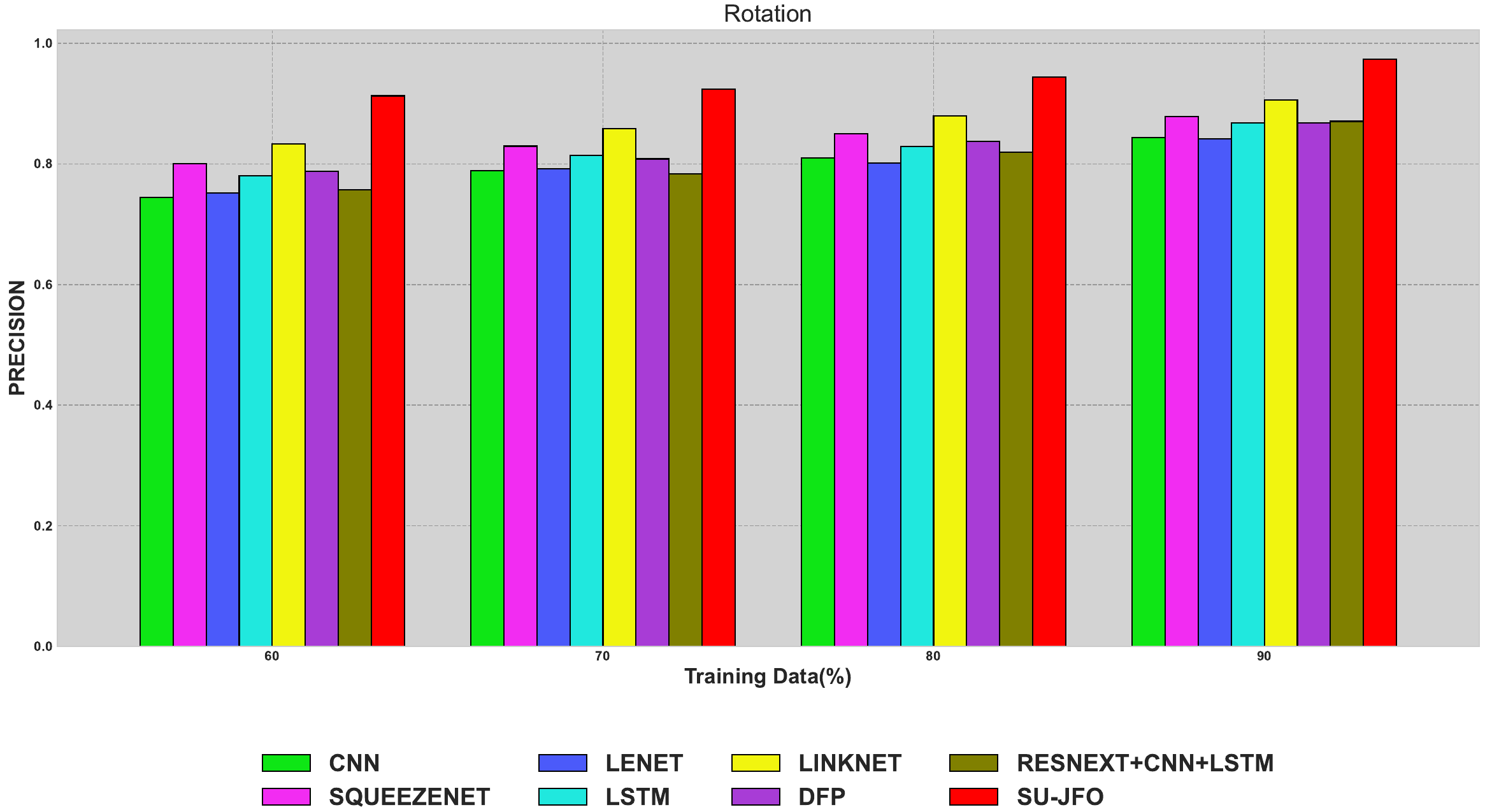}\hfill
\includegraphics[height=5cm,width=\textwidth]{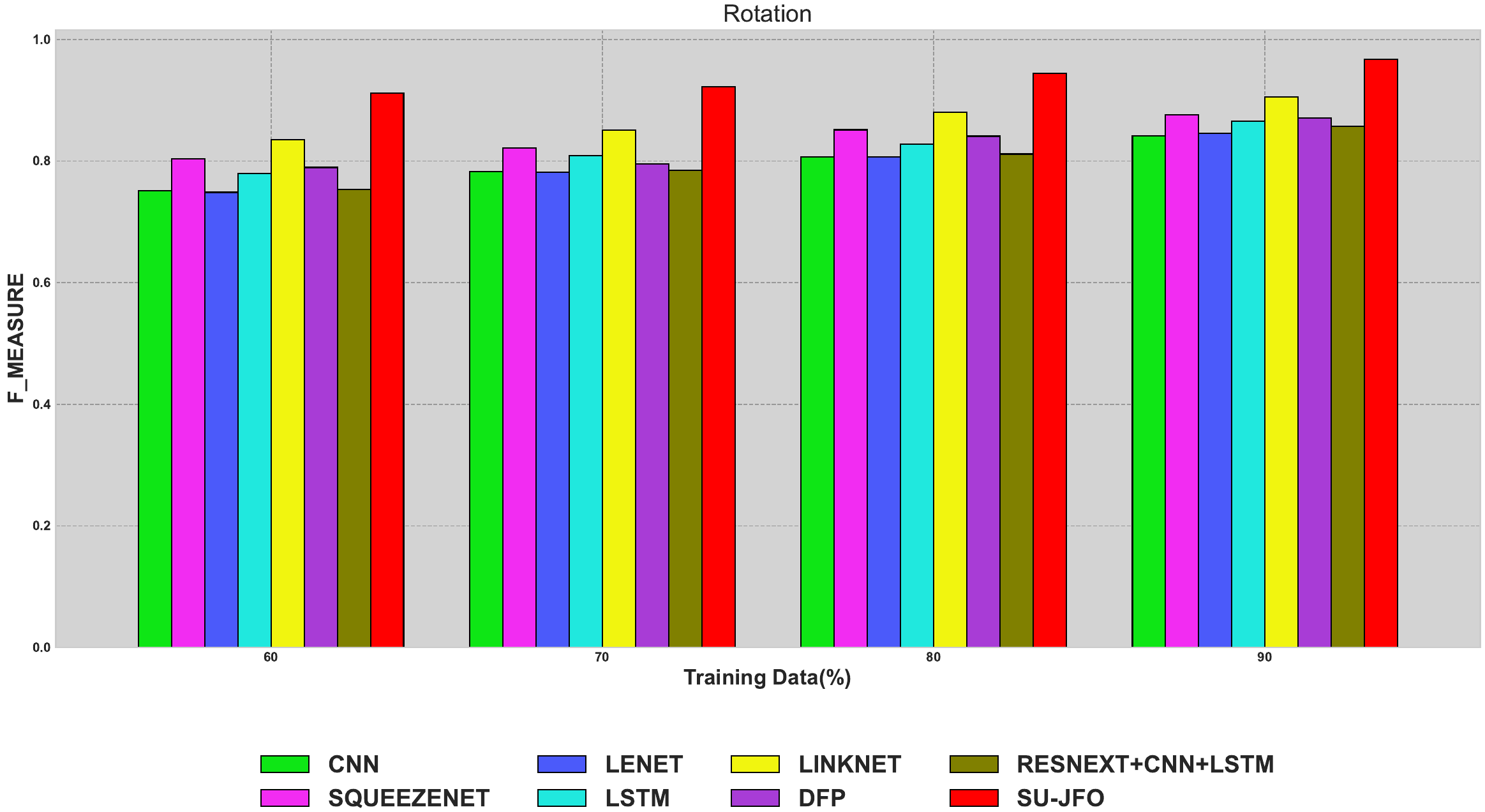}\hfill
\includegraphics[height=5cm,width=\textwidth]{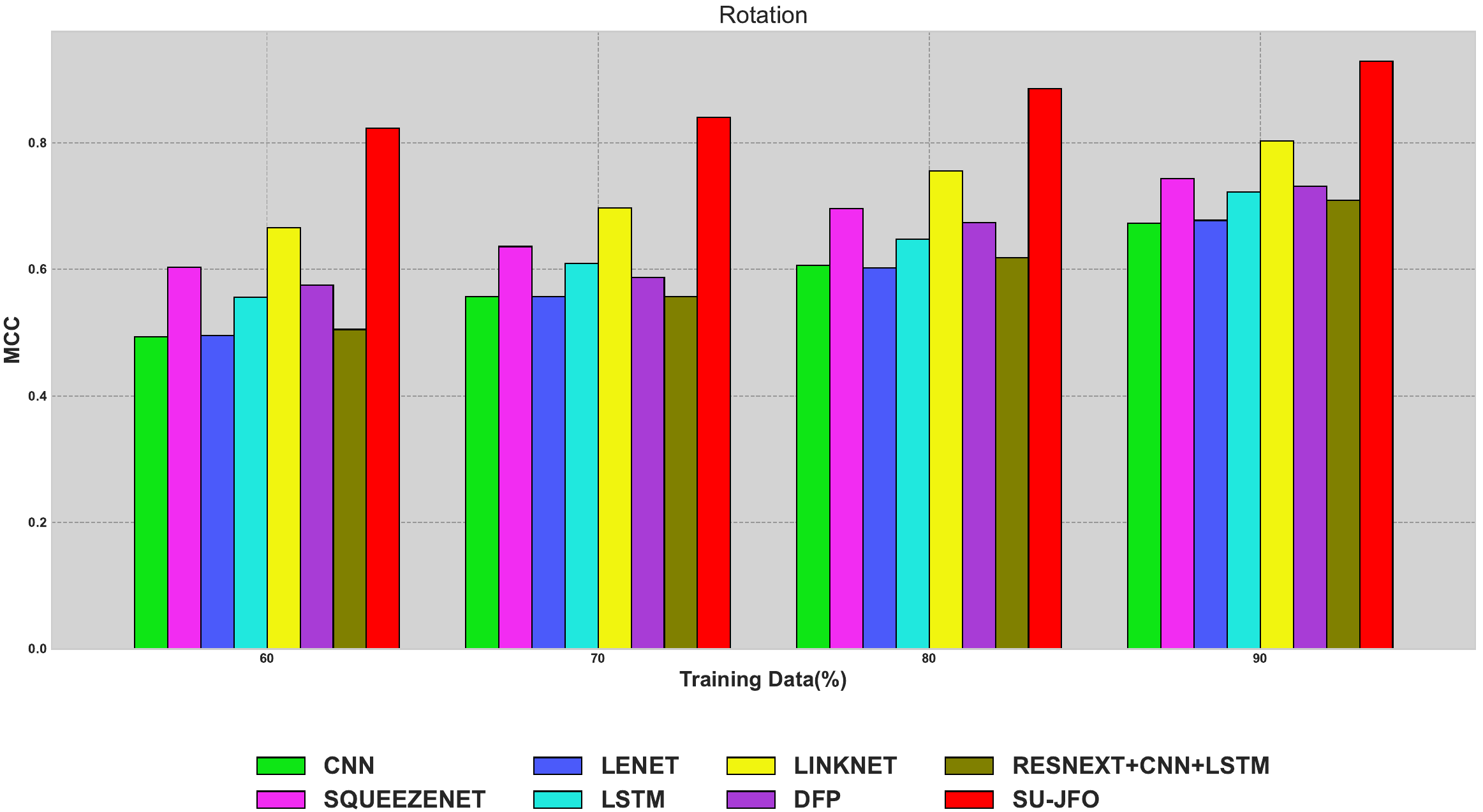}
	\caption{Assessment of SU-JFO and traditional schemes for Rotation case  on Dataset3 a) Accuracy b) Precision c) F-measure and d) MCC.}\label{Fig33}
\end{figure*}

Figure \ref{Fig33} illustrates the performance analysis of the SU-JFO method in comparison to traditional approaches for deepfake detection. The results indicate that the SU-JFO strategy achieves significantly higher values than the conventional methods highlighting its enhanced capability for accurately detecting deepfakes. 

\begin{figure*}[!t]
	\includegraphics[height=9cm,width=\textwidth]{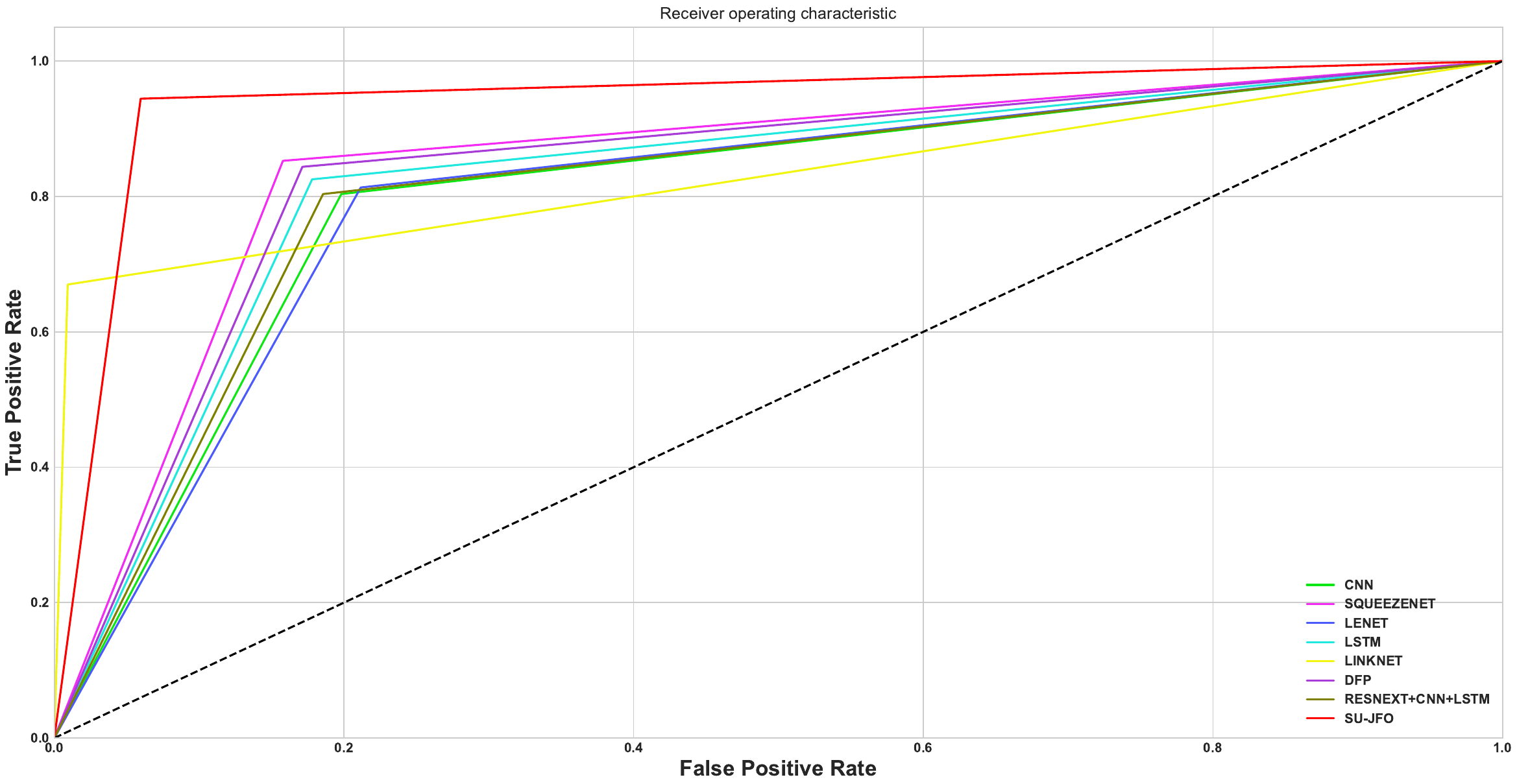}\hfill
	\caption{Evaluation on ROC for Rotation case  on Dataset3.}\label{Fig34}
\end{figure*}

\noindent \textbf{ROC Analysis on Rotation Case:}
Figure \ref{Fig34} presents the ROC analysis comparing the performance of the SU-JFO method against traditional approaches for deepfake detection. The SU-JFO method achieves a TPR exceeding 99\% while other methods such as CNN, SqueezeNet, LeNet, LSTM, LinkNet, DFP \cite{raza2022novel}, and ResNext+CNN+LSTM \cite{vamsi2022deepfake} show lower TPR. \\

\noindent \textbf{Statistical Analysis on Accuracy for Rotation Case:}
Table \ref{table13} provides a statistical comparison of the SU-JFO method with established models including CNN, SqueezeNet, LeNet, LSTM, LinkNet, DFP \cite{raza2022novel}, and ResNext+CNN+LSTM \cite{vamsi2022deepfake} . 

\begin{center}
	\begin{table}[!htbp]
		\resizebox{\textwidth}{!}{%
			\begin{tabular}{|l|l|l|l|l|l|l|l|l|}
				\hline
				\begin{tabular}[c]{@{}c@{}}\textbf{Statistical}\\ \textbf{Metrics}\end{tabular} & \textbf{CNN} & \textbf{SqueezeNet}  & \textbf{LeNet}  & \textbf{LSTM}  & \textbf{LinkNet}  & \textbf{DFP}  & \begin{tabular}[c]{@{}c@{}}\textbf{ResNext+CNN+}\\ \textbf{LSTM}\end{tabular}
   & \begin{tabular}[c]{@{}c@{}}\textbf{SU-JFO}\end{tabular}      \\ 
 \hline  
Mean & 0.791 & 0.835 & 0.792 & 0.817 & 0.865 & 0.821 & 0.799 & 0.935\\ \hline
Maximum & 0.836 & 0.872 & 0.839 & 0.861 & 0.901 & 0.866 & 0.855 & 0.965\\ \hline 
Standard Deviation  & 0.033 & 0.027 & 0.033 & 0.030 & 0.026 & 0.032 & 0.038 & 0.021  \\ \hline
Median & 0.791 & 0.833 & 0.790 & 0.814 & 0.863 & 0.815 & 0.794 & 0.932\\ \hline
Minimum & 0.747 & 0.802 & 0.748 & 0.778 & 0.833 & 0.787 & 0.753 & 0.911\\ \hline
			\end{tabular}%
		}
		\caption{Statistical Assessment on Accuracy For Rotation Case  on Dataset3.}
		\label{table13}
	\end{table}
\end{center}


\subsection{Ablation Study on SU-JFO for Dataset1 , Dataset2, and Dataset3} \label{subsection4.9}
A thorough ablation analysis was undertaken to evaluate the impact of individual attributes (a model without optimization,a model with only ear detection, and a model employing conventional RCNN) by systematically removing them or altering them thereby analyzing the overall performance. This systematic approach provides an insightful grasp of how these components contribute to the overall effectiveness of the SU-JFO framework. Table \ref{table14}, Table \ref{table15}, and Table \ref{table16} display the outcomes of the ablation evaluation performed on the SU-JFO approach, a model without optimization, a model with only ear detection, and a model employing conventional RCNN for the task of deepfake detection across dataset1 \cite{agarwal2019protecting}, dataset2 \cite{deepfaketimit}, and dataset3 \cite{yuezun2019celeb}. According to Table \ref{table14}, the SU-JFO method achieved an FNR of 0.068, the model without optimization had an FNR of 0.123, the model with only ear detection exhibited an FNR of 0.189, and the model with conventional RCNN showed an FNR of 0.113. Similarly, in dataset2 \cite{deepfaketimit} the precision values were as follows: SU-JFO approach=0.959, model without optimization=0.995, model with only ear detection=0.979, and model with conventional RCNN=0.978. Table \ref{table16} details the findings for dataset3 \cite{yuezun2019celeb}, where the SU-JFO model attained a minimum FPR of 0.054. In comparison, the model without optimization had an FPR of 0.106, the ear detection-only model recorded 0.117, and the conventional RCNN model showed an FPR of 0.084. The study demonstrates that the SU-JFO outperforms and achieves better accuracy, sensitivity, and specificity while reducing FPR and FNR. It highlights the unique strengths of SU-JFO approach in optimizing the model. The study identifies which components such as ear detection, RCNN or optimization techniques contribute most to the overall effectiveness of the model in all the three datasets.\\

\begin{table}[!htbp]
 \centering
    \resizebox{\textwidth}{!}{%
    \begin{tabular}{|l|l|l|l|l|}
     \hline
 \textbf{Metrics} & \begin{tabular}[c]{@{}c@{}}\textbf{Model}\\ \textbf{without Optimization} \end{tabular} & \begin{tabular}[c]{@{}c@{}}\textbf{Model with only}\\ \textbf{Ear Detection}\end{tabular}  & \begin{tabular}[c]{@{}c@{}}\textbf{Model}\\ \textbf{with conventional}\\ \textbf{RCNN} \end{tabular}  & \begin{tabular}[c]{@{}c@{}}\textbf{SU-JFO} \end{tabular}       \\ 
 \hline    
Accuracy & 0.881 & 0.821 & 0.903 & 0.929 \\ \hline
MCC & 0.695 & 0.579 & 0.762 & 0.855 \\ \hline 
Sensitivity  & 0.877 & 0.811 & 0.887 & 0.932 \\ \hline
F-measure & 0.921 & 0.878 & 0.935 & 0.937 \\ \hline
FNR & 0.123 & 0.189 & 0.113 & 0.068  \\ \hline
Specificity & 0.893 & 0.857 & 0.964 & 0.925  \\ \hline
NPV & 0.658 & 0.545 & 0.692 & 0.912  \\ \hline
Precision & 0.969 & 0.956 & 0.989 & 0.942  \\ \hline
FPR & 0.107 & 0.143 & 0.036 & 0.075 \\ \hline
    \end{tabular}%
    }
    \caption{ Ablation Evaluation on SU-JFO, model with only ear detection and model with conventional RCNN with Dataset1.}
    \label{table14}
\end{table}
\vspace{-5mm}
\begin{table}[!ht]
 \centering
    \resizebox{\textwidth}{!}{%
    \begin{tabular}{|l|l|l|l|l|}
     \hline
 \textbf{Metrics} & \begin{tabular}[c]{@{}c@{}}\textbf{Model}\\ \textbf{without Optimization} \end{tabular} & \begin{tabular}[c]{@{}c@{}}\textbf{Model with only}\\ \textbf{Ear Detection}\end{tabular}  & \begin{tabular}[c]{@{}c@{}}\textbf{Model}\\ \textbf{with conventional}\\ \textbf{RCNN} \end{tabular}  & \begin{tabular}[c]{@{}c@{}}\textbf{SU-JFO} \end{tabular}       \\ 
 \hline    
Accuracy & 0.896 & 0.858 & 0.873 & 0.937 \\ \hline
MCC & 0.735 & 0.669 & 0.668 & 0.835 \\ \hline 
Sensitivity  & 0.887 & 0.840 & 0.877 & 0.925 \\ \hline
F-measure & 0.931 & 0.904 & 0.916 & 0.958 \\ \hline
FNR & 0.113 & 0.160 & 0.123 & 0.075  \\ \hline
Specificity & 0.929 & 0.929 & 0.857 & 0.982  \\ \hline
NPV & 0.684 & 0.605 & 0.649 & 0.775  \\ \hline
Precision & 0.979 & 0.978 & 0.959 & 0.995  \\ \hline
FPR & 0.071 & 0.071 & 0.143 & 0.038 \\ \hline
    \end{tabular}%
    }
    \caption{ Ablation Evaluation on SU-JFO, model without optimization, model with only ear detection, and model with conventional RCNN using  Dataset2.}
    \label{table15}
\end{table}

\begin{table}[!ht]

 \centering
    \resizebox{\textwidth}{!}{%
    \begin{tabular}{|l|l|l|l|l|}
     \hline
 \textbf{Metrics} & \begin{tabular}[c]{@{}c@{}}\textbf{Model}\\ \textbf{without Optimization} \end{tabular} & \begin{tabular}[c]{@{}c@{}}\textbf{Model with only}\\ \textbf{Ear Detection}\end{tabular}  & \begin{tabular}[c]{@{}c@{}}\textbf{Model}\\ \textbf{with conventional}\\ \textbf{RCNN} \end{tabular}  & \begin{tabular}[c]{@{}c@{}}\textbf{SU-JFO} \end{tabular}       \\ 
 \hline    
Accuracy & 0.898 & 0.882 & 0.913 & 0.940 \\ \hline
MCC & 0.822 & 0.810 & 0.837 & 0.881 \\ \hline 
Sensitivity  & 0.894 & 0.883 & 0.916 & 0.946 \\ \hline
F-measure & 0.931 & 0.904 & 0.916 & 0.958 \\ \hline
FNR & 0.098 & 0.119 & 0.091 & 0.066 \\ \hline
Specificity & 0.902 & 0.881 & 0.909 & 0.934 \\ \hline
NPV & 0.892 & 0.881 & 0.913 & 0.943  \\ \hline
Precision & 0.903 & 0.883 & 0.912 & 0.938  \\ \hline
FPR & 0.106 & 0.117 & 0.084 & 0.054 \\ \hline
    \end{tabular}%
    } 
    \caption{ Ablation Evaluation on SU-JFO, model without optimization, model with only ear detection, and model with conventional RCNN using Dataset3.}
    \label{table16}
\end{table}
\subsection{Convergence analysis for Dataset1, Dataset2, and Dataset3} \label{subsection4.10}
\begin{figure}[!htbp]
	\vspace{-10mm}
	\centering
	\begin{subfigure}[b]{\textwidth} 
		\centering
		\includegraphics[width=\textwidth]{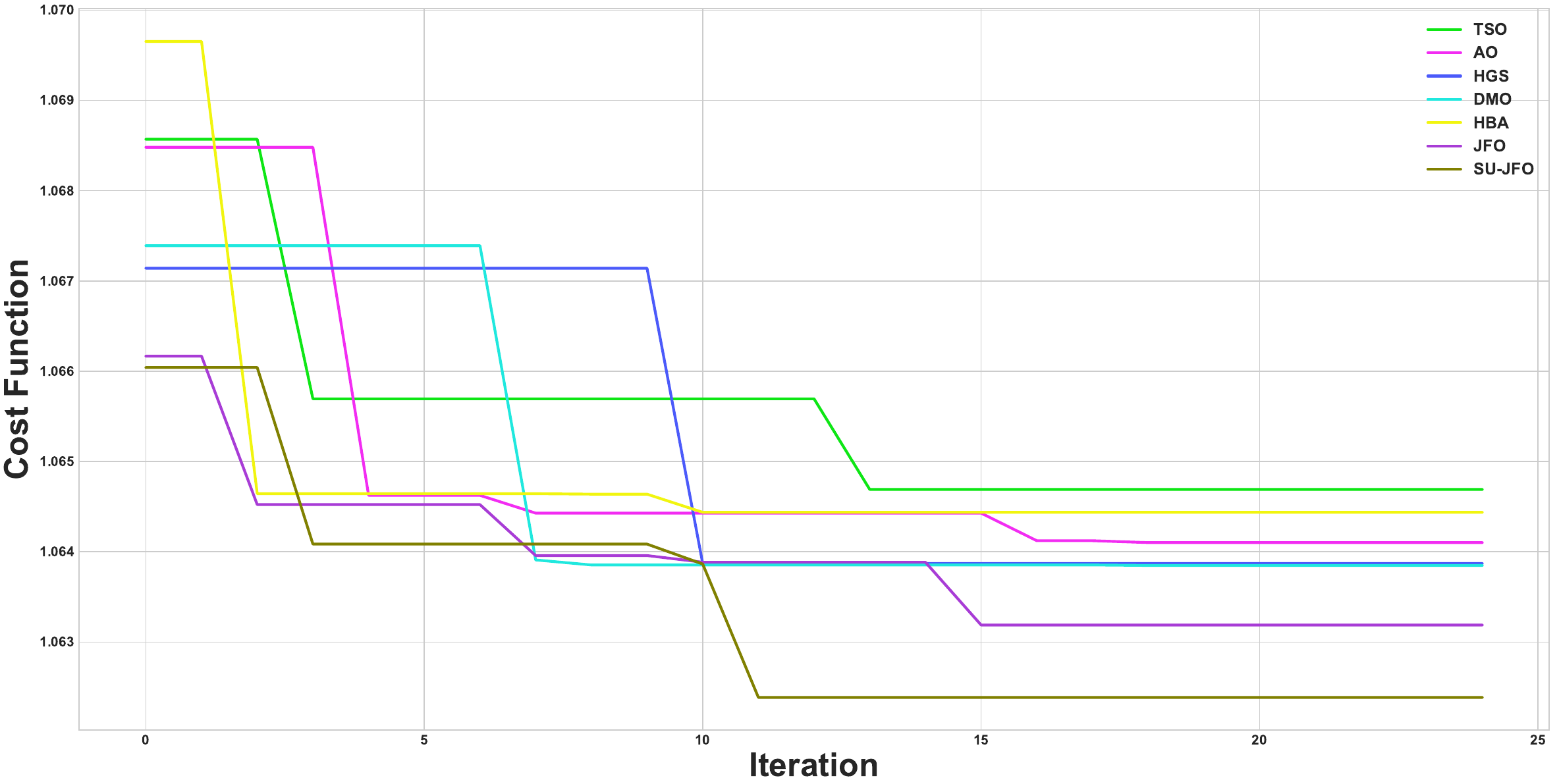}
		\caption{Convergence evaluation on SU-JFO and traditional methods on Dataset1.}
		\label{fig(a)}
	\end{subfigure}
	
	\vspace{1em} 
	
	\begin{subfigure}[b]{\textwidth}
		\vspace{-0.5em}
		\centering
		\includegraphics[width=\textwidth]{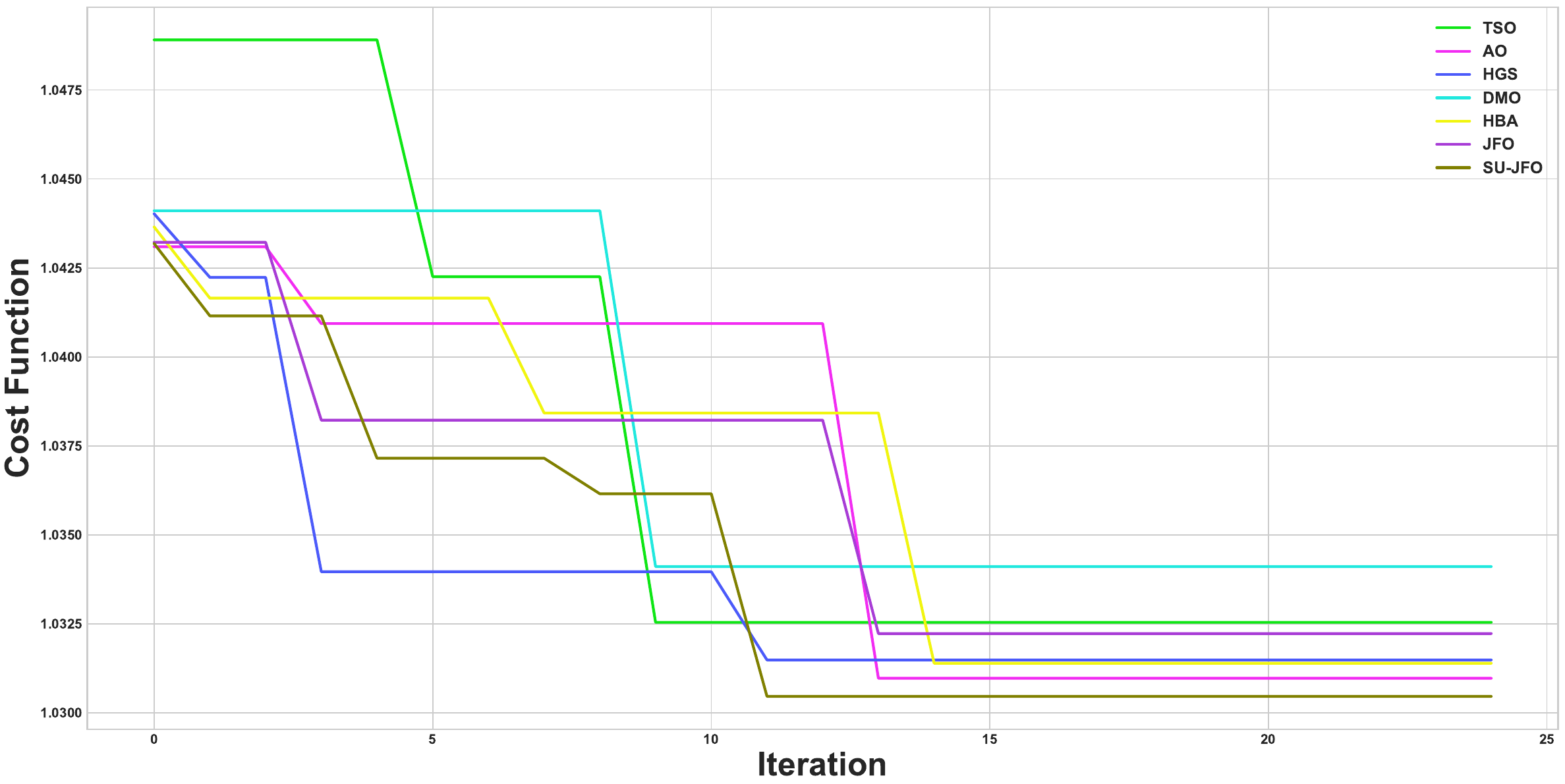}
		\caption{Convergence evaluation on SU-JFO and traditional methods on Dataset2.}
		\label{fig(b)}
	\end{subfigure}
	\begin{subfigure}[b]{\textwidth}
		\centering
		\includegraphics[width=\textwidth]{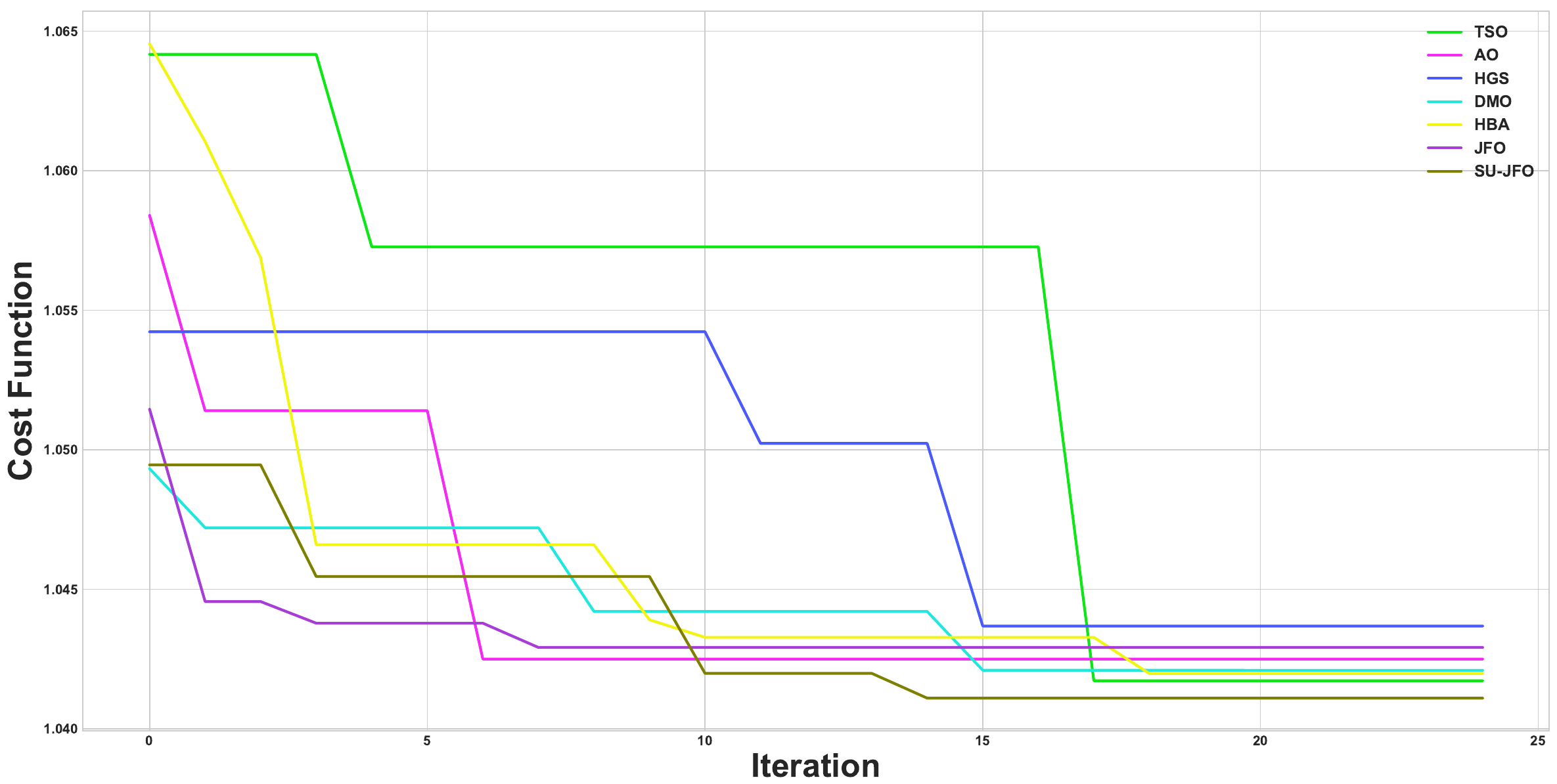}
		\caption{Convergence evaluation on SU-JFO and traditional methods on Dataset3.}
		\label{fig(c)}
	\end{subfigure}
	\caption{Convergence evaluation on SU-JFO and traditional methods on (a)Dataset1, (b)Dataset2, and (c) Dataset3.}
	\label{Fig35}
\end{figure}

Figure \ref{Fig35} illustrates the convergence study conducted on the SU-JFO method contrasting it with TSO \cite{xie2021tuna}, AO \cite{abualigah2021aquila}, HGS \cite{yang2021hunger}, DMO \cite{agushaka2022dwarf}, HBA \cite{hashim2022honey}, and JFO \cite{chou2021novel} for deepfake detection using dataset1 \cite{agarwal2019protecting}, dataset2 \cite{deepfaketimit}, and dataset3 \cite{yuezun2019celeb}. To effectively detect deepfakes, the model is expected to attain a lower cost value and achieve quicker convergence. In the initial epoch (0th iteration), both the SU-JFO and traditional methods demonstrated higher cost values. However, the cost rates decreased for all algorithms as the iterations progressed. Notably during iterations 12-25, the SU-JFO method achieved a cost rate of 1.0621 for Dataset1 \cite{agarwal2019protecting}. In contrast, traditional approaches exhibited higher cost values specifically TSO=1.0649, AO=1.0641, HGS=1.0639, DMO=1.0631, HBA=1.0647, and JFO=1.0632, respectively. Similarly, the SU-JFO scheme displayed lower cost ratings of 1.031 during 12-25 iterations  compared to conventional strategies for dataset2 \cite{deepfaketimit}. Furthermore, the SU-JFO approach also achieved lower cost ratings of 1.041 during iterations 15-25 in contrast to traditional methods such as TSO \cite{xie2021tuna}, AO \cite{abualigah2021aquila}, HGS \cite{yang2021hunger}, DMO \cite{agushaka2022dwarf}, HBA \cite{hashim2022honey}, and JFO \cite{chou2021novel} in dataset3 \cite{yuezun2019celeb}. These results indicate that the SU-JFO strategy is capable of accurately identifying deepfakes while maintaining reduced cost ratings.

\section{Conclusion} \label{section5}
This paper detects deepfakes based on ear biometric descriptor extracted with enhanced RCNN and an optimal hybrid model. Before face detection using the Viola-Jones algorithm, the input video was first converted into frames and preprocessed through resizing, normalization, gray scale conversion, and filtering. Next, the pre-processed face is subjected to feature extraction stage. AAM features have been extracted via enhanced Region-based CNN to identify ear region and extract features of the ear such as size and shape. In the end, a hybrid model leverages the strengths of both the models i.e DBN and Bi-GRU to detect deepfakes based on the retrieved features. The output from the detection phase was computed through improved score-level fusion. Additionally, self-upgraded JellyFish Optimization method (SU-JFO) was utilized to tune the weights in both the models (DBN and Bi-GRU) to enhance the detection performance. The experimental findings are reported based on four scenarios: compression, rotation, noise, pose, and illumination. A comparison is performed between the performance of the proposed work and standard models using several performance criteria such as precision, specificity, accuracy etc. Specifically, at the 70\% training data the SU-JFO method achieved a significantly higher F-measure value of 0.928 surpassing the values obtained by CNN, SqueezeNet, LeNet, LSTM, LinkNet, DFP \cite{raza2022novel}, and ResNext+CNN+LSTM \cite{vamsi2022deepfake}. In the near future, incorporation of emotions along with the ear features and movement such as the effect of anger or any other emotion that causes changes in the movement and colour of the ear could make the model more robust to detect deepfakes.

\bibliographystyle{elsarticle-num}
\bibliography{mybib}

\end{document}